\documentclass{article}
\usepackage{preprint,times}

\usepackage{amsmath,amsfonts,bm}

\def\eqref#1{equation~\ref{#1}}

\def\1{\bm{1}}

\DeclareMathAlphabet{\mathsfit}{\encodingdefault}{\sfdefault}{m}{sl}
\SetMathAlphabet{\mathsfit}{bold}{\encodingdefault}{\sfdefault}{bx}{n}

\usepackage{hyperref}
\usepackage{url}
\usepackage{amsmath}
\usepackage{amssymb}
\usepackage{booktabs}
\usepackage{makecell}
\usepackage{multirow}
\usepackage{colortbl}
\usepackage{array}
\usepackage{longtable}
\usepackage{enumitem}
\usepackage{etoc}
\usepackage{graphicx}
\usepackage{xcolor}

\definecolor{lightBlue}{RGB}{226,243,253}

\definecolor{series1}{HTML}{466F87}
\definecolor{series2}{HTML}{7BB6C8}
\definecolor{series3}{HTML}{F2D98E}
\definecolor{series4}{HTML}{E48578}
\definecolor{carddeepblue}{HTML}{466F87}
\definecolor{cardskyblue}{HTML}{7BB6C8}
\definecolor{cardyellow}{HTML}{F2D98E}
\definecolor{cardcoral}{HTML}{E48578}
\definecolor{cardgrass}{HTML}{98AF1E}
\newcommand{\seriesmark}[1]{\textcolor{#1}{\rule[0.4pt]{5pt}{5pt}}\,}

\definecolor{pophuman}{HTML}{555555}
\definecolor{popfree1136}{HTML}{466F87}
\definecolor{poptopic1136}{HTML}{7BB6C8}
\definecolor{poptag1136}{HTML}{F2D98E}
\definecolor{poptopicCL}{HTML}{E48578}
\newcommand{\popmark}[1]{\textcolor{#1}{\rule[0.4pt]{5pt}{5pt}}\,}

\newsavebox{\arpromptbox}
\newenvironment{promptbox}{%
  \setlength{\fboxsep}{7pt}%
  \begin{lrbox}{\arpromptbox}%
  \begin{minipage}{\dimexpr\linewidth-2\fboxsep-2\fboxrule\relax}%
}{%
  \end{minipage}%
  \end{lrbox}%
  \fbox{\usebox{\arpromptbox}}%
}
\usepackage{framed}
\usepackage{capt-of}
\newenvironment{promptboxlong}{%
  \setlength{\FrameSep}{7pt}%
  \setlength{\FrameRule}{\fboxrule}%
  \begin{framed}%
}{%
  \end{framed}%
}
\newcommand{\slot}[1]{{\setlength{\fboxsep}{1.5pt}\colorbox{cardskyblue!35}{<#1>}}}

\newcommand{\srctag}[2]{{\setlength{\fboxsep}{1.5pt}\colorbox{#1!22}{\textcolor{#1!75!black}{\scriptsize\bfseries\,#2\,}}}\ }
\newcommand{\srchuman}{\srctag{carddeepblue}{Human}}
\newcommand{\srcai}{\srctag{cardcoral}{AI proposal}}
\newcommand{\srcdiffer}[1]{\srctag{cardcoral}{AI #1: approach differs}}
\newcommand{\srcmatch}[1]{\srctag{cardgrass}{AI #1: approach matches}}
\newcommand{\casescore}[1]{{\color{black!60}#1\par}}
\newcommand{\caserule}{\par\smallskip{\color{black!20}\hrule height 0.4pt}\par\smallskip}

\newcommand{\MatchCite}{\textsc{Match-Cite}}
\newcommand{\MatchIdea}{\textsc{Match-Idea}}
\newcommand{\DivCite}{\textsc{Diversity-Cite}}
\newcommand{\DivIdea}{\textsc{Diversity-Idea}}

\title{IdeaTaste: Do LLMs Share the Research Taste of Human Researchers?}

\author{%
\begin{tabular}[t]{@{}l@{\hspace{2em}}l@{\hspace{2em}}l@{\hspace{2em}}l@{}}
Shanshan Zhong$^{1}$ & Chen Xu$^{1}$ & Jingjie Ning$^{1}$ & Zhengzhong Liu$^{2}$ \\[2pt]
Aviral Kumar$^{1}$ & Chenyan Xiong$^{1}$
\end{tabular} \\
\rule{0pt}{15pt}$^{1}$Carnegie Mellon University\quad $^{2}$Mohamed bin Zayed University of Artificial Intelligence
}

\begin{document}

\maketitle

\etocdepthtag.toc{main}

\begin{abstract}
Research taste, the judgment of what to read, which problems to pursue, and which methods to try, increasingly shapes the work of AI scientists, yet it remains far less understood than their ability to execute research. We introduce IdeaTaste, a benchmark that scores an LLM by how closely the research proposals it writes from a literature snapshot frozen at 2025 match the work the same communities published in 2026. Our experiments span six contemporary language models and four research areas from established to emerging fields, covering more than 90{,}000 proposals in total. LLM agents fall well short of human communities in both the literature they cite and the ideas they state, reaching about 26\% of the human level on citation matching and only 18\% on idea matching. Their populations can cite as diversely as a community and take up its tasks and methods, yet rarely arrive at the concrete approaches and research plans it pursues. Neither richer field context nor larger populations substantially close this gap. In blinded expert review, AI proposals trail the 2026 papers least in presentation and most in expected success, echoing their gap in concrete approaches, and those whose ideas no human paper shares fare worst, suggesting weaker directions rather than overlooked opportunities. These results indicate that the main bottleneck in AI research taste lies less in finding the directions a community works on than in turning them into the concrete approaches it goes on to pursue. We release IdeaTaste as an open, continuously updated leaderboard (\url{https://cxcscmu.github.io/IdeaTaste}) that benchmarks contemporary AI proposals against future publication cycles.
\end{abstract}

\section{Introduction}
\label{sec:introduction}

AI scientists can now carry out much of the research process on their own, from generating ideas and grounding them in prior work to running experiments and writing papers \citep{lu2024aiscientist,yamada2025aiscientistv2,yamada2026nature,baek2025researchagent}. Frontier laboratories are deploying them at growing scale \citep{openai2026acceleration,anthropic2026claudescience,gottweis2025coscientist,chen2026rsi}, and their work has begun to enter the scientific literature \citep{fars2026,agents4science2025}. Yet executing research well matters only when the research itself is worth doing, and this choice rests on research taste, the judgment of what to read, which problems to pursue, and which methods to try. As AI scientists take on a larger role in discovery, their taste increasingly shapes the research that comes next, but it remains far less understood than their ability to execute.
 
Evaluating research taste is challenging because it requires a reference for which choices are worth making. Existing work mostly judges one idea at a time, through expert ratings \citep{si2025canllms,si2025gap}, benchmarks with fixed targets \citep{guo2025ideabench,qiu2025aiideabench,mlrbench2025}, or model judges that predict which directions will matter \citep{tian2026foresci,wu2026cusp,tong2026taste}, and such judgments are not yet reliable \citep{sinhahajari2026rqbench,ho2026soundness,baig2026taste}. Population-level studies find that generated ideas are highly similar within and across models \citep{jiang2025hivemind,wenger2025homogeneity,chen2026diversitycollapse}, but they measure how much a population repeats itself rather than whether its choices resemble those of human researchers. What is still missing is a direct comparison between the research choices of AI scientists and those of the human community working in the same area.
 
\begin{figure}[tb]
\centering
\includegraphics[width=\textwidth]{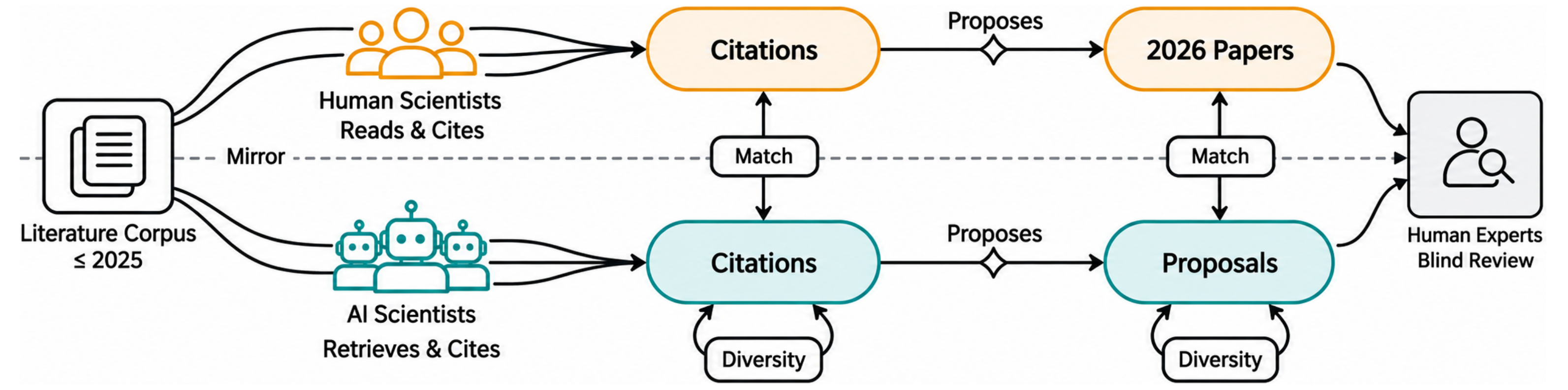}
\vspace{-15pt}
\caption{\textbf{Overview of IdeaTaste.} The literature corpus is frozen through 2025. In the upper branch, the human research community goes on to publish 2026 papers that cite work in the corpus. In the lower branch, a population of LLM agents retrieves from the corpus and writes research proposals. Matching compares the two branches, and diversity compares proposals with one another.}
\vspace{-10pt}
\label{fig:setup}
\end{figure}

The work a community goes on to publish provides such a reference, since it records which of the many possible directions its researchers actually chose. Published papers have already been used to score single ideas by their later impact and to reconstruct a held-out paper from its bibliography \citep{jiang2026hindsight,chen2026reconstruction}. We extend this reference from single papers to entire communities and introduce IdeaTaste, a benchmark that scores an LLM by how closely the population of proposals it writes matches a research community. As shown in Fig.~\ref{fig:setup}, IdeaTaste freezes the literature of a research area through 2025, lets LLM agents, standing in for the ideation stage of AI scientists, write research proposals from this snapshot, and compares them with the papers the same community published in 2026. A proposal that no 2026 paper shares may still be a good direction the community has not yet taken, so we also test, through blinded expert review, whether such proposals are overlooked opportunities or weaker ideas.

Since new research both builds on prior work and recombines it into new ideas \citep{uzzi2013atypical,foster2015tradition}, we compare proposals and papers at two levels. The literature they cite reflects what they read, and the ideas they state reflect which problems they pursue and which methods they propose. At each level, matching measures agreement with the human papers and diversity measures how varied the proposals are, and both are read against the level the 2026 papers reach among themselves. The benchmark covers four areas ranging from established to emerging fields, namely machine learning systems, code generation, world models, and automated research. Four methods differ in what an LLM agent knows about its area, from nothing beyond the corpus to a single topic \citep{lu2024aiscientist}, a map of the whole area, and the proposals its population wrote in earlier rounds \citep{schmidgall2025agentrxiv}. We benchmark six recent language models under each method, two of which were trained before 2026 and thus cannot have seen any reference paper.
 
Across more than 90{,}000 proposals, LLM agents fall well short of the research taste of human communities. Citation matching reaches about 26\% of the human level on average and idea matching only 18\%, and no population reaches half of it on either metric. The shortfall lies in what LLM agents propose more than in what they read. Given a direction, their populations can cite as diversely as a community and take up its tasks and methods, yet their proposals meet human papers far more often on problems and motivations than on concrete approaches and research plans. The gap also resists simple remedies. Models trained before 2026 match more human ideas than later ones that may have seen the reference, richer field context changes how widely a population spreads far more than how closely it follows human research, and populations up to 17 times larger level off well below the human level. In blinded expert review, AI proposals trail the 2026 papers least in presentation and most in expected success, echoing their gap in concrete approaches, and those whose ideas no human paper shares fare worst, suggesting weaker directions rather than overlooked opportunities. The main bottleneck in AI research taste thus lies less in finding the directions a community works on than in turning them into the concrete approaches it goes on to pursue. We release IdeaTaste as an open, continuously updated leaderboard that scores proposals written today against the research communities publish next. Our main contributions are as follows.

\begin{itemize}[leftmargin=0.3cm, itemsep=-0.3em, topsep=-0.2em]
\item We introduce IdeaTaste, a benchmark that evaluates the research taste of LLM agents against the papers human communities publish after a frozen literature snapshot, scoring matching and diversity on cited literature and stated ideas across four research areas.
\item Through experiments with six language models and four methods, covering more than 90{,}000 proposals, we find that LLM agents reach only a small fraction of the human level, lag more on ideas than on citations, and gain little from possible exposure to the reference, richer context, or larger populations.
\item Through direction coverage, slot-level decomposition, and blinded expert review, we show that LLM agents reach the directions a community pursues far more readily than the approaches it proposes, and that proposals missing the human record are weaker rather than merely different.
\end{itemize}

\section{Related Work}
\label{sec:related-work}

\paragraph{AI scientists and research agents.}
AI scientists are now built to run much of the research loop, from idea to experiment to written paper \citep{lu2024aiscientist,yamada2025aiscientistv2,yamada2026nature}, and the space is growing quickly \citep{luo2025llm4sr,wei2025agenticscience}. Agent Laboratory executes a human-provided idea, Co-Scientist refines hypotheses through a team of agents \citep{schmidgall2025agentlab,gottweis2025coscientist}, SciAgents and the Virtual Lab carry multi-agent designs into materials and biomedical problems \citep{ghafarollahi2025sciagents,swanson2025virtuallab}, and CycleResearcher adds automated review \citep{weng2025cycleresearcher}. Evaluation has largely followed one unit, a single system, even as agent-written work begins to enter the record from deployments at scale and a conference with AI authors and reviewers \citep{fars2026,agents4science2025,zhong2026agentwebbench}. This paper moves up a level, to a population of LLM agents proposing research from one shared literature.

\paragraph{Automated idea generation.}
Within such systems the proposal comes from an ideation step, where methods turn retrieved prior work into problems, hypotheses, and experiment designs \citep{baek2025researchagent,wang2024scimon,li2025chainofideas,hu2025nova}, and related work forecasts active topics from a knowledge graph \citep{krenn2023forecasting,gu2025impact4cast}. Ideation has also been run at population scale, on a shared archive or inside a simulated research community \citep{schmidgall2025agentrxiv,yu2025researchtown}, where outputs turn out to be substantially similar within and across models \citep{jiang2025hivemind,wenger2025homogeneity}, with further work relating this to knowledge, group size, and crowding against human output \citep{deng2026barriers,chen2026diversitycollapse,azad2026exante}, as agent societies outside research also develop shared conventions and collective bias \citep{park2023generativeagents,ashery2025conventions} and agent ecosystems concentrate into homogeneous offerings \citep{ling2026agentskills}. Such comparisons are mostly internal to the population. This study measures that similarity in the literature a proposal cites and in the idea it states, and pairs each with agreement against human papers.

\paragraph{Evaluating research ideas.}
Evaluation supplies the external reference such comparisons lack. Experts judge one idea at a time, though the initial advantage of model ideas narrowed once executed \citep{si2025canllms,si2025gap}, while benchmarks score ideas against fixed targets \citep{guo2025ideabench,qiu2025aiideabench,mlrbench2025}, often leaving the verdict to model judges \citep{zheng2023judge,huang2025lotbench}, now asked which directions will matter \citep{tian2026foresci,wu2026cusp,tong2026taste}. Those verdicts are not yet reliable, as judges disagree with experts on novelty \citep{sinhahajari2026rqbench}, trail them at comparison \citep{baig2026taste}, and rate unsound ideas as sound \citep{ho2026soundness}, while human review is itself inconsistent \citep{cortes2021inconsistency}. Published research offers a steadier reference, used to score ideas by future impact \citep{jiang2026hindsight}, to recover a held-out paper \citep{chen2026reconstruction}, and to profile ideas against a taxonomy on which model output is narrower \citep{chen2026gap}, though citation lists are an imperfect standard \citep{sahu2026citationgt}. This paper extends that reference from single papers to what a research community published after its literature was frozen, comparing that with what a population cites and proposes.

\vspace{0.1cm}
\section{IdeaTaste}
\label{sec:benchmark}

IdeaTaste evaluates the research taste of LLM agents by comparing the proposals they write from a frozen literature with the papers a human community later publishes. Sec.~\ref{sec:corpus} builds the literature corpus and the human reference, Sec.~\ref{sec:metrics} defines four metrics on what each side cites and proposes, and Sec.~\ref{sec:systems} describes the methods.

\subsection{Data Construction}
\label{sec:corpus}

We cover four research areas: machine learning systems and code generation, established fields with large literatures, and world models and automated research, smaller fields that are still emerging. For each area, we split the literature at the end of 2025. Papers published through 2025 form the corpus $U$, which LLM agents search and propose from. Papers published in 2026 form the human reference $H$, which records what the community actually pursued after this cutoff. The cutoff can be moved forward as new papers appear.

Both sets are collected by following citations from 20 to 22 landmark papers hand-picked for each area. The corpus first takes in papers published through 2025 that cite a landmark and then adds every paper cited by at least two of these citing papers. This second step brings in shared background that may not use the terminology of the area, such as program synthesis work for code generation. The human reference combines 2026 papers that cite a landmark with papers returned by topic search, screens them against a written area definition, and removes surveys. It covers January to August 2026, and Table~\ref{tab:corpus} reports the sizes of both sets. Appendix~\ref{app:corpus} gives the area definitions, the landmarks, further construction details, and a coverage audit against authoritative surveys.

\begin{table}[tb]
\centering
\caption{Data statistics. Left: sizes of the corpus $U$, published through 2025, and the human reference $H$, published in 2026. Right: papers in the human reference by publication month. Collection ends in August 2026, so the last month is incomplete.
}
\label{tab:corpus}
\begin{minipage}[c]{0.45\textwidth}%
\centering
\footnotesize
\setlength{\tabcolsep}{3pt}
\resizebox{\linewidth}{!}{%
\begin{tabular}{@{}lrr@{}}
\toprule
Area & Corpus $U$ & Reference $H$ \\
\midrule
\seriesmark{series1}Machine learning systems & 88{,}184 & 5{,}188 \\
\seriesmark{series2}Code generation          & 94{,}674 & 3{,}375 \\
\seriesmark{series3}World models             & 50{,}271 & 1{,}867 \\
\seriesmark{series4}Automated research       & 30{,}264 & 1{,}136 \\
\bottomrule
\end{tabular}
}
\end{minipage}
\hfill
\begin{minipage}[c]{0.54\textwidth}%
\centering
\vspace*{5pt}%
\includegraphics[height=0.24\linewidth]{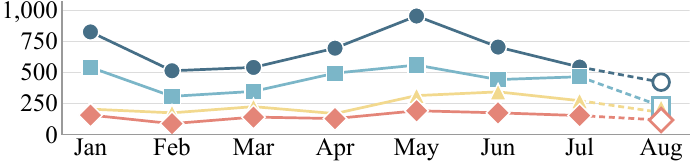}%
\end{minipage}
\vspace{-5pt}
\end{table}

\subsection{Evaluation Design}
\label{sec:metrics}
\label{sec:formulation}

Writing a research document, whether a human paper or an LLM agent proposal, involves two choices, selecting the prior work to build on and formulating an idea in relation to it. LLM agents make these choices using only the frozen corpus $U$, while the human reference $H$ records the community's subsequent choices after the cutoff. We therefore compare $H$ with the proposals $A$ of a population at the two corresponding levels \citep{uzzi2013atypical,foster2015tradition}. The resulting information asymmetry across the cutoff is discussed in Appendix~\ref{app:leakage}. At the citation level, each document $d\in H\cup A$ is represented by its citation set $\mathcal{C}(d)$, the papers it cites that lie in $U$. At the idea level, it is represented by a summary that a language model extracts from its abstract under a fixed schema of five slots $\mathcal{I}$, namely contribution type, problem, motivation, proposed approach, and research plan, as detailed in Appendix~\ref{app:eval}. At each level, matching measures agreement with the human reference and diversity measures how varied the proposals are, giving citation matching \MatchCite{}, citation diversity \DivCite{}, idea matching \MatchIdea{}, and idea diversity \DivIdea{}. All four lie in $[0,1]$, and higher values mean closer agreement with the human reference or greater variety. Applying them to $H$ itself, with each human paper compared only with the other human papers, gives the human level against which populations are read.

\subsubsection{Level 1: Literature Selection}

\paragraph{\MatchCite{}.} We compare a human paper $d_h\in H$ and a proposal $d_a\in A$ by the Jaccard similarity of their citation sets, $J(d_h,d_a)=|\mathcal{C}(d_h)\cap\mathcal{C}(d_a)|/|\mathcal{C}(d_h)\cup\mathcal{C}(d_a)|$, and match each document with its closest counterpart on the other side:
\begin{equation}
J_{A\to H}=\frac{1}{|A|}\sum\nolimits_{d_a\in A}\max_{d_h\in H}J(d_h,d_a),
\qquad
J_{H\to A}=\frac{1}{|H|}\sum\nolimits_{d_h\in H}\max_{d_a\in A}J(d_h,d_a).
\label{eq:cite_match}
\end{equation}
$J_{A\to H}$ measures whether each proposal cites like some human paper, which a population can achieve from a narrow corner of the literature. $J_{H\to A}$ measures whether the sources of each human paper appear among the proposals, which requires the population to spread across the literature. \MatchCite{} is  $M_{\mathrm{cite}}=2J_{A\to H}J_{H\to A}/(J_{A\to H}+J_{H\to A})$, which is high only when both are.

\paragraph{\DivCite{}.} Citation diversity asks whether proposals read enough without returning to the same sources. It is the product $D_{\mathrm{cite}}=Q\,S$ of the sufficiency $Q$ and the spread $S$,
\begin{equation}
Q=\frac{1}{|A|}\sum\nolimits_{d_a\in A}\min\left(|\mathcal{C}(d_a)|/\tau,1\right),
\qquad
S=\left|\bigcup\nolimits_{d_a\in A}\mathcal{C}(d_a)\right|/\sum\nolimits_{d_a\in A}|\mathcal{C}(d_a)|,
\label{eq:cite_diversity}
\end{equation}
where $\tau$ is the mean citation set size of the human papers in the area. See Appendix~\ref{app:tau}. Because $Q$ is capped at one, citing more than a typical human paper earns nothing, and $S$ falls as proposals cite the same papers. $D_{\mathrm{cite}}$ is thus high only when proposals both read enough and read differently.

\subsubsection{Level 2: Idea Formulation}

A language-model judge compares two idea summaries slot by slot, with $b_i(d,d')\in\{0,1\}$ recording whether $d$ and $d'$ agree on slot $i\in\mathcal{I}$. To keep judging feasible, each $d$ is compared only with $\mathcal{N}_X(d)$, its five nearest other documents in a pool $X$ under hybrid sparse and dense retrieval over titles and abstracts, and $m_i(d,Y)=\max_{d'\in Y}b_i(d,d')$ indicates whether slot $i$ of $d$ is matched within a set $Y$. As most paper pairs share a contribution type but few share a research plan, each slot is weighted by its rarity, $w_i=-\log_2 p_i$ for $p_i>0$, where $p_i$ is the fraction of human papers $d_h$ whose slot $i$ is matched within $\mathcal{N}_H(d_h)$. With the weights and their sum $W$ fixed per area, $s(d,Y)=\sum_{i\in\mathcal{I}}w_i\,m_i(d,Y)/W$ is the weighted fraction of slots of $d$ matched within $Y$.

\paragraph{\MatchIdea{}.} A slot of a human paper counts as covered when at least one proposal retrieved for it matches on that slot, and \MatchIdea{} is the weighted fraction of human slots covered.

\paragraph{\DivIdea{}.} Applying the same matching within a population, we compare each proposal with its peers $\mathcal{P}(d_a)=\mathcal{N}_A(d_a)\cup\{d_{a'}\in A: d_a\in\mathcal{N}_A(d_{a'})\}$, namely the proposals it retrieves and those that retrieve it, and \DivIdea{} is one minus the weighted fraction of its slots that some peer shares. Both idea metrics are thus averages of $s$, taken over human papers against the proposals and over proposals against their peers:
\begin{equation}
M_{\mathrm{idea}}=\sum\nolimits_{d_h\in H}s\big(d_h,\mathcal{N}_A(d_h)\big)/|H|,
\qquad
D_{\mathrm{idea}}=1-\sum\nolimits_{d_a\in A}s\big(d_a,\mathcal{P}(d_a)\big)/|A|.
\label{eq:idea_metrics}
\end{equation}
Appendix~\ref{app:eval} gives the retrieval and judging configurations that both idea metrics depend on, and validates the judge against human annotators.

\subsection{Methods}
\label{sec:systems}

All four methods share one LLM agent implementation and differ only in what it knows about its area. The LLM agent receives the area definition used to screen the human reference and writes one proposal within a fixed step budget, taking one action per step: planning, searching the corpus, pinning retrieved papers, revising the draft, compressing its history into notes, or submitting. It can cite only retrieved papers, so its citations always lie in $U$. The methods vary its profile and actions:
\begin{itemize}[leftmargin=0.3cm, itemsep=-0.3em, topsep=-0.2em]
\item \textsc{free} only states that the LLM agent is a researcher in the area, revealing what it pursues without an explicit prior.
\item \textsc{topic} assigns one research topic and leaves the choice of problem open \citep{lu2024aiscientist}, testing how a broad direction shapes proposals. Topics are allocated in proportion to corpus frequency, with recent years upweighted.
\item \textsc{landscape} assigns no topic but lists the area's most common task tags, method tags, and task--method pairs with their shares among 2025 papers, plus an action to retrieve papers by tag or pair. It thus exposes the area's structure and how crowded each direction is, which researchers weigh when choosing between established and new directions \citep{foster2015tradition}.
\item \textsc{frontier} lets LLM agents see one another's work \citep{schmidgall2025agentrxiv}. It assigns topics like \textsc{topic} but runs in rounds, freezing each round's proposals and research traces into a snapshot that later rounds query through two extra actions, as researchers read peers' work. Peer records are separate from literature retrieval, uncitable, and marked as not scientifically validated.
\end{itemize}
These methods test whether an LLM agent's view of its area changes the literature and ideas its population selects. See Appendix~\ref{app:methods} for more details. %

\begin{table}[tb]
\centering
\scriptsize
\setlength{\tabcolsep}{2.5pt}
\caption{Main results (\%). For each area, Match reports \MatchCite{} and \MatchIdea{}, and Diversity reports \DivCite{} and \DivIdea{}. The Human row scores the 2026 papers against one another and gives the human level. Model groups are ordered by release date, and bold marks the best score in each column within a group.}%
\label{tab:main}
\resizebox{\textwidth}{!}{%
\begin{tabular}{l *{16}{r}}
\toprule
& \multicolumn{4}{c}{Machine learning systems} & \multicolumn{4}{c}{Code generation} & \multicolumn{4}{c}{World models} & \multicolumn{4}{c}{Automated research} \\
\cmidrule(lr){2-5}\cmidrule(lr){6-9}\cmidrule(lr){10-13}\cmidrule(lr){14-17}
& \multicolumn{2}{c}{Match} & \multicolumn{2}{c}{Diversity} & \multicolumn{2}{c}{Match} & \multicolumn{2}{c}{Diversity} & \multicolumn{2}{c}{Match} & \multicolumn{2}{c}{Diversity} & \multicolumn{2}{c}{Match} & \multicolumn{2}{c}{Diversity} \\
\addlinespace[1pt]
Population & Cite & Idea & Cite & Idea & Cite & Idea & Cite & Idea & Cite & Idea & Cite & Idea & Cite & Idea & Cite & Idea \\
\midrule
Human & 20.84 & 13.56 & 13.50 & 84.97 & 18.75 & 11.25 & 18.64 & 87.41 & 16.38 & 13.87 & 14.68 & 84.31 & 18.51 & 10.53 & 21.69 & 87.52 \\
\midrule
\rowcolor{lightBlue}\multicolumn{17}{l}{\textbf{2025-04-05, Llama 4 Scout}~\citep{meta2025llama4}} \\
\addlinespace[2pt]
\textsc{free} & 3.44 & 4.32 & 3.89 & 34.91 & 2.72 & 3.41 & 5.35 & 42.70 & 2.65 & 2.74 & 2.96 & 35.32 & \textbf{6.42} & \textbf{4.79} & 3.87 & 38.58 \\
\textsc{topic} & \textbf{3.87} & 4.93 & 13.71 & 67.47 & \textbf{3.29} & \textbf{4.21} & 12.77 & 63.90 & 3.42 & \textbf{4.42} & 9.41 & 61.39 & 5.44 & 3.85 & 13.84 & \textbf{68.53} \\
\textsc{landscape} & 2.51 & 4.91 & 5.51 & 33.05 & 2.10 & 2.19 & 2.44 & 35.27 & 1.91 & 1.81 & 2.20 & 27.65 & 5.14 & 4.37 & 3.70 & 38.62 \\
\textsc{frontier} & 2.76 & \textbf{5.95} & \textbf{14.93} & \textbf{71.64} & 3.14 & 3.72 & \textbf{12.85} & \textbf{64.20} & \textbf{3.95} & 3.94 & \textbf{9.91} & \textbf{62.91} & 4.71 & 3.32 & \textbf{14.06} & 67.21 \\
\midrule
\rowcolor{lightBlue}\multicolumn{17}{l}{\textbf{2025-10-15, Claude Haiku 4.5}~\citep{anthropic2025haiku45}} \\
\addlinespace[2pt]
\textsc{free} & \textbf{4.48} & 1.70 & 13.02 & 45.34 & 3.90 & \textbf{1.93} & 17.36 & 51.53 & 4.37 & 2.17 & 12.04 & 48.29 & 5.86 & 2.79 & 12.11 & 28.34 \\
\textsc{topic} & 4.24 & 1.99 & \textbf{30.78} & 77.60 & \textbf{4.00} & 1.81 & \textbf{31.68} & \textbf{72.51} & 4.94 & \textbf{2.85} & \textbf{22.31} & 66.17 & \textbf{6.14} & 2.55 & \textbf{27.86} & \textbf{67.64} \\
\textsc{landscape} & 3.97 & 1.48 & 12.79 & 45.83 & 3.71 & 1.57 & 17.78 & 57.87 & 4.29 & 2.59 & 12.06 & 40.44 & 5.80 & 3.14 & 13.20 & 33.69 \\
\textsc{frontier} & 4.41 & \textbf{2.30} & 29.82 & \textbf{80.66} & 3.90 & 1.81 & 28.76 & 70.32 & \textbf{5.03} & 2.41 & 22.07 & \textbf{69.06} & 6.07 & \textbf{3.17} & 26.82 & 65.44 \\
\midrule
\rowcolor{lightBlue}\multicolumn{17}{l}{\textbf{2026-07-21, Gemini 3.5 Flash-Lite}~\citep{google2026gemini35flashlite}} \\
\addlinespace[2pt]
\textsc{free} & 3.73 & 1.36 & 3.01 & 16.88 & 3.72 & 1.52 & 6.29 & 15.40 & 4.89 & 3.38 & 4.76 & 18.76 & 7.18 & 2.22 & 5.52 & 12.92 \\
\textsc{topic} & 4.59 & 2.29 & \textbf{14.11} & \textbf{53.98} & 4.23 & \textbf{2.19} & \textbf{17.19} & 49.28 & \textbf{5.33} & 3.86 & \textbf{13.35} & 52.52 & 6.58 & 2.14 & 20.22 & 62.21 \\
\textsc{landscape} & 3.67 & 1.38 & 2.85 & 15.01 & 3.88 & 1.47 & 6.07 & 16.56 & 4.44 & 2.31 & 3.62 & 14.35 & \textbf{7.18} & 2.14 & 5.12 & 9.86 \\
\textsc{frontier} & \textbf{4.65} & \textbf{2.32} & 14.05 & 53.28 & \textbf{4.23} & 1.85 & 16.86 & \textbf{52.38} & 5.28 & \textbf{3.87} & 12.90 & \textbf{52.68} & 6.34 & \textbf{2.28} & \textbf{20.57} & \textbf{63.31} \\
\midrule
\rowcolor{lightBlue}\multicolumn{17}{l}{\textbf{2026-07-31, DeepSeek-V4-Flash}~\citep{deepseekai2026v4}} \\
\addlinespace[2pt]
\textsc{free} & 3.94 & 0.92 & 7.65 & 33.13 & 3.51 & 0.79 & 13.00 & 35.38 & 4.77 & 1.45 & 13.50 & 42.79 & 6.40 & 1.67 & 7.57 & 23.16 \\
\textsc{topic} & 4.63 & \textbf{1.18} & 33.88 & 67.65 & \textbf{4.63} & 0.99 & 37.46 & 64.30 & \textbf{5.16} & 1.76 & 32.51 & 69.06 & \textbf{7.72} & \textbf{2.09} & 25.39 & 55.80 \\
\textsc{landscape} & 3.53 & 1.05 & 10.82 & 29.33 & 3.62 & 0.96 & 14.24 & 38.46 & 4.22 & \textbf{1.90} & 13.00 & 40.58 & 5.33 & 1.87 & 9.55 & 22.50 \\
\textsc{frontier} & \textbf{4.78} & 0.94 & \textbf{42.15} & \textbf{82.55} & 4.58 & \textbf{1.00} & \textbf{46.96} & \textbf{77.25} & 4.98 & 1.41 & \textbf{39.83} & \textbf{71.38} & 7.02 & 1.83 & \textbf{36.17} & \textbf{73.59} \\
\midrule
\rowcolor{lightBlue}\multicolumn{17}{l}{\textbf{2026-08-14, Qwen3.8-27B}~\citep{qwen2026qwen38}} \\
\addlinespace[2pt]
\textsc{free} & 3.96 & 0.95 & 8.22 & 31.92 & 4.44 & 1.11 & 17.38 & 45.91 & 5.32 & 1.51 & 13.30 & 37.35 & 6.30 & 1.79 & 6.94 & 18.01 \\
\textsc{topic} & 4.89 & 1.38 & 32.36 & 74.25 & 4.21 & \textbf{1.40} & 40.35 & 66.94 & 5.59 & 2.20 & 36.12 & \textbf{66.34} & 7.17 & \textbf{2.66} & 29.08 & 59.14 \\
\textsc{landscape} & 3.70 & 1.00 & 10.92 & 37.27 & 3.47 & 1.01 & 16.18 & 45.42 & 4.73 & 1.59 & 11.28 & 25.58 & 6.58 & 1.86 & 8.98 & 22.32 \\
\textsc{frontier} & \textbf{4.94} & \textbf{1.49} & \textbf{37.97} & \textbf{78.03} & \textbf{5.05} & 1.33 & \textbf{46.53} & \textbf{67.09} & \textbf{5.91} & \textbf{2.35} & \textbf{36.94} & 65.75 & \textbf{7.72} & 2.39 & \textbf{31.05} & \textbf{63.81} \\
\midrule
\rowcolor{lightBlue}\multicolumn{17}{l}{\textbf{2026-09-03, K2 Horizon 375B A23B}~\citep{ifm2026k2horizon}} \\
\addlinespace[2pt]
\textsc{free} & 4.21 & 0.94 & 9.30 & 48.63 & 3.82 & 1.03 & 17.96 & 60.97 & 5.01 & 1.56 & 14.16 & 65.68 & \textbf{7.21} & 1.80 & 10.52 & 38.38 \\
\textsc{topic} & 4.44 & 1.19 & \textbf{35.70} & 82.82 & \textbf{4.56} & 1.05 & 36.83 & 76.99 & 4.97 & 1.32 & \textbf{29.70} & 80.48 & 7.04 & 2.09 & 33.37 & \textbf{80.26} \\
\textsc{landscape} & 4.09 & 1.10 & 10.31 & 51.96 & 3.73 & \textbf{1.10} & 21.35 & 66.92 & 4.51 & \textbf{1.65} & 12.00 & 53.09 & 6.75 & 1.87 & 10.40 & 39.87 \\
\textsc{frontier} & \textbf{4.59} & \textbf{1.27} & 34.85 & \textbf{84.58} & 4.24 & 1.03 & \textbf{38.14} & \textbf{77.50} & \textbf{5.05} & 1.52 & 29.45 & \textbf{82.76} & 6.82 & \textbf{2.36} & \textbf{34.13} & 78.02 \\
\bottomrule
\end{tabular}%
}
\vspace{-10pt}
\end{table}

\section{Experiments}
\label{sec:experiments}

\paragraph{LLM agents.} We run the four methods of Sec.~\ref{sec:systems} in every area with the six language models in Table~\ref{tab:main}. Each LLM agent has a budget of 20 steps, and each population holds 300 proposals, which \textsc{frontier} writes in two rounds of 150. The six models form two groups by design. Four were released in 2026 and power current AI scientists, so their training data may overlap the human reference, but leaving them out would confine the benchmark to outdated models. The other two were trained before 2026 and cannot have seen any 2026 paper, so their results are free of leakage by construction. Comparing the two groups tests whether possible exposure to the reference helps LLM agents anticipate 2026 research. Appendix~\ref{app:methods} gives implementation details.

\paragraph{Automatic evaluation.} All populations are scored with the metrics of Sec.~\ref{sec:metrics}, and scoring the 2026 papers against one another gives the human level. The slot-level judge is GPT-5.6 Luna, outside the evaluated model families and validated against human annotators in Appendix~\ref{app:judge}. 

\paragraph{Human evaluation.} Beyond whether LLM agents propose what the community pursues, we use human evaluation to examine how good their proposals are, and in particular whether those without a human counterpart are weak directions or sound ones the community has not yet taken. Eleven reviewers, all with reviewing experience, assess 192 documents without knowing their source. The documents are split evenly between proposals and 2026 papers and balanced across areas and models. Half come from matched pairs, each a proposal and the 2026 paper it overlaps under the judge of \MatchIdea{}, and the other half have no counterpart on the other side. Each document receives two reviews. Besides the dimensions of a standard machine learning conference review form, reviewers rate feasibility and expected success, since an idea can look original yet fail in execution \citep{si2025gap}, and guess whether the document is AI-generated. Appendix~\ref{app:human} details the design.

\begin{table}[tb]
\centering
\scriptsize
\caption{Human evaluation of idea quality. Reviewers score 2026 papers and LLM agent proposals without knowing their source, and scores are mapped onto $[0,100]$. ``Judged AI" is the share of reviews that label a document as AI-generated. Match is decided by slot-level judge \MatchIdea{}.}
\label{tab:human}
\resizebox{\textwidth}{!}{%
\begin{tabular}{l *{8}{r}}
\toprule
& \multicolumn{3}{c}{Quality} & \multicolumn{2}{c}{Execution} & \multicolumn{3}{c}{Overall} \\
\cmidrule(lr){2-4}\cmidrule(lr){5-6}\cmidrule(lr){7-9}
Population & Soundness & Presentation & Contribution & Feasibility & Success & Rating & Confidence & Judged AI \\
\midrule
Human                 & 49.83 & 62.50 & 53.99 & \textbf{75.39} & \textbf{70.70} & \textbf{49.48} & 56.51 & 39.06 \\
\rowcolor{lightBlue}LLM agent (mean) & 38.54 & 58.16 & 45.31 & 63.80 & 50.39 & 38.54 & 59.90 & 86.46 \\
\addlinespace[2pt]
\quad Llama 4 Scout         & 13.54 & 30.21 & 17.71 & 55.47 & 35.94 & 22.50 & 60.94 & 81.25 \\
\quad Claude Haiku 4.5      & 38.54 & 57.29 & 41.67 & 67.97 & 50.78 & 39.38 & 55.47 & 84.38 \\
\quad Gemini 3.5 Flash-Lite & 39.58 & 56.25 & 45.83 & 64.06 & 58.59 & 35.62 & 62.50 & 90.62 \\
\quad DeepSeek-V4-Flash     & \textbf{51.04} & \textbf{75.00} & \textbf{63.54} & 67.19 & 61.72 & 46.88 & 60.16 & 87.50 \\
\quad Qwen3.8-27B           & 44.79 & 70.83 & 52.08 & 60.94 & 46.88 & 44.38 & 59.38 & 87.50 \\
\quad K2 Horizon 375B A23B  & 43.75 & 59.38 & 51.04 & 67.19 & 48.44 & 42.50 & 60.94 & 87.50 \\
\midrule
\rowcolor{lightBlue}\multicolumn{9}{l}{\textbf{Match}, proposals and the 2026 papers they match} \\
\addlinespace[1pt]
Human                 & 50.00 & 60.42 & 53.47 & 74.48 & 69.01 & 50.00 & 55.99 & 40.62 \\
LLM agent          & 40.62 & 58.68 & 45.49 & 66.15 & 54.17 & 39.17 & 60.16 & 86.46 \\
$\Delta$ LLM agent $-$ Human & $-$9.38 & $-$1.74 & $-$7.99 & $-$8.33 & $-$14.84 & $-$10.83 & $+$4.17 & $+$45.83 \\
\midrule
\rowcolor{lightBlue}\multicolumn{9}{l}{\textbf{Non-match}, documents with no counterpart on the other side} \\
\addlinespace[1pt]
Human                 & 49.65 & 64.58 & 54.51 & 76.30 & 72.40 & 48.96 & 57.03 & 37.50 \\
LLM agent          & 36.46 & 57.64 & 45.14 & 61.46 & 46.61 & 37.92 & 59.64 & 86.46 \\
$\Delta$ LLM agent $-$ Human & $-$13.19 & $-$6.94 & $-$9.37 & $-$14.84 & $-$25.79 & $-$11.04 & $+$2.61 & $+$48.96 \\
\bottomrule
\end{tabular}%
}
\vspace{-10pt}
\end{table}

\paragraph{Main results.}
\label{sec:main}
\label{sec:human}
Table~\ref{tab:main} scores every population on the four metrics against the human level, and Table~\ref{tab:human} reports how reviewers rate a blinded sample of proposals and 2026 papers. Together they yield four findings.
\begingroup
\setlength{\leftmargini}{1.2em}
\begin{itemize}
\setlength{\itemsep}{3pt}\setlength{\parskip}{0pt}\setlength{\topsep}{4pt}
\vspace{-3pt}
\item \textbf{LLM agents fall far short of the research taste of human communities.} On average, LLM agents reach about 26\% of the human level on \MatchCite{} and only 18\% on \MatchIdea{}, and no population, under any model or in any area, reaches half of it on either metric. The gap is narrowest in automated research, the newest and smallest area, and widens as the area grows, peaking in machine learning systems and code generation. LLM agents thus track a community most closely where it has least to choose from and fall furthest behind where it has most.

\item \textbf{The gap lies in what LLM agents propose, not in what they read.} For five of the six models, matching falls further behind the human level on ideas than on citations, and diversity sharpens this split. Once \textsc{topic} or \textsc{frontier} assigns a direction, four of the six models cite up to three times as diversely as the community, yet no population reaches the human level of idea diversity. A population can thus read as widely as a community and still repeat itself in what it proposes, so drawing on the literature a community builds on is not enough to arrive at the ideas it pursues.

\item \textbf{Knowing more, from training or from context, does not close the gap.} Although four of the six models may have seen the 2026 papers in training, the two trained before 2026 match more human ideas, about 26\% of the human level against 14\% as Appendix~\ref{app:leakage} details. Idea matching thus follows neither recency nor capability, as Llama 4 Scout matches the most human ideas yet receives the lowest ratings in Table~\ref{tab:human}. More context does no better. Across the four methods, matching stays within a narrow band while citation diversity varies about threefold, so what an LLM agent is told shapes how widely its population spreads far more than how closely it follows human research. The fullest view helps least, as \textsc{landscape} ranks last on \MatchCite{} in 19 of 24 settings, while the direction of \textsc{topic} and the peer proposals of \textsc{frontier} broaden a population without bringing it near the human level.

\item \textbf{The quality gap centers on execution and deepens where proposals miss the human record.} In blind review, AI proposals score below the 2026 papers on every quality dimension, by 20 points on expected success but only 4 on presentation, so the gap lies less in how the proposals are written than in whether they would work. Reviewers also flag 86\% of them as AI-generated, against 39\% of the papers. Even DeepSeek-V4-Flash, which surpasses the papers on soundness, presentation, and contribution, trails them in feasibility and expected success. Split by match, both groups of proposals trail the papers by the same 11 points in overall rating, but those with no human counterpart fall much further behind in execution, as the gap widens from 8 to 15 points in feasibility and from 15 to 26 in expected success. The proposals that miss the human record are thus the ones reviewers expect least to succeed.
\end{itemize}
\endgroup

Together, these findings show that LLM agents come closer to a community in what they read than in what they propose, that neither newer models nor richer context closes this gap, and that the proposals outside the human record read to reviewers as weaker directions rather than overlooked opportunities. Low matching thus reflects a deficit in research taste rather than a mere difference in preference. Sec.~\ref{sec:analysis} examines which part of an idea LLM agents miss and whether scale recovers it.

\paragraph{Case study.}
\label{sec:main_case}
We provide interesting cases with expert review as shown in Figure~\ref{fig:maincase}. Case~1 shows that the proposal scored higher on soundness and on expected success. Case~2 collects three documents that no counterpart matched. The first is a 2026 paper rated below both and also taken for AI.  The first proposal reads well enough for a middling overall rating, yet its feasibility and expected success sit near the bottom of the scale and the reviewer identified it as AI-generated. A reviewer took the second proposal for a published paper. Therefore, the averages in Table~\ref{tab:human} place proposals below papers there, while individual documents on both sides span the range. 

\begin{figure}[t]
\centering
\begin{promptbox}
{\scriptsize\raggedright \textbf{Case 1: a matched pair.} Problem: Both documents falsify an AI scientist's own hypotheses before accepting them, rather than trusting a frozen evaluator, and reviewers place neither side.\caserule \srchuman\textbf{\href{https://arxiv.org/abs/2603.29045}{Let the Abyss Stare Back}~\citep{li2026abyss}.} Co-evolve an innovator, an adversarial falsifier that builds counterexample environments, and a causal extractor, on a controlled loss-discovery problem.\par
\casescore{Overall 75 $\cdot$ Soundness 50 $\cdot$ Success 40 $\cdot$ Guess: \textbf{unsure}}
\smallskip
\srcai(Gemini 3.5 Flash-Lite, \textsc{free}). Add a verification agent that writes adversarial tests, probes boundary conditions, and tries to falsify each hypothesis before ratifying it.\par
\casescore{Overall 75 $\cdot$ Soundness \textbf{100} $\cdot$ Success 60 $\cdot$ Guess: \textbf{unsure}}
}
\end{promptbox}
\vspace{-10pt}
\par\medskip
\begin{promptbox}
{\scriptsize\raggedright \textbf{Case 2: examples of score heterogeneity among non-matching documents.} \caserule \srchuman\textbf{\href{https://arxiv.org/abs/2607.19321}{ResearchArena}~\citep{libon2026researcharena}.} Pair four long-horizon AI research tasks with hidden side tasks that sabotage the submitted artifact, to test whether an untrusted agent's output is safe to deploy.\par
\casescore{Overall \textbf{25} $\cdot$ Soundness 0 $\cdot$ Success 20 $\cdot$ Guess: \textbf{AI-generated}}
\smallskip
\srcai (Qwen3.8-27B, \textsc{frontier}). Key the KV cache by a graph of mutation relations among propagating artifacts rather than by request prefix, so long-horizon agent-society simulations reuse cache across agents and turns.\par
\casescore{Overall 50 $\cdot$ Feasibility \textbf{25} $\cdot$ Success \textbf{20} $\cdot$ Guess: AI-generated}
\smallskip
\srcai(Claude Haiku 4.5, \textsc{frontier}). Augment a learned latent dynamics model with a topic model, so that recurring configurations and action-effect modes surface as discrete behavioral topics.\par
\casescore{Overall \textbf{75} $\cdot$ Feasibility 75 $\cdot$ Success 40 $\cdot$ Guess: \textbf{human paper}}
}
\end{promptbox}
\caption{Cases from the blind review, with the scores and the source guess of one reviewer under each document. Case 1 pairs a proposal with the 2026 paper whose idea the judge marks as overlap, and Case 2 is three documents with no counterpart in the human record.}
\label{fig:maincase}
\end{figure}

\section{Analysis}
\label{sec:analysis}

The main results show that LLM agents fall furthest behind on idea matching. We trace where this gap lies, first across research directions, then within the idea summary and in concrete cases, and finally ask whether larger populations close it.

\begin{figure}[t]
\centering
\begin{minipage}[t]{0.43\textwidth}%
\vspace{0pt}%
{\setlength{\tabcolsep}{5pt}
\renewcommand{\arraystretch}{1.15}
\resizebox{\linewidth}{!}{%
\begin{tabular}{@{}lrrr@{}}
\toprule
Population & \# Task & \# Method & \# Pair \\
\midrule
\popmark{pophuman}Human & 200 & 268 & 2,871 \\
\popmark{popfree1136}\textsc{free} & 36 & 103 & 710 \\
\popmark{poptopic1136}\textsc{topic} & 125 & 191 & 1,734 \\
\popmark{poptag1136}\textsc{landscape} & 39 & 114 & 847 \\
\popmark{cardcoral}\textsc{frontier} & 192 & 220 & 3,441 \\
\bottomrule
\end{tabular}}}\end{minipage}%
\hfill
\begin{minipage}[t]{0.55\textwidth}%
\vspace{0pt}%
\includegraphics[width=\linewidth]{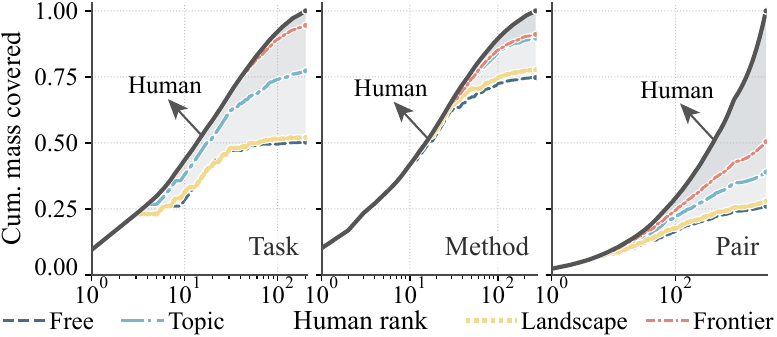}%
\end{minipage}
\caption{Tag use in automated research: the task a document works on, the method it uses, and the task--method pair. Left: how many distinct tags of each kind each population uses. Right: the cumulative share of human tag mass each population reaches, with human tags ordered from the most to the least used and the human curve as the ceiling.}
\label{fig:tagcoverage}
\end{figure}

\paragraph{Do LLM agents cover the human research directions?}
\label{sec:gap}
Low idea matching may arise because a population never takes up the directions the 2026 papers pursue, or because it takes them up and still arrives at different ideas. To tell these apart, we treat a direction as a task paired with a method, that is, what a paper works on and how, and label both sides with the task and method vocabulary that \textsc{landscape} draws on. We run this comparison for DeepSeek-V4-Flash in automated research, with each population holding as many proposals as the area has papers so that both sides are compared at equal size. Figure~\ref{fig:tagcoverage} reports the share of human tag mass each population reaches. On each axis alone, the populations given a direction come close to the community, and \textsc{frontier} misses only a small share of the human task mass and of the method mass. On pairs, every population falls away, and even \textsc{frontier} reaches only about half of the human mass, so LLM agents work on the community's tasks with its methods but combine the two differently. Neither kind of coverage, however, carries over to ideas. \textsc{Frontier} reaches about twice the human pair mass of \textsc{free} but matches barely more human ideas in Table~\ref{tab:main}, where the four methods sit within a narrow band. LLM agents thus reach the directions a community pursues and still propose something else, so what they miss lies inside a direction rather than in the choice of one.

\paragraph{Where do AI and human research ideas differ?}
\label{sec:slots}
To locate the difference inside an idea, we decompose \MatchIdea{} into the slots of the idea summary. Table~\ref{tab:slots} reports per-slot matching for Llama 4 Scout and DeepSeek-V4-Flash, one model from each side of the 2026 cutoff. Both populations reach the problems and motivations of human papers many times more often than their approaches and research plans, so LLM agents agree with a community on what to study and why far more readily than on how to do it. Part of this ordering is intrinsic, since the 2026 papers also match one another far more often on problems than on approaches, an approach being the more specific of the two. Read against this human level, the decline persists for DeepSeek-V4-Flash, which covers a small share of human problems and almost none of their research plans, while Llama 4 Scout stays at roughly a third of the human level on every slot. Either way, almost no human approach or research plan finds a counterpart among the proposals, and this is where the idea gap lies.

\begin{table}[tb]
\centering
\scriptsize
\setlength{\tabcolsep}{2.5pt}
\vspace{-10pt}
\caption{Per-slot \MatchIdea{} over the slots of the idea summary, namely problem, motivation, proposed approach, and research plan. Each entry is the share of human papers a population matches on that slot.}
\label{tab:slots}
\resizebox{\textwidth}{!}{%
\begin{tabular}{l *{16}{r}}
\toprule
& \multicolumn{4}{c}{Machine learning systems} & \multicolumn{4}{c}{Code generation} & \multicolumn{4}{c}{World models} & \multicolumn{4}{c}{Automated research} \\
\cmidrule(lr){2-5}\cmidrule(lr){6-9}\cmidrule(lr){10-13}\cmidrule(lr){14-17}
Population & Problem & Motivation & \makecell{Approach} & \makecell{Plan} & Problem & Motivation & \makecell{Approach} & \makecell{Plan} & Problem & Motivation & \makecell{Approach} & \makecell{Plan} & Problem & Motivation & \makecell{Approach} & \makecell{Plan} \\
\midrule
Human & 28.32 & 42.58 & 3.99 & 7.21 & 24.74 & 33.19 & 4.21 & 3.64 & 25.12 & 29.41 & 7.12 & 6.16 & 19.10 & 38.29 & 4.49 & 2.99 \\
\midrule
\rowcolor{lightBlue}\multicolumn{17}{l}{\textbf{2025-04-05, Llama 4 Scout}~\citep{meta2025llama4}} \\
\addlinespace[2pt]
\textsc{free} & 7.09 & 17.62 & 0.81 & 1.45 & 6.73 & 6.99 & \textbf{1.87} & 1.07 & 4.02 & 5.25 & 1.18 & 0.91 & \textbf{6.95} & 14.96 & \textbf{3.35} & 0.44 \\
\textsc{topic} & 10.00 & 20.34 & 0.87 & 0.94 & 6.13 & 10.10 & 1.63 & \textbf{2.61} & \textbf{7.02} & \textbf{10.28} & 2.09 & 1.23 & 5.28 & 14.08 & 1.58 & 0.53 \\
\textsc{landscape} & 8.54 & 18.04 & 0.87 & \textbf{2.35} & 2.93 & 4.30 & 1.13 & 1.01 & 1.71 & 2.25 & 0.37 & \textbf{1.45} & 5.99 & \textbf{15.32} & 2.38 & \textbf{0.62} \\
\textsc{frontier} & \textbf{12.10} & \textbf{23.46} & \textbf{1.29} & 1.52 & \textbf{7.50} & \textbf{10.16} & 1.63 & 0.77 & 5.52 & 8.03 & \textbf{2.25} & 1.39 & 4.23 & 13.12 & 1.41 & 0.09 \\
\midrule
\rowcolor{lightBlue}\multicolumn{17}{l}{\textbf{2026-07-31, DeepSeek-V4-Flash}~\citep{deepseekai2026v4}} \\
\addlinespace[2pt]
\textsc{free} & 1.04 & 1.12 & \textbf{0.23} & 0.08 & 0.89 & 1.21 & 0.00 & 0.00 & 2.95 & 1.98 & 0.27 & 0.00 & 2.11 & 3.43 & 0.26 & 0.00 \\
\textsc{topic} & \textbf{1.77} & \textbf{3.01} & 0.10 & 0.04 & \textbf{1.63} & 1.69 & 0.12 & 0.00 & 2.68 & 3.00 & \textbf{0.64} & \textbf{0.21} & \textbf{3.35} & \textbf{4.75} & 0.44 & \textbf{0.09} \\
\textsc{landscape} & 1.31 & 2.18 & 0.12 & \textbf{0.12} & 1.19 & 1.60 & 0.18 & 0.00 & \textbf{3.48} & \textbf{4.18} & 0.16 & 0.05 & 2.38 & 3.43 & \textbf{0.62} & \textbf{0.09} \\
\textsc{frontier} & 1.04 & 2.04 & 0.06 & 0.02 & 1.42 & \textbf{2.04} & \textbf{0.21} & 0.00 & 2.36 & 2.73 & 0.11 & 0.00 & 3.08 & 4.23 & 0.26 & 0.00 \\
\bottomrule
\end{tabular}%
}
\vspace{-10pt}
\end{table}

\paragraph{What do LLM agents propose instead?}
\label{sec:case}
Figure~\ref{fig:case} pairs each of two human papers with two proposals from one language model that share its problem. In both cases, the proposal that misses the human approach leaves the trained network unchanged and adds machinery around it, whereas the human paper and the matching proposal both modify the network itself. The Case 1 paper fine-tunes the router toward expert reuse, while proposal A predicts expert demand and prefetches experts around an unchanged network. The Case 2 paper updates low-rank adapters on a frozen network, while proposal A keeps the weights fixed and adapts through retrieved transitions. The same problem thus admits very different mechanisms, and a shared problem does not imply a shared approach. Appendix~\ref{app:case} provides the source schemas and further examples.

\begin{figure}[t]
\centering
\begin{promptbox}
{\scriptsize\raggedright \textbf{Case 1: } fetching uncached MoE experts causes substantial I/O overhead.\caserule \srchuman\textbf{\href{https://arxiv.org/abs/2605.27081}{ReMoE}~\citep{zhu2026remoe}.} Fine-tune the router to favor recently used experts, reducing transfers without adding inference-time computation.\par
\smallskip
\srcdiffer{A}(Qwen3.8-27B, \textsc{frontier}). Predict expert demand from prompt topics, then prefetch experts and coordinate caching and scheduling without changing the model.\par
\smallskip
\srcmatch{B}(Qwen3.8-27B, \textsc{topic}). Train only the router to increase expert reuse over consecutive tokens while preserving language-model performance.\par
}
\end{promptbox}
\vspace{-10pt}
\par\medskip
\begin{promptbox}
{\scriptsize\raggedright \textbf{Case 2: } pretrained world models need to adapt to changing environments while retaining prior capabilities.\caserule \srchuman\textbf{\href{https://doi.org/10.1109/LRA.2026.3688061}{SPREAD}~\citep{moon2026spread}.} Freeze the pretrained model and update low-rank adapters online, aiming for rapid adaptation without replay.\par
\smallskip
\srcdiffer{A}(DeepSeek-V4-Flash, \textsc{topic}). Keep model weights fixed, store observed transitions, and combine retrieved predictions with the model's predictions.\par
\smallskip
\srcmatch{B}(DeepSeek-V4-Flash, \textsc{free}). Adapt a frozen model through low-rank updates and compose adapters for combinations of dynamics changes.\par
}
\end{promptbox}
\caption{Two human papers, each paired with two AI proposals from a single language model. Under the stored automatic judgments, all four proposals match their human paper on problem and motivation, and only proposal B also matches on proposed approach and research plan.}
\label{fig:case}
\end{figure}

\begin{figure}[t]
\centering
\includegraphics[width=\textwidth]{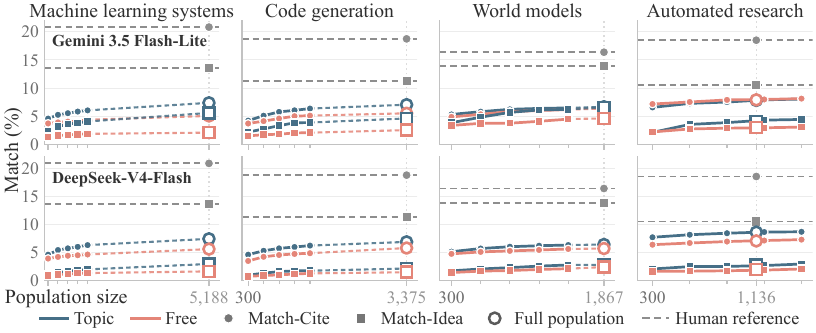}
\caption{Matching against population size. Rows are Gemini 3.5 Flash-Lite and DeepSeek-V4-Flash from top to bottom, and columns are the four areas. Color marks the method, \textsc{topic} or \textsc{free}, and marker shape the level, circles for \MatchCite{} and squares for \MatchIdea{}.}
\label{fig:scale}
\end{figure}

\paragraph{Can more LLM agents close the gap?}
\label{sec:scale}
If the gap stemmed mainly from population size, sampling more proposals should close it. We grow the \textsc{free} and \textsc{topic} populations of Gemini 3.5 Flash-Lite and DeepSeek-V4-Flash in every area from 300 proposals to the size of the human reference, 3.8 to 17 times larger, and report matching in Figure~\ref{fig:scale}. Matching grows far more slowly than the population. Every curve rises only gently and levels off as proposals accumulate, staying well below the human level even when a population holds as many proposals as the community holds papers. An assigned direction helps somewhat, as \textsc{topic} climbs more steeply on ideas than \textsc{free} in most settings, yet both remain far from the human level. The gap is thus not an artifact of population size. Together with the analyses above, this shows that neither reaching more of a community's directions nor writing more proposals brings LLM agents closer to the approaches the community arrives at. Improving research taste therefore means improving how a recognized problem becomes a concrete approach, not how widely or how much a population proposes.

\section{Conclusion}
\label{sec:conclusion}
\vspace{-10pt}
We introduced IdeaTaste, a benchmark that measures the research taste of LLM agents against the research human communities go on to publish. Our results point to a consistent picture. LLM agents share much of a community's literature and directions, yet seldom arrive at the concrete approaches it pursues, and the proposals that depart from human research tend to be weaker rather than overlooked. This picture holds across newer models, richer context, and larger populations. The research taste of LLM agents is thus limited less by knowing what to work on than by deciding how to pursue it, which we see as the central challenge for future AI scientists.

\section*{Acknowledgments}
This work was supported in part by the National Institute of Standards and Technology (NIST) (\href{https://ror.org/05xpvk416}{ror.org/05xpvk416}) and the Carnegie Mellon University (\href{https://ror.org/05x2bcf33}{ror.org/05x2bcf33}) AI Measurement Science and Engineering Center (AIMSEC). Shanshan Zhong is supported by the LTI Presidential Fellowship from School of Computer Science, Carnegie Mellon University.

\bibliography{references}
\bibliographystyle{preprint}

\clearpage
\appendix
\etocdepthtag.toc{appendix}
\etocsettagdepth{main}{none}
\etocsettagdepth{appendix}{subsection}
\etocsettocstyle{\section*{Appendix Contents}}{}
\begingroup\hypersetup{hidelinks}
\tableofcontents
\endgroup
\clearpage
\section{Data Construction Details}
\label{app:corpus}

Each area is represented by two sets of papers: a corpus $U$ of work published through 2025, which is what LLM agents retrieve from, and a human reference $H$ of papers from 2026, which is what proposals are measured against. The two are built to different standards. The corpus has to be wide enough that the background a 2026 paper builds on lies inside it, and it must carry nothing from 2026. The reference set has to be as complete as we can make it and has to hold research in the area. This appendix describes the area definitions, the sources data come from, data construction details, and the survey reference coverage audit.

\subsection{Research Areas}

Each area is defined in writing before collection begins, and that definition does not change afterwards. It states what belongs to the area and what does not, and it is what every screening step below applies. The LLM agents receive the same text as the description of their research community, so both sides work under one definition of the area. Table~\ref{tab:areadefs} gives the four definitions together with the topic search queries used to collect the reference set.

\begin{longtable}{@{}>{\raggedright\arraybackslash}p{0.2\textwidth}@{\hspace{1em}}>{\raggedright\arraybackslash}p{\dimexpr0.8\textwidth-2em\relax}@{}}
\caption{Area definitions and topic search queries recorded in each area's frozen configuration. Topic search applies to the reference set only.}\label{tab:areadefs}\\
\toprule
Area & Definition and topic search \\
\midrule
\endfirsthead
\toprule
Area & Definition and topic search \\
\midrule
\endhead
\bottomrule
\endlastfoot
Machine learning systems & Systems research for efficient training and inference of ML and LLM models: inference serving and scheduling, KV-cache and memory management, distributed training and parallelism, quantization, pruning and distillation for efficiency, speculative decoding, efficiency-motivated attention and architecture changes, ML compilers and kernels, hardware acceleration, and MoE systems. Not in scope: architecture work aimed at task quality, downstream applications that merely use these systems, retrieval-augmented generation, alignment, and generic high-performance computing unrelated to ML.\newline Topic search: the OpenAlex topics Parallel Computing and Optimization Techniques and Stochastic Gradient Optimization Techniques, and title and abstract search for ``llm serving'', ``inference serving'', and ``distributed training''. \\
\addlinespace[4pt]
Code generation & Research on automatically producing program code from natural-language or other specifications with learned models: code LLMs and pretrained code models, program synthesis, natural language to code including text-to-SQL, code benchmarks and code-specific evaluation, program repair and self-debugging, coding agents and LLM-based software engineering, and code data curation. Not in scope: generic LLM and NLP foundation work such as GPT, LLaMA, RLHF and LoRA, generic agent or reasoning work without a code focus, generic machine translation metrics, and papers that merely use code as an application domain without studying generation.\newline Topic search: OpenAlex keywords ``code-generation'' and ``program-synthesis''. \\
\addlinespace[4pt]
World models & Learned internal models of an environment that represent its current state and predict future states under actions, used for planning and decision-making, simulation, and data generation. The lineages are model-based reinforcement learning with latent dynamics, video-generative world simulators, JEPA-style predictive representation learning, and domain instances in robotics, autonomous driving and games. Not in scope: generic video generation or vision backbones not positioned as environment models, model-free reinforcement learning, and generic LLM and NLP work.\newline Topic search: title and abstract search for ``world model''. \\
\addlinespace[4pt]
Automated research & AI systems, especially LLM agents, that automate stages of the scientific research process: hypothesis and idea generation, experiment planning and execution including ML engineering and laboratory automation, analysis, scientific writing and peer review, up to end-to-end autonomous AI scientists. Research-agent benchmarks and equation or function discovery are included. Not in scope: philosophy of science without AI, generic agent frameworks without a research-automation focus, domain-science papers that merely use AI as a tool, and generic literature-retrieval tools.\newline Topic search: title and abstract search for ``ai scientist'', ``research agent'', ``automated scientific discovery'', and ``autonomous research''. \\
\end{longtable}

\subsection{Data Collection}

Collection has to satisfy one requirement: the corpus must hold the earlier work a 2026 paper builds on. We therefore run a single expansion out of hand-picked landmark papers along citation edges, and a paper's publication year decides which set it joins, the corpus through 2025 and the reference set in 2026. The reference set carries a second requirement, since it is the record every metric is read against. Research the community did that we fail to collect is invisible to the scores, and a proposal that reached it would count as reaching nothing, so the set has to be as complete as we can make it. We therefore reach it by topic search over the area as well as by citation, and screen what both paths return against the area definition.

\paragraph{Landmark papers and citation expansion.} Each area starts from 20 to 22 landmark papers chosen by hand, listed in Table~\ref{tab:landmarks}. The corpus then grows in two steps. Descendants are papers published through 2025 that cite at least one landmark paper. Ancestors are papers cited by at least two distinct descendants. The second step brings in shared background that may not use the area's terminology, and the threshold of two is what keeps it from pulling in everything a descendant happens to cite. Nothing on this side is screened for the area, since the corpus is a pool to retrieve from and a paper outside the area costs little as long as the background is there. 

\begin{longtable}{@{}r@{\hspace{1em}}>{\raggedright\arraybackslash}p{\dimexpr\textwidth-3.2em\relax}@{}}
\caption{Landmark papers for each research area. Each area's corpus grows from these by the two-step citation expansion described above. Years are those recorded in the frozen area configuration.}
\label{tab:landmarks}\\
\toprule
Year & Landmark paper \\
\midrule
\endfirsthead
\toprule
Year & Landmark paper \\
\midrule
\endhead
\bottomrule
\endlastfoot
\multicolumn{2}{@{}l}{\textbf{Machine learning systems}} \\
\addlinespace[2pt]
2018 & TVM: An Automated End-to-End Optimizing Compiler for Deep Learning \\
2019 & GPipe: Efficient Training of Giant Neural Networks using Pipeline Parallelism \\
2019 & Megatron-LM: Training Multi-Billion Parameter Language Models Using Model Parallelism \\
2019 & Triton: an intermediate language and compiler for tiled neural network computations \\
2020 & DeepSpeed: System Optimizations Enable Training Deep Learning Models with Over 100 Billion Parameters \\
2020 & ZeRO: Memory optimizations Toward Training Trillion Parameter Models \\
2021 & Switch Transformers: Scaling to Trillion Parameter Models with Simple and Efficient Sparsity \\
2022 & Alpa: Automating Inter- and Intra-Operator Parallelism for Distributed Deep Learning \\
2022 & Efficiently Scaling Transformer Inference \\
2022 & FlashAttention: Fast and Memory-Efficient Exact Attention with IO-Awareness \\
2022 & LLM.int8(): 8-bit Matrix Multiplication for Transformers at Scale \\
2022 & Orca: A Distributed Serving System for Transformer-Based Generative Models \\
2023 & Efficient Memory Management for Large Language Model Serving with PagedAttention \\
2023 & Fast Inference from Transformers via Speculative Decoding \\
2023 & FlexGen: High-Throughput Generative Inference of Large Language Models with a Single GPU \\
2023 & GPTQ: Accurate Post-Training Quantization for Generative Pre-trained Transformers \\
2023 & SmoothQuant: Accurate and Efficient Post-Training Quantization for Large Language Models \\
2024 & AWQ: Activation-aware Weight Quantization for LLM Compression and Acceleration \\
2024 & DistServe: Disaggregating Prefill and Decoding for Goodput-optimized Large Language Model Serving \\
2024 & FlashAttention-2: Faster Attention with Better Parallelism and Work Partitioning \\
2024 & Mooncake: A KVCache-centric Disaggregated Architecture for LLM Serving \\
2024 & SGLang: Efficient Execution of Structured Language Model Programs \\
\addlinespace[3pt]
\addlinespace[3pt]
\multicolumn{2}{@{}l}{\textbf{Code generation}} \\
\addlinespace[2pt]
2018 & Spider: A Large-Scale Human-Labeled Dataset for Complex and Cross-Domain Semantic Parsing and Text-to-SQL Task \\
2020 & CodeBERT: A Pre-Trained Model for Programming and Natural Languages \\
2021 & CodeT5: Identifier-aware Unified Pre-trained Encoder-Decoder Models for Code Understanding and Generation \\
2021 & CodeXGLUE: A Machine Learning Benchmark Dataset for Code Understanding and Generation \\
2021 & Evaluating Large Language Models Trained on Code \\
2021 & Measuring Coding Challenge Competence With APPS \\
2021 & Program Synthesis with Large Language Models \\
2022 & Competition-Level Code Generation with AlphaCode \\
2022 & InCoder: A Generative Model for Code Infilling and Synthesis \\
2023 & Code Llama: Open Foundation Models for Code \\
2023 & CodeGen: An Open Large Language Model for Code with Multi-Turn Program Synthesis \\
2023 & MetaGPT: Meta Programming for A Multi-Agent Collaborative Framework \\
2023 & Reflexion: Language Agents with Verbal Reinforcement Learning \\
2023 & StarCoder: may the source be with you! \\
2023 & Teaching Large Language Models to Self-Debug \\
2023 & WizardCoder: Empowering Code Large Language Models with Evol-Instruct \\
2024 & BigCodeBench: Benchmarking Code Generation with Diverse Function Calls and Complex Instructions \\
2024 & DeepSeek-Coder: When the Large Language Model Meets Programming \\
2024 & Magicoder: Empowering Code Generation with OSS-Instruct \\
2024 & Qwen2.5-Coder Technical Report \\
2024 & SWE-agent: Agent-Computer Interfaces Enable Automated Software Engineering \\
2024 & SWE-bench: Can Language Models Resolve Real-World GitHub Issues \\
\addlinespace[3pt]
\addlinespace[3pt]
\multicolumn{2}{@{}l}{\textbf{World models}} \\
\addlinespace[2pt]
2018 & Recurrent World Models Facilitate Policy Evolution \\
2018 & World Models \\
2019 & Learning Latent Dynamics for Planning from Pixels \\
2020 & Dream to Control: Learning Behaviors by Latent Imagination \\
2020 & Mastering Atari, Go, chess and shogi by planning with a learned model \\
2021 & Mastering Atari with Discrete World Models \\
2022 & A Path Towards Autonomous Machine Intelligence \\
2022 & DayDreamer: World Models for Physical Robot Learning \\
2023 & DriveDreamer: Towards Real-world-driven World Models for Autonomous Driving \\
2023 & GAIA-1: A Generative World Model for Autonomous Driving \\
2023 & Mastering Diverse Domains through World Models \\
2023 & Self-Supervised Learning from Images with a Joint-Embedding Predictive Architecture \\
2023 & Transformers are Sample-Efficient World Models \\
2024 & Diffusion for World Modeling: Visual Details Matter in Atari \\
2024 & Genie: Generative Interactive Environments \\
2024 & Learning Interactive Real-World Simulators \\
2024 & Revisiting Feature Prediction for Learning Visual Representations from Video \\
2024 & TD-MPC2: Scalable, Robust World Models for Continuous Control \\
2024 & Video generation models as world simulators \\
2024 & Vista: A Generalizable Driving World Model with High Fidelity and Versatile Controllability \\
2025 & Cosmos World Foundation Model Platform for Physical AI \\
\addlinespace[3pt]
\addlinespace[3pt]
\multicolumn{2}{@{}l}{\textbf{Automated research}} \\
\addlinespace[2pt]
2020 & AI Feynman: A physics-inspired method for symbolic regression \\
2021 & Nobel Turing Challenge: creating the engine for scientific discovery \\
2023 & Autonomous chemical research with large language models \\
2023 & Emergent autonomous scientific research capabilities of large language models \\
2023 & PaperQA: Retrieval-Augmented Generative Agent for Scientific Research \\
2024 & Can LLMs Generate Novel Research Ideas: A Large-Scale Human Study with NLP Researchers \\
2024 & ChemCrow: Augmenting large language models with chemistry tools \\
2024 & Large Language Models for Automated Open-domain Scientific Hypotheses Discovery \\
2024 & MLAgentBench: Evaluating Language Agents on Machine Learning Experimentation \\
2024 & MLE-bench: Evaluating Machine Learning Agents on Machine Learning Engineering \\
2024 & MLR-Copilot: Autonomous Machine Learning Research based on Large Language Models Agents \\
2024 & Mathematical discoveries from program search with large language models \\
2024 & ResearchAgent: Iterative Research Idea Generation over Scientific Literature with Large Language Models \\
2024 & SciMON: Scientific Inspiration Machines Optimized for Novelty \\
2024 & The AI Scientist: Towards Fully Automated Open-Ended Scientific Discovery \\
2025 & Agent Laboratory: Using LLM Agents as Research Assistants \\
2025 & CycleResearcher: Improving Automated Research via Automated Review \\
2025 & SciAgents: Automating Scientific Discovery Through Bioinspired Multi-Agent Intelligent Graph Reasoning \\
2025 & The AI Scientist-v2: Workshop-Level Automated Scientific Discovery via Agentic Tree Search \\
2025 & Towards an AI co-scientist \\
\addlinespace[3pt]
\end{longtable}
\suppressfloats[t]

Table~\ref{tab:corpusyears} gives what the two steps collect, by publication year. 2025 sits below 2024 in every area. That gap belongs to the indexes. A paper may enter an index well after it first appears \citep{haustein2015published}, and even once it is there, an index carries a parsed reference list for only part of what it holds \citep{culbert2025openalex}, so the most recent year is still filling when we take the snapshot. 

\begin{table}[t]
\centering
\small
\setlength{\tabcolsep}{4pt}
\caption{The collection by publication year. }
\label{tab:corpusyears}
\resizebox{\textwidth}{!}{%
\begin{tabular}{l rrrrrrrr r}
\toprule
& \multicolumn{8}{c}{Corpus ($\le 2025$)} & Reference \\
\cmidrule(lr){2-9}\cmidrule(lr){10-10}
Area & Total & $\le$2009 & 2010--14 & 2015--19 & 2020--21 & 2022--23 & 2024 & 2025 & 2026 \\
\midrule
Machine learning systems & 88{,}184 & 5{,}626 & 3{,}646 & 16{,}572 & 14{,}393 & 19{,}383 & 14{,}868 & 13{,}696 & 5{,}188 \\
Code generation          & 94{,}674 & 6{,}779 & 4{,}282 & 14{,}232 & 12{,}285 & 20{,}524 & 18{,}288 & 18{,}284 & 3{,}375 \\
World models             & 50{,}271 & 5{,}541 & 3{,}116 & 11{,}245 & 8{,}758 & 10{,}591 & 6{,}004 & 5{,}016 & 1{,}867 \\
Automated research       & 30{,}264 & 3{,}134 & 1{,}505 & 4{,}886 & 4{,}213 & 7{,}321 & 5{,}362 & 3{,}843 & 1{,}136 \\
\bottomrule
\end{tabular}%
}
\end{table}

\begin{table}[t]
\centering
\small
\setlength{\tabcolsep}{4pt}
\caption{The reference set by month of publication. The snapshot is taken at the end of August, so that month is only partly filled.}
\label{tab:refmonths}
\begin{tabular}{l r rrrrrrrr}
\toprule
& & \multicolumn{8}{c}{Month of 2026} \\
\cmidrule(lr){3-10}
Area & Total & 01 & 02 & 03 & 04 & 05 & 06 & 07 & 08 \\
\midrule
Machine learning systems & 5{,}188 & 825 & 512 & 539 & 694 & 953 & 703 & 542 & 420 \\
Code generation          & 3{,}375 & 540 & 306 & 346 & 493 & 558 & 441 & 464 & 227 \\
World models             & 1{,}867 & 203 & 173 & 223 & 167 & 312 & 342 & 271 & 176 \\
Automated research       & 1{,}136 & 155 & 86 & 139 & 127 & 190 & 173 & 151 & 115 \\
\bottomrule
\end{tabular}
\end{table} 

\begin{figure}[t]
\centering
\begin{promptbox}
{\scriptsize\ttfamily\raggedright \textbf{System}\par\smallskip
You are a strict topic classifier for a research-paper corpus. Decide whether the paper below belongs to the research area defined here.\par
Area definition:\par
\slot{the area's definition, as given in Table~\ref{tab:areadefs}}\par
\par
Judge by the paper's main contribution as described in its title and abstract; a paper that only mentions the area in passing or uses its tools for another goal does NOT belong. If no abstract is given, judge from the title alone and lower your confidence.\par
Reply with JSON only: \{"relevant": true or false, "confidence": <0-1>, "reason": "<one short sentence>"\}\par
\medskip
\textbf{User}\par\smallskip
Title: \slot{title}\par
Abstract: \slot{abstract}\par
}
\end{promptbox}
\caption{The area screening prompt. One call decides one paper.}
\label{fig:screen_prompt}
\end{figure}

\begin{table}[t]
\centering
\small
\caption{Record completeness after gap-filling. ``Abstract" means the percentage of papers with an abstract. ``References" means the percentage of papers with a reference list. Every reference paper has an abstract and almost all of their reference lists resolve into the area's corpus.}
\label{tab:completeness}
\begin{tabular}{l rcc rcc}
\toprule
& \multicolumn{3}{c}{Corpus ($\le 2025$)} & \multicolumn{3}{c}{Reference (2026)} \\
\cmidrule(lr){2-4}\cmidrule(lr){5-7}
Area & Papers & Abstract & References & Papers & Abstract & References \\
\midrule
Machine learning systems & 88{,}184 & 96.9 & 65.2 & 5{,}188 & 100 & 96.2 \\
Code generation          & 94{,}674 & 96.5 & 62.5 & 3{,}375 & 100 & 97.6 \\
World models             & 50{,}271 & 96.0 & 70.5 & 1{,}867 & 100 & 91.4 \\
Automated research       & 30{,}264 & 95.3 & 71.8 & 1{,}136 & 100 & 93.7 \\
\midrule
All                      & 263{,}393 & 96.4 & 66.0 & 11{,}566 & 100 & 95.6 \\
\bottomrule
\end{tabular}
\end{table}

\paragraph{Reference set construction.} The reference set is built in three stages.

\begin{itemize}[leftmargin=0.3cm, itemsep=-0.3em, topsep=-0.2em]
\item \textbf{Collection paths.} Papers published in 2026 that cite a landmark continue the citation expansion directly, but many of them only use a released model or system rather than doing research in the area. Topic search over the area reaches what that path misses, including work that cites no landmark at all, so we collect along both.
\item \textbf{Screening.} Both paths are screened against the area definition, one DeepSeek-V4-Flash call per paper in JSON mode with the prompt and the rule it applies given in Figure~\ref{fig:screen_prompt}. The model returns a decision, a confidence, and a one-sentence reason, and each paper is judged once. The reference set is the union of what survives.
\item \textbf{Filtering.} We then remove surveys, duplicate records, and records that also appear in the corpus (using title and author list to filtering), which happens when a preprint and its published version are indexed separately.
\end{itemize}

Collection covers January to August 2026 as shown in Table~\ref{tab:refmonths}, with the snapshot taken at the end of August, so the final month is not filled.

\paragraph{Sources and paper records.} We collect one record per paper, carrying its title and abstract, its publication date, its reference list, and the papers that cite it. Records come from OpenAlex \citep{priem2022openalex} and Semantic Scholar \citep{kinney2023s2}. Records from the two sources are merged by identifier (DOI, arXiv id, etc.). Each index leaves reference lists unparsed for part of the corpus, and the edges they do report overlap only in part, so we merge the two and treat the union as a paper's reference list. Publication dates missing from both are recovered from the arXiv API for preprints and from Crossref otherwise. Abstracts that neither index carried were recovered from the Springer, Elsevier and IEEE metadata services, from the first page of open-access PDFs, and by hand from a publisher's page. Reference lists for 2026 papers that no index had parsed were recovered from full text where an index held it and otherwise from the PDF itself. After this pass every paper in the reference set has an abstract and almost all carry reference lists that resolve into the corpus for their area, while in the corpus nearly all records have abstracts and about two thirds carry reference lists as shown in Table~\ref{tab:completeness}.

\subsection{Coverage of Survey References}

References from authoritative surveys provide an external check on how much of an area's literature the collection holds. The audit screens each survey's references for the area and measures how many of them that area's own collection holds, its corpus together with its reference set. A reference we collected under another area counts as missing, since an LLM agent working in this area never retrieves it. Table~\ref{tab:surveys} lists the surveys behind the audit. The references are screened one at a time by a language model with the same prompt shown in Figure~\ref{fig:screen_prompt}, which decides from the title since a reference record carries no abstract. 

\begin{table}[t]
\centering
\small
\setlength{\tabcolsep}{4pt}
\caption{The surveys behind the coverage audit. References is the length of a survey's reference list, read from Semantic Scholar, or parsed from the survey's arXiv page for those that Semantic Scholar returns no references for. In area is what remains after screening those references for the area. The area rows count such a work once, which is why they are smaller than the sum of the surveys below them, and they match the References column of Table~\ref{tab:coverage}.}
\label{tab:surveys}
\begin{tabular}{@{}>{\raggedright\arraybackslash}p{0.60\textwidth} l r r@{}}
\toprule
Survey & arXiv & References & In area \\
\midrule
\rowcolor{lightBlue}\multicolumn{2}{@{}l}{\textbf{Machine learning systems}} & \textbf{761} & \textbf{409} \\
\addlinespace[2pt]
Towards Efficient Generative LLM Serving & 2312.15234 & 334 & 240 \\
A Survey on Efficient Inference for LLMs & 2404.14294 & 304 & 202 \\
LLM Inference Serving: Recent Advances & 2407.12391 & 63 & 54 \\
A Comprehensive Survey of Speculative Decoding & 2401.07851 & 60 & 31 \\
\addlinespace[3pt]
\rowcolor{lightBlue}\multicolumn{2}{@{}l}{\textbf{Code generation}} & \textbf{1{,}482} & \textbf{718} \\
\addlinespace[2pt]
A Survey on Large Language Models for Code Generation & 2406.00515 & 289 & 149 \\
Unifying the Perspectives of NLP and Software Engineering & 2311.07989 & 902 & 561 \\
LLMs for Code Generation: Challenges, Techniques, Evaluation & 2503.01245 & 60 & 38 \\
AI Agentic Programming: A Survey & 2508.11126 & 231 & 78 \\
\addlinespace[3pt]
\rowcolor{lightBlue}\multicolumn{2}{@{}l}{\textbf{World models}} & \textbf{1{,}348} & \textbf{435} \\
\addlinespace[2pt]
Understanding World or Predicting Future? A Comprehensive Survey of World Models & 2411.14499 & 340 & 109 \\
Is Sora a World Simulator? & 2405.03520 & 255 & 63 \\
World Models for Autonomous Driving: An Initial Survey & 2403.02622 & 116 & 57 \\
World Models: Architectures, Methodologies, Reasoning Paradigms, and Applications & 2606.00133 & 430 & 202 \\
World Model for Robot Learning & 2605.00080 & 207 & 113 \\
\addlinespace[3pt]
\rowcolor{lightBlue}\multicolumn{2}{@{}l}{\textbf{Automated research}} & \textbf{814} & \textbf{457} \\
\addlinespace[2pt]
From Automation to Autonomy: LLMs in Scientific Discovery & 2505.13259 & 117 & 80 \\
LLM4SR: LLMs for Scientific Research & 2501.04306 & 198 & 120 \\
Autonomous Research Agents and the Verification Gap & 2608.05179 & 499 & 315 \\
\bottomrule
\end{tabular}
\end{table}

Table~\ref{tab:coverage} reports the result. For code generation the table also audits a corpus built by filtering OpenAlex~\citep{priem2022openalex} for the code generation keyword instead of expanding along citations, which reaches only a small fraction of the same references. That gap is the argument for expanding along citations: much of what an area builds on does not call itself by the area's name. Automated research has the lowest coverage of the four, and its audit is the one that reaches furthest into 2026. It carries the most references to 2026 work of any area, most of them from a 2026 survey~\citep{ding2026autonomous}. A paper enters the collection either because something already inside it cites the paper or because an index files the paper under the area, and a preprint posted weeks before the snapshot satisfies neither.

\begin{table}[t]
\centering
\small
\caption{Survey-reference coverage. References is the number of survey references the screen places in the area, which Table~\ref{tab:surveys} breaks down by survey, and Covered is matched by identifier (DOI or arXiv ID) or by title. }
\label{tab:coverage}
\begin{tabular}{llrr}
\toprule
Area & Construction & References & Covered \\
\midrule
Machine learning systems & Citation expansion & 409 & 88.3\% \\
Code generation          & Citation expansion & 718 & 92.9\% \\
World models             & Citation expansion & 435 & 94.0\% \\
Automated research       & Citation expansion & 457 & 79.0\% \\
\addlinespace[2pt]
All                      & Citation expansion & 2{,}019 & 89.1\% \\
\midrule
Code generation          & Keyword search     & 718 & 17.8\% \\
\bottomrule
\end{tabular}
\end{table}

\subsection{Interpretation of the Human Reference}
\label{app:reference}

The reference we score against is the set of 2026 papers in each area. By 2026 some of these papers are written with model assistance, and many are preprints that no venue has reviewed, so the set is not peer-reviewed, purely human work. This does not weaken its role in the benchmark as a measure of human research activity. Every paper in the reference was released by its authors. Releasing a paper is a human decision: the authors judged the problem worth pursuing and the work worth putting on the record. That judgment is what we compare against, and it stands whether or not a model helped write the text and whether or not a venue has reviewed it. What we compare against is research taste, i.e., the problems and ideas the authors chose. Whether a 2026 paper is strong or weak therefore does not change the measurement, since a weak paper still marks a direction a human researcher chose to pursue. What the benchmark asks is whether an LLM agent population proposes the problems and ideas that human researchers pursued, that is, whether it can reach ideas at the human level.

\subsection{Interpretation of research taste.}
IdeaTaste uses research taste in a descriptive sense. A research community faces many plausible problems and methods but chooses to pursue only a small subset. We therefore treat papers released after the cutoff as observations of these realized research choices. The human reference captures which prior work researchers build on, which problems they pursue, and which approaches they develop. It does not assume that every published paper is high quality or that publication defines scientific value. Accordingly, MATCH-CITE and MATCH-IDEA measure how closely an LLM agent population anticipates the subsequent choices of the corresponding human community. High matching does not by itself imply higher research quality, and low matching does not by itself imply lower quality. The benchmark measures agreement with realized community choices rather than a normative standard of scientific merit. The blinded expert study helps interpret the gap beyond matching alone. A non-matching proposal could represent a valuable direction that the human community did not pursue. However, non-matching AI proposals are judged less feasible and less likely to succeed than non-matching human papers. This suggests that the observed gap is not explained only by different research preferences. We therefore view IdeaTaste as measuring one operational aspect of research taste: whether LLM agents anticipate the research choices that a human community later makes.

\subsection{Interpreting Corpus Size}

Corpus ($\le 2025$) size follows from how the citation expansion runs rather than from how large an area is. Code generation ends with the largest corpus of the four and a smaller reference set than machine learning systems, and both follow from the same cause. Several code generation landmarks are cited across natural language processing, semantic parsing, and software engineering, so their descendants reach well beyond the area, while the machine learning systems landmarks are cited mostly inside the systems community. Ancestors are drawn from those descendants, so the difference compounds at the second step. For the reference set the same breadth works the other way, since screening for the area removes the general natural language processing papers that cite those landmarks. A corpus size therefore records how widely an area's research is built on elsewhere, not how much work the area itself produces.

\subsection{Remaining Collection Gaps}
\label{app:missing}

Three gaps remain that no source we tried closes. 
\begin{itemize}[leftmargin=0.3cm, itemsep=-0.3em, topsep=-0.2em]
\item The corpus records still without an abstract are older closed-access work that none of the indexes or publisher services carries. 
\item Every index also lags on recent preprints, which thins the most recent layer of the corpus and leaves a small share of reference papers without a parsed reference list; those papers stay in the reference set and are left out of the denominator when we report how citations resolve into the corpus.
\item A preprint that nothing has cited yet may fall outside both collection paths, since a paper is reached either by a citation from inside the collection or by an index filing it under the area. 
\end{itemize}

\section{Evaluation Details}
\label{app:eval}

We give here the details of the evaluation, including how the citation threshold $\tau$ is set, how the idea schema is extracted from abstracts, and how human evaluations assess the idea judge and idea quality.

\subsection{Citation Sufficiency Threshold}
\label{app:tau}

\DivCite{} needs a point at which a proposal counts as having read enough, and that point is $\tau$. It is taken from the literature.

\paragraph{Definition.} $\tau$ is the mean of $|\mathcal{C}(h)|$ over an area's human reference, computed after citations are restricted to the corpus $U$, so the threshold is measured on the same sets the metric scores. One value serves an area. We leave $\tau$ unrounded: citation counts are integers.

\paragraph{Interpretation.} Table~\ref{tab:tau} gives the four thresholds with the human reference split into its two factors. The human papers do not reach the ceiling on $Q$, because $\tau$ is the mean of a right-skewed distribution and fewer than half of them cite that many papers. The human row of \DivCite{} is therefore a level a population can pass and a population that passes it reads more per proposal or spreads wider than the community does.

\begin{table}[t]
\centering
\small
\caption{The citation threshold $\tau$ for each area, with the human reference's \DivCite{} split into sufficiency $Q$ and spread $S$. The last three columns are percentages; $\tau$ is a number of citations.}
\label{tab:tau}
\begin{tabular}{lrrrr}
\toprule
Area & $\tau$ & $Q_H$ & $S_H$ & $Q_H S_H$ (\DivCite{}) \\
\midrule
Machine learning systems & 20.81 & 70.68 & 19.10 & 13.50 \\
Code generation          & 20.94 & 73.58 & 25.33 & 18.64 \\
World models             & 26.46 & 68.05 & 21.57 & 14.68 \\
Automated research       & 18.28 & 73.51 & 29.51 & 21.69 \\
\bottomrule
\end{tabular}
\end{table}

\subsection{Idea Schema and Extraction}

A language model reads each abstract and fills a fixed schema. The schema records the proposed work and has five slots: contribution type, problem, motivation, proposed approach, and research plan. A slot whose content the abstract does not state is recorded as "None". The language model we use for schema extraction is DeepSeek-V4-Pro. We use a strong model as we want to maximize the accuracy of the extracted schemas. The prompt is given in Figure~\ref{fig:schema_prompt}. 

\begin{promptboxlong}
{\scriptsize\ttfamily\raggedright \textbf{System}\par\smallskip
You are a research schema extractor. Given a paper abstract, extract that paper's idea schema as a single JSON object.\par
\par\smallskip
FOUR RULES THAT OVERRIDE EVERYTHING BELOW\par
\par\smallskip
R1. Faithfulness outranks the rubric. The field descriptions below describe what a well-posed research idea looks like; they are NOT a standard the paper must be edited to meet. Extract what this abstract actually says. Where the abstract falls short of a description --- an evaluation with no stated failure condition --- record it as it is and note the shortfall in quality\_note. Never repair, sharpen, or complete the paper. An invented scope, baseline, metric, or failure condition is a far worse error than a thin field.\par
R2. Missing means missing. If the abstract supplies nothing for a schema field, that field's entire value is the string "None", and for a list field the single element ["None"]. Use that exact word --- not a sentence explaining what is missing, not JSON null, not an empty string --- and never back-fill the field from another one that does have content. Three exemptions: `contribution\_type.primary` is always one of the listed types (it is your classification, not something the abstract has to supply); `contribution\_type.secondary` is [] when there is no genuine second contribution; and the three quality\_check fields follow their own instructions below.\par
R3. A defective abstract does not lower the bar and does not raise it. If the text is truncated, a stub, or otherwise not a real abstract, extract whatever it genuinely supports and set every other field to "None" --- do not compensate by inferring from the title, the venue, or what such papers usually do. Report the defect through abstract\_complete and quality\_note. The schema of a defective abstract is expected to be mostly not-stated; that is the correct output, not a failure.\par
R4. Two bans apply to every field.\par
  (a) Anonymity. Do NOT copy any name this paper coins for its own contribution --- not the system, tool, framework, benchmark or dataset it proposes, and equally not a metric, principle, concept or phenomenon it names, in full form or as an acronym. 'We propose FooBar' becomes 'the proposed framework'; 'introducing Near-Free Parallelism (NFP), the maximum number of positions executable at near-free latency' becomes 'the maximum number of positions executable at near-free latency', with every later NFP replaced the same way; 'we introduce PASS@(k,T), a two-dimensional metric' becomes 'a two-dimensional metric'. A coined name is identifying even when it reads like ordinary technical vocabulary --- if the abstract is the thing that introduces the term, describe the concept instead of naming it. Also no title and no author names. Names of PRIOR work, baselines, and third-party models, datasets and libraries are NOT covered by this ban: keep them wherever the abstract gives them.\par
  (b) No outcomes. Do NOT copy results or performance numbers (accuracies, speedups, improvement ratios, win rates, acceptance counts) into any field: give the claims, the approach, and the evaluation design, not how they turned out. Quantities that define the setting rather than report a result --- model scale, token budget, dataset size, number of released tasks --- are allowed, and belong in a claim's scope or in the approach.\par
\par\smallskip
OUTPUT\par
\par\smallskip
English only, whatever the abstract's language. Budgets, which are caps and not targets:\par
  problem, motivation --- 1-2 sentences each.\par
  proposed\_approach --- 2-4 sentences.\par
  research\_plan --- up to 6 sentences (it usually needs them; the others usually do not).\par
  quality\_note --- 1 sentence.\par
When a field's content does not fit its budget, drop by this order rather than compressing everything into an unreadable list: in research\_plan, keep which evidence bears on which claim, then the failure conditions and what is held matched, and drop the enumeration of every baseline and metric name first; in proposed\_approach, keep the mechanism and drop release/scale detail first. Never drop a field to "None" to save room --- that value means the abstract said nothing, and using it for anything else corrupts the downstream not-stated statistics.\par
\par\smallskip
SCHEMA FIELDS\par
\par\smallskip
contribution\_type --- The form of the contribution, exactly one of these five in `primary`:\par
  theory --- establishes what is or is not possible, or gives a formalism: proofs, bounds, characterizations, and the framings and problem formulations they rest on.\par
  method --- a new or modified technique meant to do a task better or newly at all.\par
  empirical\_analysis --- measures or explains behaviour of things that already exist; the contribution is the finding, not a technique.\par
  artifact --- a dataset, benchmark, system, library, or model release; the contribution is the thing itself and its fitness for use.\par
  position --- argues that the field should frame, evaluate, or prioritize something differently; the contribution is the argument.\par
Put another type in `secondary` only if the work genuinely delivers it too, else []. This is a contribution axis, not a domain --- a paper applying diffusion models to protein design is still `method`. Classify by which kind of evidence the work must bring, not by what it ships: a benchmark paper whose point is what the benchmark reveals about existing models is `empirical\_analysis` with `artifact` secondary. When the abstract is too thin to classify confidently, choose the closest type anyway and say in quality\_note that the classification is uncertain.\par
\par\smallskip
problem --- What is currently not known, not achievable, or not correctly formulated. Descriptive, and stated without naming the paper's solution: someone who has not heard the idea should still recognize it as a gap. For `position` work the gap may be a defect in how the field frames or evaluates something rather than an open technical question.\par
\par\smallskip
motivation --- Why this is worth solving now: what becomes possible, what cost is avoided, or what newly available capability makes it tractable. Do not restate the gap.\par
\par\smallskip
proposed\_approach --- What is built and how: method design, proof strategy, construction protocol, or system architecture, including what is released and at what scale. Execution only --- not the assertion, not the evidence it worked.\par
\par\smallskip
research\_plan --- The evidence the paper brings to bear on its claims, and what would count as support or as failure. Number the claims when they get different evidence ('(1) ... (2) ...'). Evidence means proofs and assumption-tightness arguments, experiments with named baselines and metrics, ablations isolating the component a claim credits, robustness checks, or quality audits; which of these counts depends on contribution\_type (a `theory` paper's evidence is its proof, an `artifact` paper's is its quality audit and coverage analysis). Give a failure condition only where the abstract implies one, and say what is held matched (compute, parameters, latency) for trade-off claims when the abstract says. Describe the design, never the outcome.\par
\par\smallskip
QUALITY CHECK --- about the abstract as a text, not about the paper. These three fields are exempt from R2: they always carry a real value.\par
\par\smallskip
abstract\_complete --- boolean. true if this is a real, untruncated abstract; false if it is a stub, a citation placeholder, a fragment, or otherwise not an abstract.\par
abstract\_sufficient --- boolean. true if it carries enough to fill the schema without guessing.\par
quality\_note --- one sentence naming which fields you left as "None", whether the contribution\_type classification is uncertain, and where the abstract fell short of a field's description. If none of those apply, the value is exactly "None".\par
\par\smallskip
Respond with a single JSON object only.\par
\medskip
\textbf{User}\par\smallskip
Abstract:\par
\slot{abstract}\par
\par\smallskip
Return JSON with exactly this shape:\par
\{"schema": \{"contribution\_type": \{"primary": "method", "secondary": []\}, "problem": "...", "motivation": "...", "proposed\_approach": "...", "research\_plan": "..."\}, "quality\_check": \{"abstract\_complete": true/false, "abstract\_sufficient": true/false, "quality\_note": "..."\}\}\par
`primary` must be one of: theory, method, empirical\_analysis, artifact, position --- never "None"; `secondary` is [] when there is no genuine second contribution. Any other schema field the abstract supplies nothing for is exactly "None".\par
}
\end{promptboxlong}
\nopagebreak
\captionof{figure}{The schema extraction prompt, run with the same model on all research documents. The rules hold a summary to what the abstract states: a name the paper coins for its own work becomes a description, and reported outcomes are dropped, so two documents are compared on their research idea rather than on their wording or their results. }
\label{fig:schema_prompt}

\begin{figure}[t]
\centering
\begin{promptbox}
{\scriptsize\ttfamily\raggedright \textbf{System}\par\smallskip
You judge whether two research idea schemas describe the SAME core idea. Each schema has these 5 fields: contribution\_type, problem, motivation, proposed\_approach, research\_plan. Judge in two stages and report both.\par
STAGE 1 (per field): for each field in turn, decide whether A and B are substantially the same ON THAT FIELD ALONE, as if the other fields were hidden from you.\par
If a field has no corresponding content in A or B (including when both are missing or use the placeholder 'None'), its per-field verdict must be false.\par
STAGE 2 (whole idea): same\_idea is true only if the two schemas address substantially the same problem AND propose substantially the same approach. This is your own judgement on the full schema, NOT a vote or an average over stage 1 --- agreeing on problem, motivation, contribution\_type alone is not enough.\par
In both stages: different wording, naming, or level of detail does not matter; merely sharing a topic, task, dataset, or a general technique family does NOT count as the same.\par
Reply JSON only: \{"fields": \{"contribution\_type": true/false, "problem": true/false, "motivation": true/false, "proposed\_approach": true/false, "research\_plan": true/false\}, "same\_idea": true/false, "reason": "<one sentence>"\}.\par
\medskip
\textbf{User}\par\smallskip
SCHEMA A:\par
\slot{the first summary, one line per slot}\par
\par\smallskip
SCHEMA B:\par
\slot{the second summary, in the same form}\par
}
\end{promptbox}
\caption{The judging prompt for schemas for two research documents. Part one decides each slot on its own, so that a pair can agree on the problem and differ on the approach; a slot that either side leaves empty counts as a difference. Part two is the judge's own reading of the whole idea rather than a tally of part one. }
\label{fig:judge_prompt}
\end{figure}

\subsection{Candidate Retrieval, Idea Matching, and Slot Weights}

\paragraph{Retrieval.} Retrieval runs over the title and abstract of each research document and combines a sparse and a dense channel. The sparse channel is BM25 with $k_1=1.5$ and $b=0.75$. The dense channel encodes the same text with all-MiniLM-L6-v2 \citep{reimers2019sbert} and ranks by cosine similarity. The two rankings are fused over the top 200 of each channel, where a document placed $r$th in a channel contributes $1/(60+r)$, and the five highest fused documents become $\mathcal{N}_X(d)$. 

\paragraph{Pairwise idea matching.} Each pair is judged once. The two summaries are presented in a random order, which removes a systematic position preference. The judge rules on every slot separately and also returns an overall verdict on whether the two documents state the same idea. The prompt is given in Figure~\ref{fig:judge_prompt}. The judge is a language model, GPT-5.6 Luna, which is strong enough to read the summaries and apply the rules.

\paragraph{Slot weights.} Weights come from the human reference for each research area. For each human 2026 paper we ask whether any of the other human 2026 papers retrieved for it matches on slot $i$; $p_i$ is the fraction of human papers with such a match, and $w_i=-\log_2 p_i$. A slot that two papers share by default therefore carries almost no weight, while a slot that rarely matches carries more. The weights and their sum $W$ are estimated once per area and held fixed across methods.

\subsection{Human Evaluation of the Idea Judge}
\label{app:judge}

The idea metrics rest on one decision, whether two summaries agree on a given slot. In this section, we sample pairs the judge has already ruled on and have people make the same decision without seeing its answer.

\begin{figure}[t]
\centering
\includegraphics[width=\textwidth]{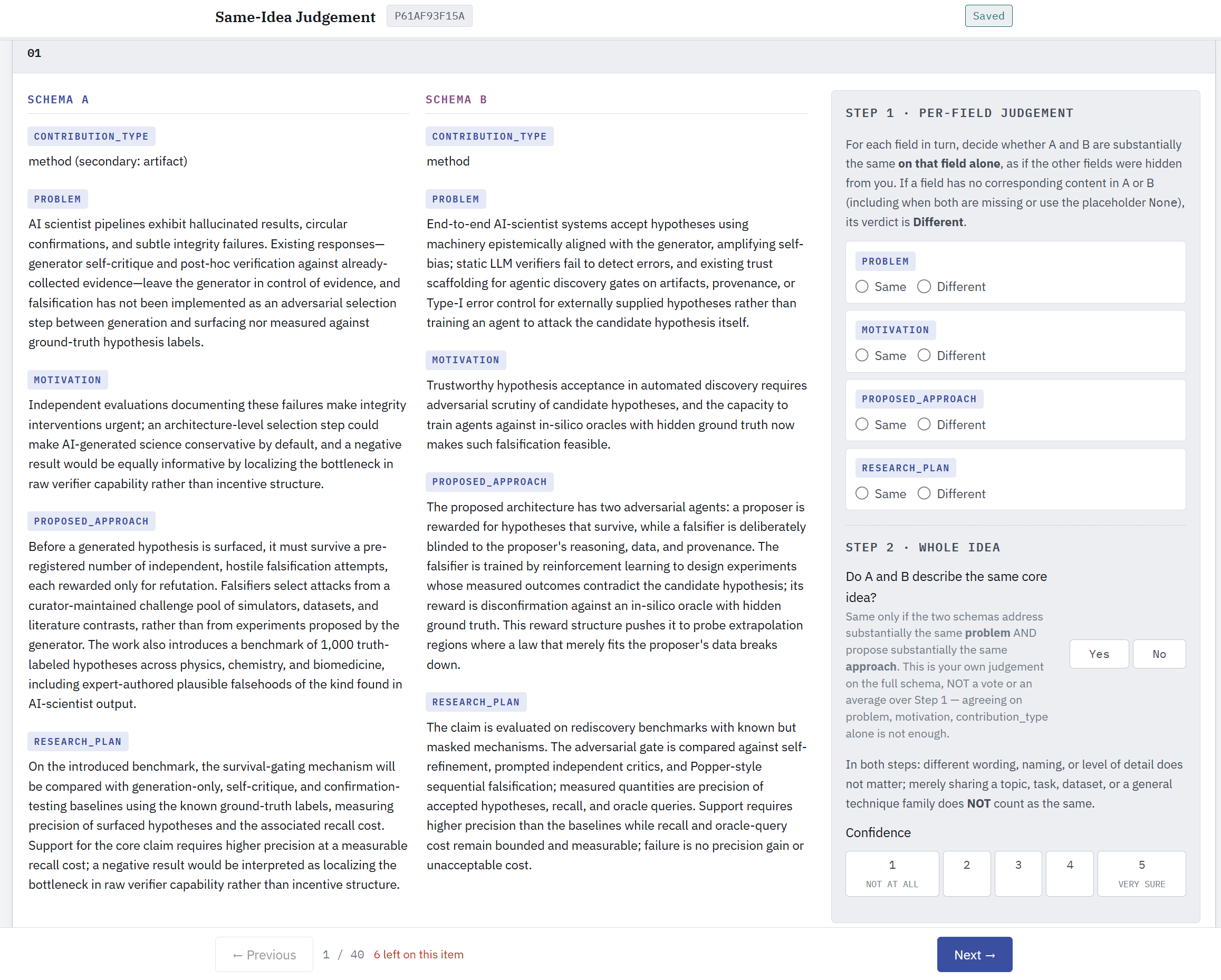}
\caption{The same-idea judgement interface, showing one example. The left panes give the two idea summaries; the right pane states the criterion the judge is given and collects the four slot decisions, the whole-idea decision, and confidence.}
\label{fig:judge-ui}
\end{figure}

\paragraph{Questionnaire.} Figure~\ref{fig:judge-ui} shows one example. The two left panes give the two idea summaries, the five slots contribution type, problem, motivation, proposed approach, and research plan, so an annotator decides on the same material the judge reads. The right pane carries the judge's own instruction, narrowed to the four slots asked, and collects the answers in the two stages the judge works in. First, problem, motivation, proposed approach, and research plan are each marked same or different, judged as if the other slots were hidden, and a slot that one side leaves empty counts as different. Second, the annotator says whether the two documents describe the same core idea, and records a confidence from one to five. 

\paragraph{Sampling design.} The study covers automated research and uses 40 judged summary pairs containing 160 slot pairs. Half are a proposal against a 2026 paper, taken from the \MatchIdea{} judgments, and half are a proposal against another proposal of the same population, taken from the \DivIdea{} judgments. Sampling at random would leave little to compare, since the judge finds agreement on the whole idea in fewer than one in a thousand of the \MatchIdea{} pairs it rules on, so nearly every drawn pair would be an easy disagreement. We instead stratify on what the judge decided: pair type crossed with its decisions on problem and motivation gives eight cells, and we draw five pairs from each. The sample then realizes 14 of the 16 possible combinations of the four decisions an annotator is asked to repeat, and is close to balanced on each of them as shown in Table~\ref{tab:judgehuman}.

\paragraph{Judge consistency and agreement with annotators.} Three annotators, a master's student, a PhD student, and a PhD in computer science, each answer all 40 pairs, so every slot decision carries three independent human answers, which we reduce to a majority label. To gauge the judge's stability, we run it five times on the same pairs. Table~\ref{tab:judgehuman} reports, slot by slot, how often the judge marks a match, how far its runs agree, and how closely it agrees with the annotators' majority. The five runs agree 77--$90\%$ of the time, so the verdict is largely reproducible. The judge also tracks the annotators' majority closely on the problem and the proposed approach, while the more interpretive motivation and research-plan slots run lower. On most slots this alignment matches or exceeds the agreement among the annotators themselves, so the judge reads a pair about as consistently as a human would.

\begin{table}[t]
\centering
\small
\caption{Human study for the same-idea judge. Judge match counts the sampled pairs the judge marks as matching. Judge--judge (\%) is the percentage agreement among the judge runs (robustness), and Judge--human (\%) the percentage agreement between the judge and the annotators' majority.}
\label{tab:judgehuman}
\begin{tabular}{lccc}
\toprule
Slot & Judge match & Judge--judge & Judge--human \\
\midrule
Problem           & 20 / 40 & 83.00 & 87.50 \\
Motivation        & 20 / 40 & 77.00 & 70.00 \\
Proposed approach & 19 / 40 & 90.00 & 87.50 \\
Research plan     & 18 / 40 & 79.00 & 72.50 \\
\bottomrule
\end{tabular}
\end{table}

\begin{figure}[t]
\centering
\includegraphics[width=\textwidth]{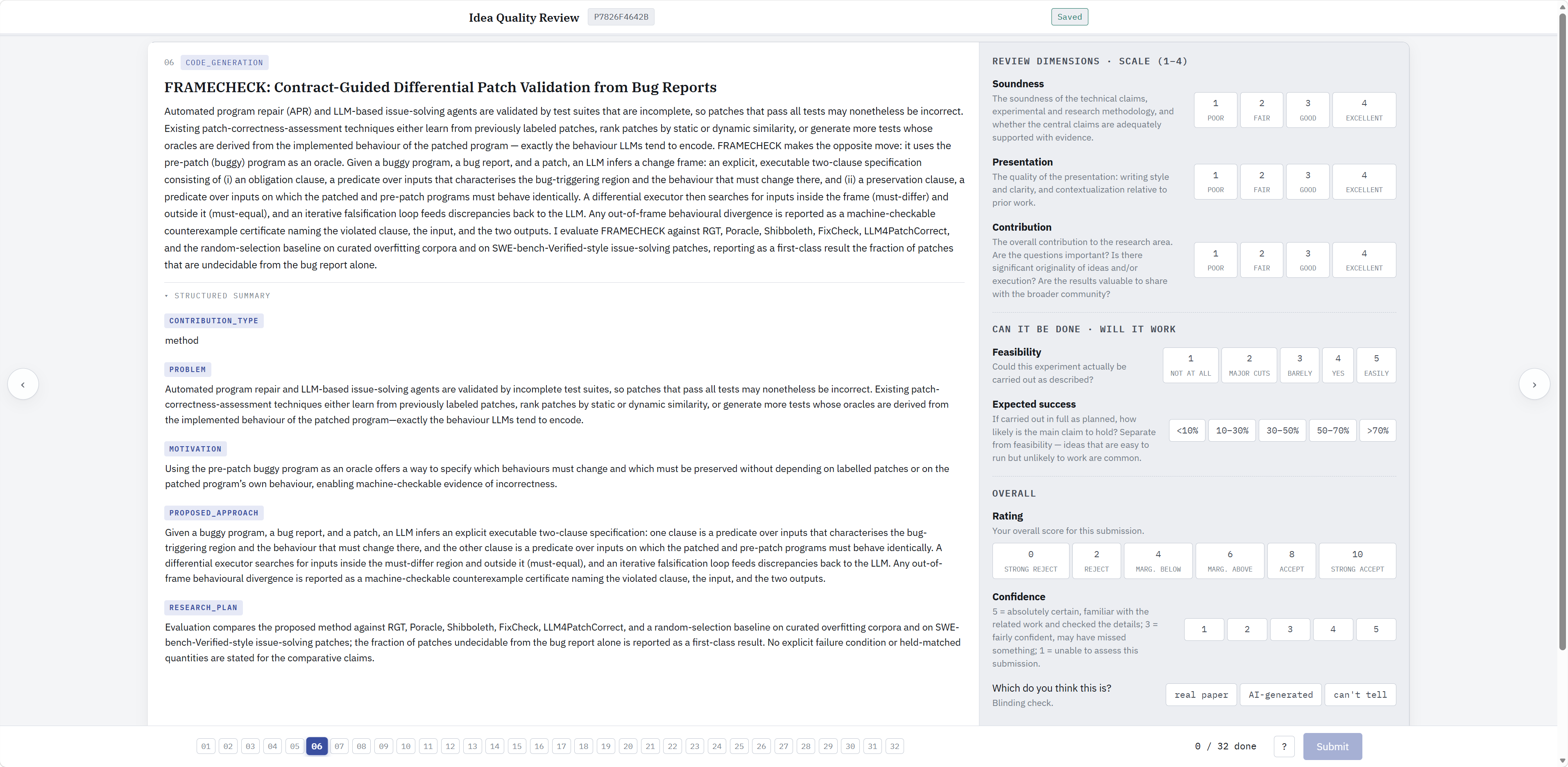}
\caption{The idea quality review interface. The left pane gives the title, abstract, and structured summary; the right pane collects the review dimensions, the two execution questions, the overall rating and confidence, and the blinding check.}
\label{fig:human-ui}
\end{figure}

\begin{table}[t]
\centering
\small
\caption{Sample composition of human study for idea quality. The four document types contribute equally and are balanced across the four areas. Proposals are further spread over the six models at 16 each, and the 192 documents are split into six blocks of 32.}
\label{tab:human-sample}
\begin{tabular}{lrr}
\toprule
Document type & Per area & Total \\
\midrule
Proposal overlapping a 2026 paper & 12 & 48 \\
2026 paper overlapped by a proposal & 12 & 48 \\
Proposal, non-overlapping & 12 & 48 \\
2026 paper, non-overlapping & 12 & 48 \\
\midrule
Total &  & 192 \\
\bottomrule
\end{tabular}
\end{table}

\begin{figure}[t]
\centering
\includegraphics[width=\textwidth]{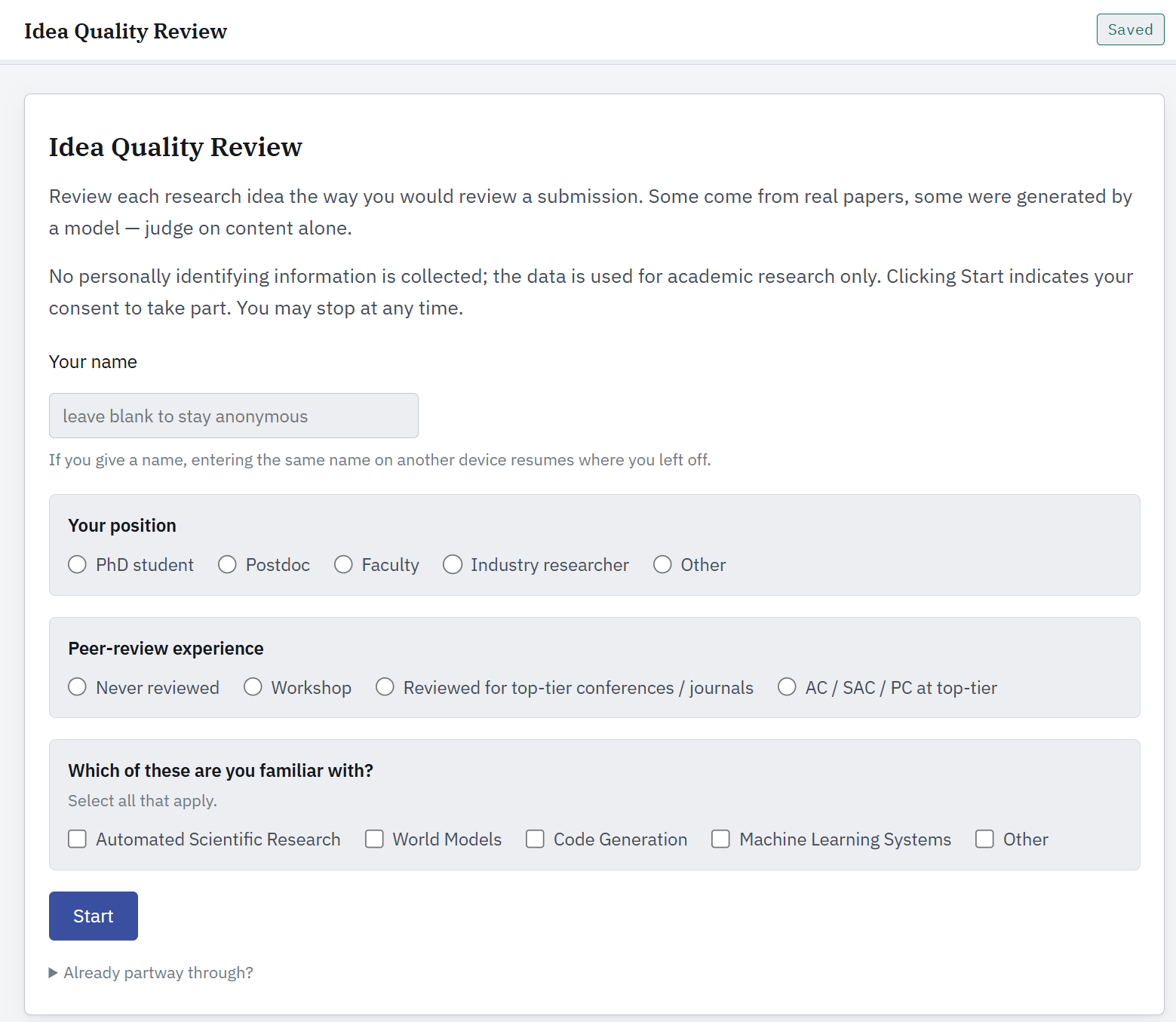}
\caption{The idea quality review interface. Every human reviewer records their background and experience.}
\label{fig:human-ui-identify}
\end{figure}

\subsection{Human Evaluation of Idea Quality}
\label{app:human}

This appendix describes the questionnaire of Sec.~\ref{sec:human}: what a rater sees, how the sample is composed, who rated it, and the comparison split by overlap with the human record.

\paragraph{Questionnaire.} Figure~\ref{fig:human-ui} shows one example. The left pane presents the work as its title, abstract, and the structured summary of Sec.~\ref{sec:metrics}, the five slots contribution type, problem, motivation, proposed approach, and research plan. The right pane collects the ratings: soundness, presentation, and contribution on the review form's four-point scale, feasibility on a five-point scale from not at all to easily, expected success as one of five probability bands, an overall rating from strong reject to strong accept, a confidence from one to five, and a blinding check of real paper, AI-generated, or unclear.

\paragraph{Sampling design.} The questionnaire holds 192 research documents, split evenly between LLM agent proposals and 2026 papers so the blinding check cannot be won by favoring the more common label. Each area contributes 48 documents and each model 16 proposals, drawn from the \textsc{free}, \textsc{topic}, and \textsc{landscape}. The documents come in four kinds, shown in Table~\ref{tab:human-sample}: an AI proposal that overlaps a 2026 paper, the paper it overlaps, an AI proposal that overlaps no candidate paper, and a 2026 paper that no proposal overlaps. Overlap is decided by the slot-level judge of \MatchIdea{}. Among the overlapping proposals, some match their paper on the whole idea and some only on the problem, and both are kept.

\paragraph{Reviewer recruitment and block assignment.} The documents are split into six blocks of 32, each covering the four areas and a subset of the models. A proposal and the 2026 paper it overlaps are placed in different blocks, so no human reviewers see both and a paired comparison cannot turn into a within-reviewer ranking; the paired comparison is therefore made across reviewers, with a reviewer random effect in the analysis. Each reviewer takes one block one time and rates all 32 documents one time. One reviewer may take multiple blocks and record their background and experience as shown in Figure~\ref{fig:human-ui-identify}. We only record reviewers who have review experience. We finally recruited 11 reviewers and collected two records for each block, that is, two review records for each research document.

\paragraph{Rating scales and score normalization.} As shown in Figure~\ref{fig:human-ui}, review design mainly follows the official review form of a major machine learning conference and scores an idea on seven dimensions.
\begin{itemize}[leftmargin=0.3cm, itemsep=-0.3em, topsep=-0.2em]
\item Soundness, Presentation and Contribution use four levels (1 = poor, 2 = fair, 3 = good, 4 = excellent).
\item Feasibility uses five levels (1 = not at all, 5 = easily).
\item Expected success uses five probability bins ($<$10\%, 10--30\%, 30--50\%, 50--70\%, $>$70\%), coded 1--5.
\item Rating uses six levels (0 = strong reject, 2 = reject, 4 = marginally below, 6 = marginally above, 8 = accept, 10 = strong accept).
\item Confidence uses five levels (1 = unable to assess, 5 = absolutely certain).
\item Blinding check asks whether the reviewer thinks the entry is a real 2026 paper, an AI-generated proposal, or they cannot tell.
\end{itemize}

For each population we average the raw scores of a dimension and map the mean linearly onto $[0, 100]$ as shown in Table \ref{tab:human}: $s = 100 \times \frac{\bar{x} - x_{\min}}{x_{\max} - x_{\min}}$, where $x_{\min}$ and $x_{\max}$ are the lowest and highest levels of that dimension's scale.

\section{Method Details}
\label{app:methods}

All four methods run the same loop. This section specifies that loop and then the profile and actions each method changes.

\subsection{Shared Agent Loop}

For LLM agents, an agent harness provides the interface through which a language model observes and acts in an environment. Our harness follows ReAct~\citep{yao2022react}, interleaving reasoning, actions, and observations in the model's prompt, similar tool-based interaction loops supporting research planning in Idea2Plan~\citep{huang2025idea2plan}, idea generation in The AI Scientist-v2~\citep{yamada2025aiscientistv2}, and skill learning~\citep{zhong2026skilllearnbench}. In our setting, one run of the LLM agent within this harness produces a single proposal. The harness provides the agent with a profile, a step budget, and a set of available actions. The agent follows the ReAct loop, deciding for itself which actions to take and how to spend the budget.

\paragraph{Actions.} Every response carries exactly one tagged action, and every step costs one unit of a budget of at most 20 model calls. \texttt{plan} records the current working hypothesis and replaces the previous plan. \texttt{search} takes a query with an optional publication-year filter and runs the hybrid retrieval above over the area's corpus. \texttt{read\_papers} pins retrieved documents into a working set of 50 cards, evicting the least recently requested card when the set is full. \texttt{draft} replaces the working draft. \texttt{summary} condenses the agent history into notes that replace it, leaving the working set intact. \texttt{proposal} submits and ends the agent loop. Figure~\ref{fig:agent_prompt} gives the prompt that defines them.

\begin{promptboxlong}
{\scriptsize\ttfamily\raggedright \textbf{System}\par\smallskip
You are a research scientist preparing ONE new research proposal.\par
\par\smallskip
Your research community is defined as:\par
\slot{the area's definition, as given in Table~\ref{tab:areadefs}}\par
 The literature you can access ends at 2025-12-31.\par
\par\smallskip
You don't need to finish in one step --- you may plan, search, read, draft and revise across multiple steps, one action per step.\par
\par\smallskip
You will be provided with:\par
1. Your researcher profile.\par
2. Your pinned papers --- the full cards you pinned with $<$read\_papers$>$, listed in the order you pinned them, with the count at the end of the block. Nothing is ever lost: the corpus does not change, so any [CAND] id you have seen can be pinned back and its full card reappears here.\par
3. Your history --- an append-only log of what you did, one entry per step, in the order it happened. A $<$search$>$ entry is followed by its $<$information$>$ block (which opens with a receipt counting new vs. already-retrieved results); a $<$read\_papers$>$ entry is followed by a receipt; a $<$plan$>$ or $<$draft$>$ entry records only that you rewrote it --- the text itself lives in your workspace. Past steps are never rewritten --- only a $<$summary$>$ replaces the earlier entries wholesale. In the history every [CAND] result is a title-only line; a line marked '(shown above)' is one that appeared earlier.\par
4. Latest search results --- the full cards of your most recent search. Once you search again this block is replaced, so pin anything you want to keep reading.\par
5. Your workspace --- what you are writing: your latest $<$plan$>$, your latest $<$draft$>$, and any $<$feedback$>$. A $<$summary$>$ clears the plan and the draft but keeps your pinned papers, which is why that action asks you to carry them into your notes.\par
6. Your remaining step budget (each response consumes one step).\par
\par\smallskip
IMPORTANT: You must strictly adhere to the following rules:\par
1. Choose ONLY ONE action from the list below for each response, DO NOT perform more than one action per step.\par
2. Follow the exact syntax format for the selected action, DO NOT create or use any actions other than those listed.\par
3. The corpus is a bounded snapshot of one research community. Thin or empty results mean "not found in this corpus", NOT proof that no such work exists anywhere; ground research-gap claims in the papers you actually retrieved and cited.\par
4. Form a hypothesis early and let evidence attack it. Put a first $<$plan$>$ on the table within your first few steps, then use $<$search$>$ to test it --- searching is how you check, refute or sharpen your idea.\par
5. The episode ends when you output $<$proposal$>$, and you may output it on any step. "Steps remaining: 1" is the deadline: your response then MUST be the $<$proposal$>$, otherwise the episode ends with nothing.\par
6. Every id in your References section MUST be a paper you actually retrieved: copy it verbatim from a [CAND] line anywhere in your history --- search results, your working set, or ids you kept in your summary notes. You are NOT allowed to invent ids or cite papers that never appeared. List each paper at most once.\par
7. Cite inline with bracketed numbers [1], [2], ... matching your References entries. The same paper always reuses the same number. Do NOT write raw [CAND] ids inside the prose --- they belong only in the References section.\par
8. The proposal must be concrete and technically grounded: a specific falsifiable core idea, not a survey or a vague direction. It must be new relative to the literature you retrieved, and your proposal must say exactly what is new: in its related-work discussion, name the closest papers you found and state what your proposal does that they do not --- a different problem framing, a different mechanism, a different evaluation that would settle something they left open. Restating or incrementally tuning a retrieved paper is not a proposal. Every proposal has related work: if you found none, that almost always means the search was too narrow, not that the idea is unprecedented --- claim the latter only if you are prepared to defend it. Claim novelty only against what you actually retrieved and cited (see rule 3).\par
\par\smallskip
Valid actions:\par
1. $<$plan$>$ research state $<$/plan$>$: your current working hypothesis and anything else worth carrying forward --- including but not limited to (a) the idea you are proposing, stated as a specific falsifiable claim; (b) what would show it is wrong, already done, or not worth doing; (c) a finding, a constraint you hit, or a question to settle later. A rough hypothesis on little evidence is fine and expected. Issue $<$plan$>$ again to replace it: sharpen it as evidence accumulates, and when evidence kills it --- the claim is wrong, or the retrieved literature already does it --- move to a different gap instead of defending it. Apart from your $<$draft$>$, this is the only note that survives into the next step.\par
2. $<$search$>$ query $<$/search$>$: search the local literature corpus (a short natural-language or keyword query, $<$= 25 words). Optionally append a publication-year filter after a pipe, e.g. $<$search$>$ repository level code generation | years: 2020-2025 $<$/search$>$ (a single year also works). The engine combines keyword (BM25) and semantic matching.\par
3. $<$read\_papers$>$ id1, id2, ... $<$/read\_papers$>$: pin papers you have already retrieved into your $<$working\_set$>$ --- use it to read a paper whose card has been compressed to a title-only line, or to keep an important one in full while you work. Copy the ids verbatim from [CAND] lines. The working set holds 50 cards; when it is full the least recently requested one is evicted. A card carries the paper's metadata and its abstract --- the corpus has no full text, so the abstract is all there is of a paper's content.\par
4. $<$draft$>$ revised or newly written working draft of your proposal $<$/draft$>$: write or revise your working draft based on what you have learned --- this is where your $<$plan$>$ hypothesis becomes an actual proposal. The latest draft replaces the previous one. Ideally you start drafting while 5-10 steps remain, so there is room left to sharpen the proposal rather than writing it in the last step.\par
5. $<$summary$>$ research notes $<$/summary$>$: condense your history into notes. Keep the plan, the current draft state, key findings, and the [CAND] ids you may still want to cite (one line per id is enough --- you can pin any of them back into your working set later with $<$read\_papers$>$). Your subsequent history will be replaced by this summary; your $<$working\_set$>$ is preserved automatically.\par
6. $<$proposal$>$ final proposal in markdown $<$/proposal$>$: output the final proposal as a self-contained markdown document:\par
   - First line: "\# $<$specific title$>$".\par
   - Then a paragraph starting with "**Abstract**: " --- 4-8 sentences summarizing the whole proposal (this is what the community will first see).\par
   - Then a related-work section: the closest papers you retrieved, named, and what your proposal does that they do not.\par
   - Then the rest of the proposal body in free-form prose, structured however serves the idea best. A strong proposal covers: motivation, the research gap that follows from the related work, the core idea (crisp and falsifiable), method, implementation plan, experiment plan (datasets, baselines, metrics, ablations), and expected contribution --- but the organization, depth, and voice are yours.\par
   - Cite inline with [1], [2], ... throughout the body.\par
   - End with a "\#\# References" section, exactly one line per cited paper, in the format:\par
     [n] $<$the id copied from the [CAND] line, WITHOUT any brackets around it$>$ | role: method|baseline|motivation|extension|background | one sentence on how this paper is used\par
     For example: [1] D3f2a9c01d2e4 | role: baseline | We compare against this retrieval-augmented generator.\par
\par\smallskip
Format: use exactly one tag pair per response --- $<$plan$>$, $<$search$>$, $<$read\_papers$>$, $<$draft$>$, $<$summary$>$, or $<$proposal$>$. No text outside the tags.\par
}
\end{promptboxlong}
\nopagebreak
\captionof{figure}{The system prompt shared by every method. It fixes the loop: what the LLM agent is shown at each step and the six actions it may take, one per response. }
\label{fig:agent_prompt}

\subsection{Method Configurations}

The methods differ in the profile placed at the top of that prompt and in the actions appended to it.

\paragraph{Agent profiles.} \textsc{Free} states only that the LLM agent is an independent researcher in the community defined above, with no direction and no prescription of how to work. \textsc{Topic} names one research area and adds that the LLM agent chooses its own problems within it. \textsc{Frontier} reuses the \textsc{topic} profile and allocation unchanged, so that the only difference between them is access to earlier rounds.

\paragraph{Topic allocation.} The areas named in \textsc{topic} profiles come from the topic labels already carried by OpenAlex records. The set of topics itself was built from the citation network. OpenAlex started with works that have incoming and outgoing citations and clustered them by citation relationships~\citep{openalex2024topics,vaneck2024classification}.
We collect topic label on a $\le 2025$ paper, and we weight topics by an exponential decay in publication year with a half-life of two years, which favors recent practice without cutting the earlier years off. The resulting distribution is allocated across the population by largest remainder. As shown in Table~\ref{tab:topic_allocation}, topics read like Topic Modeling, Machine Learning in Materials Science, Scientific Computing and Data Management, and Multi-Agent Systems and Negotiation. Some topics look unrelated to the specific research area, but the topic allocation is a rough guide to the areas the LLM agent is expected to explore, and it does not constrain the search or the proposal. The LLM agent may search for any topic in the corpus and propose any idea it can justify. At a population of 300 the allocation reaches between 146 and 177 distinct topics out of the several thousand that carry prior mass.

\begin{table}[t]
\centering
\small
\setlength{\tabcolsep}{4pt}
\caption{The \textsc{topic} allocation in each area. \# Used is the number of topics that receive at least one LLM agent at a population of 300, out of the \# $\le 2025$ topics that carry prior mass in the area's corpus. The last column lists the fifteen topics that take the largest share, in that order.}
\label{tab:topic_allocation}
\begin{tabular}{@{}l@{\hspace{0.8em}}rr@{\hspace{0.8em}}>{\raggedright\arraybackslash}p{0.55\textwidth}@{}}
\toprule
Area & \# Used & \# $\le 2025$ & Topics, most allocated first \\
\midrule
Machine learning systems & 156 & 2{,}665 & Topic Modeling, Natural Language Processing Techniques, Multimodal Machine Learning Applications, Advanced Neural Network Applications, Domain Adaptation and Few-Shot Learning, Parallel Computing and Optimization Techniques, Neural Networks and Applications, Speech Recognition and Synthesis, Advanced Image and Video Retrieval Techniques, Generative Adversarial Networks and Image Synthesis, Adversarial Robustness in Machine Learning, Semantic Web and Ontologies, Speech and dialogue systems, Explainable Artificial Intelligence (XAI), Artificial Intelligence in Healthcare and Education \\
\addlinespace[4pt]
Code generation & 146 & 2{,}609 & Topic Modeling, Natural Language Processing Techniques, Software Engineering Research, Multimodal Machine Learning Applications, Software Testing and Debugging Techniques, Semantic Web and Ontologies, Advanced Malware Detection Techniques, Artificial Intelligence in Healthcare and Education, Adversarial Robustness in Machine Learning, Explainable Artificial Intelligence (XAI), Domain Adaptation and Few-Shot Learning, Speech and dialogue systems, Software Reliability and Analysis Research, Multi-Agent Systems and Negotiation, Software System Performance and Reliability \\
\addlinespace[4pt]
World models & 156 & 2{,}296 & Reinforcement Learning in Robotics, Multimodal Machine Learning Applications, Generative Adversarial Networks and Image Synthesis, Topic Modeling, Domain Adaptation and Few-Shot Learning, Natural Language Processing Techniques, Human Pose and Action Recognition, Advanced Vision and Imaging, Robot Manipulation and Learning, Advanced Neural Network Applications, Autonomous Vehicle Technology and Safety, Adversarial Robustness in Machine Learning, Computer Graphics and Visualization Techniques, Human Motion and Animation, Advanced Image and Video Retrieval Techniques \\
\addlinespace[4pt]
Automated research & 177 & 2{,}387 & Topic Modeling, Machine Learning in Materials Science, Natural Language Processing Techniques, Computational Drug Discovery Methods, Artificial Intelligence in Healthcare and Education, Scientific Computing and Data Management, Explainable Artificial Intelligence (XAI), Multimodal Machine Learning Applications, Semantic Web and Ontologies, Model Reduction and Neural Networks, Multi-Agent Systems and Negotiation, Neural Networks and Applications, Biomedical Text Mining and Ontologies, Evolutionary Algorithms and Applications, Ethics and Social Impacts of AI \\
\bottomrule
\end{tabular}
\end{table}

\paragraph{Tag extraction.} The tags behind \textsc{landscape} are extracted from the 2025 papers. A language model with prompt as shown in Figure~\ref{fig:tag_prompt} reads one paper at a time and names the task it addresses and the method it uses, drawing on a tag set that starts empty and grows as papers are processed in publication order. After each batch a second pass merges newly proposed tags into existing entries. A final pass by hand merges the variants that survive, such as modifier and hyphen differences and the same name appearing on both axes. In automated research this yields task tags such as autonomous experimentation, equation discovery, AI scientist, and scientific hypothesis generation, and method tags such as llm agent, multi-agent system, benchmark construction, and retrieval-augmented generation.

\begin{promptboxlong}
{\scriptsize\ttfamily\raggedright \textbf{System}\par\smallskip
You tag a paper with the tasks it addresses and the methods it uses, using a controlled vocabulary that grows over time.\par
\par\smallskip
CURRENT TASK VOCABULARY\par
\slot{every task tag introduced so far, one id and label per line}\par
\par\smallskip
CURRENT METHOD VOCABULARY\par
\slot{every method tag introduced so far, one id and label per line}\par
\par\smallskip
task - WHAT is being done or improved: the capability or object of study.\par
method - HOW: the mechanisms, one tag each.\par
\par\smallskip
FOR EACH TAG, PREFER AN EXISTING ONE.\par
Reuse an existing tag whenever it genuinely covers this paper, even if the paper's own wording differs - that is the point of a controlled vocabulary. Give its id. Create a new tag ONLY when no existing tag covers the concept: give id "NEW" and put the new tag in the label field. Never create a near-duplicate of an existing tag.\par
\par\smallskip
TAGS ARE ATOMIC.\par
Each tag names ONE concept. Never bundle two concepts into one tag. "multi-stage llm agents with human feedback" is WRONG - that is three concepts. Emit them separately: "llm agent", "multi-stage pipeline", "human-in-the-loop". 1-3 task tags and 1-4 method tags for the paper as a whole.\par
\par\smallskip
TAG STYLE\par
1-4 words, lowercase, no articles, no trailing punctuation. A canonical name other papers would also carry, not a description of this paper. Discriminative: a tag that would apply to most papers in this area carries no information. Specific enough to tell research directions apart, general enough to recur.\par
\par\smallskip
RULES\par
R0. English only, whatever the language of the input text.\par
R1. Community vocabulary, never this paper's coined names. NEVER use a name this paper coins for its own system, framework, benchmark, metric or concept. Names of PRIOR work, third-party models, datasets and libraries ARE allowed.\par
R2. Tags must be substantive AND discriminative. These are BANNED as tags, and so is anything equally broad: machine learning, deep learning, artificial intelligence, natural language processing, computer vision, neural network, large language model, llm, transformer, foundation model, data analysis, optimization, evaluation, improving performance, novel framework. They name a whole field or a truism, not a research direction. If a side genuinely supports nothing more specific, emit fewer tags rather than a broad one.\par
R3. Do not invent. Tag only what the text states.\par
R4. No results or numbers anywhere.\par
R5. A survey, review or taxonomy paper contributes no tags of its own: return empty lists with extractable=false.\par
\par\smallskip
Also return ONE contribution\_type for the whole paper, exactly one of these five in `primary':\par
\slot{the five contribution type definitions of Figure~\ref{fig:schema_prompt}}\par
Put another type in `secondary' only if the work genuinely delivers it too, else [].\par
\par\smallskip
And ONE theme judgement for the whole paper.\par
Area definition:\par
\slot{the area's definition, as given in Table~\ref{tab:areadefs}}\par
Judge by the paper's main contribution. A paper that only mentions the area in passing, or uses its tools for another goal, does NOT belong.\par
\medskip
\textbf{User}\par\smallskip
Text: \slot{title and abstract}\par
\par\smallskip
Return JSON with exactly this shape:\par
\{"tasks": [\{"id": "T001 or NEW", "label": "..."\}], "methods": [\{"id": "M001 or NEW", "label": "..."\}], "contribution\_type": \{"primary": "method", "secondary": []\}, "in\_theme": true, "theme\_confidence": 0.0, "theme\_reason": "...", "extractable": true, "note": "..."\}\par
Always fill both id and label: reusing an existing tag means giving its id and copying its label verbatim; creating one means id "NEW" and your new tag as the label. `primary' must be one of: theory, method, empirical\_analysis, artifact, position. contribution\_type and the theme judgement are ALWAYS filled, even when the tag lists are empty.\par
}
\end{promptboxlong}
\nopagebreak
\captionof{figure}{The tag extraction prompt.}
\label{fig:tag_prompt}

\begin{promptboxlong}
{\scriptsize\ttfamily\raggedright RESEARCH LANDSCAPE OF THIS COMMUNITY (2025)\par
\par\smallskip
The 2025 literature of this community is indexed into \slot{task tags} task tags and \slot{method tags} method tags; a typical paper carries \slot{task tags per paper} task tags and \slot{method tags per paper} method tags. The distribution is long-tailed --- about half of the tags appear in only one paper.\par
\par\smallskip
Listed below are the 100 most widely used tags on each axis and the 100 most common task-method combinations, each with the share of the papers carrying it, ordered from most to least common. Tags outside these lists exist and are queryable by name.\par
\par\smallskip
These tags describe what was studied in 2025. New work often continues these directions, but it also opens directions that are not in this list at all. Treat it as a reference, not a menu.\par
\par\smallskip
TASK TAGS\par
\slot{the hundred most common task tags, each with its share, separated by semicolons}\par
\par\smallskip
METHOD TAGS\par
\slot{the hundred most common method tags, each with its share}\par
\par\smallskip
COMMON TASK-METHOD COMBINATIONS\par
\slot{the hundred most common task-method pairs, each with its share}\par
}
\end{promptboxlong}
\nopagebreak
\captionof{figure}{The landscape block, which closes the system prompt of \textsc{landscape} and follows the description of the tag retrieval action. One block is built per area from that area's 2025 papers, and every \textsc{landscape} LLM agent in that area receives the same text.}
\label{fig:landscape_block}

\begin{longtable}{@{}l@{\hspace{1em}}>{\raggedright\arraybackslash}p{\dimexpr\textwidth-4.5em\relax}@{}}
\caption{What each area fills into the landscape block of Figure~\ref{fig:landscape_block}. Each group header gives the papers the shares are taken over and the size of the two tag vocabularies. }\label{tab:landscape}\\
\toprule
Axis & Ten most common, with the share of the area's 2025 papers carrying each \\
\midrule
\endfirsthead
\toprule
Axis & Ten most common, with the share of the area's 2025 papers carrying each \\
\midrule
\endhead
\bottomrule
\endlastfoot
\multicolumn{2}{@{}l}{\textbf{Machine learning systems}, 824 task tags and 2{,}072 method tags} \\
\addlinespace[2pt]
Task & llm inference acceleration (7.6\%), llm serving (5.6\%), low-bit quantization (4.8\%), mixture-of-experts (4.6\%), large language model compression (4.1\%), large language model training (3.7\%), post-training quantization (3.7\%), edge llm inference (3.6\%), kv cache compression (3.5\%), distributed training (3.2\%) \\
Method & post-training quantization (4.3\%), speculative decoding (4.1\%), knowledge distillation (3.8\%), sparse attention (3.4\%), algorithm-hardware co-design (3.2\%), token pruning (3.1\%), mixture of experts (2.8\%), mixed-precision quantization (2.4\%), reinforcement learning (2.4\%), performance modeling (2.3\%) \\
Pair & low-bit quantization $\times$ post-training quantization (1.8\%), post-training quantization $\times$ post-training quantization (1.6\%), visual token compression $\times$ token pruning (1.5\%), kv cache compression $\times$ key-value cache compression (1.2\%), efficient vision-language model inference $\times$ token pruning (1.2\%), autoregressive decoding acceleration $\times$ speculative decoding (1.2\%), llm inference acceleration $\times$ speculative decoding (1.0\%), mixture-of-experts $\times$ expert routing (1.0\%), long-context inference $\times$ sparse attention (0.9\%), low-bit quantization $\times$ mixed-precision quantization (0.9\%) \\
\addlinespace[5pt]
\multicolumn{2}{@{}l}{\textbf{Code generation}, 618 task tags and 805 method tags} \\
\addlinespace[2pt]
Task & code generation (19.9\%), text-to-sql (10.0\%), code generation evaluation (8.3\%), automated program repair (6.6\%), benchmark construction (5.5\%), code generation benchmark (3.2\%), software engineering agent (3.1\%), code translation (2.9\%), verilog code generation (2.5\%), code completion (2.3\%) \\
Method & benchmarking (15.5\%), comparative evaluation (13.8\%), fine-tuning (12.3\%), retrieval-augmented generation (8.5\%), iterative refinement (7.8\%), llm agent (7.3\%), multi-agent system (7.0\%), benchmark generation (6.9\%), prompt engineering (6.2\%), execution feedback (5.2\%) \\
Pair & code generation evaluation $\times$ benchmarking (4.1\%), code generation evaluation $\times$ comparative evaluation (3.3\%), benchmark construction $\times$ benchmarking (2.8\%), code generation $\times$ fine-tuning (2.6\%), code generation $\times$ iterative refinement (2.5\%), code generation $\times$ comparative evaluation (2.2\%), code generation $\times$ retrieval-augmented generation (2.1\%), code generation $\times$ benchmarking (2.0\%), code generation benchmark $\times$ benchmarking (1.9\%), code generation $\times$ prompt engineering (1.8\%) \\
\addlinespace[5pt]
\multicolumn{2}{@{}l}{\textbf{World models}, 337 task tags and 516 method tags} \\
\addlinespace[2pt]
Task & world model learning (25.7\%), autonomous driving (11.9\%), robot manipulation (10.6\%), model-based reinforcement learning (8.9\%), future state prediction (4.5\%), interactive world generation (4.4\%), world model evaluation (4.0\%), driving scene generation (2.3\%), controllable video synthesis (2.0\%), continuous control (1.9\%) \\
Method & model-based reinforcement learning (16.3\%), diffusion model (11.7\%), generative world model (5.1\%), autoregressive generation (4.9\%), action-conditioned dynamics model (4.7\%), model predictive control (3.9\%), synthetic data generation (3.9\%), benchmark framework (3.7\%), model-based planning (3.7\%), diffusion video generation (3.0\%) \\
Pair & world model learning $\times$ diffusion model (3.7\%), world model learning $\times$ autoregressive generation (2.8\%), autonomous driving $\times$ model-based reinforcement learning (2.7\%), world model evaluation $\times$ benchmark framework (2.3\%), world model learning $\times$ model-based reinforcement learning (2.2\%), model-based reinforcement learning $\times$ model-based reinforcement learning (2.0\%), world model learning $\times$ action-conditioned dynamics model (2.0\%), world model learning $\times$ generative world model (1.9\%), autonomous driving $\times$ diffusion model (1.6\%), robot manipulation $\times$ diffusion model (1.6\%) \\
\addlinespace[5pt]
\multicolumn{2}{@{}l}{\textbf{Automated research}, 318 task tags and 424 method tags} \\
\addlinespace[2pt]
Task & autonomous experimentation (10.9\%), equation discovery (10.6\%), AI scientist (10.0\%), symbolic regression (6.8\%), scientific hypothesis generation (6.4\%), research idea generation (5.2\%), agent benchmarking (4.2\%), human-AI collaboration (4.1\%), automated peer review (3.9\%), scientific discovery (3.5\%) \\
Method & llm agent (22.1\%), multi-agent system (17.3\%), benchmark construction (10.9\%), retrieval-augmented generation (10.1\%), iterative refinement (7.6\%), symbolic regression (7.6\%), human-in-the-loop (6.8\%), llm-as-judge (4.8\%), self-driving laboratory (4.6\%), closed-loop discovery workflow (4.5\%) \\
Pair & equation discovery $\times$ symbolic regression (6.0\%), autonomous experimentation $\times$ self-driving laboratory (4.2\%), AI scientist $\times$ llm agent (4.1\%), agent benchmarking $\times$ benchmark construction (3.9\%), AI scientist $\times$ multi-agent system (3.5\%), autonomous experimentation $\times$ closed-loop discovery workflow (3.1\%), autonomous experimentation $\times$ llm agent (2.7\%), human-AI collaboration $\times$ human-in-the-loop (2.4\%), autonomous experimentation $\times$ multi-agent system (2.1\%), symbolic regression $\times$ genetic programming (1.9\%) \\
\end{longtable}

\paragraph{Landscape prompt block.} The prompt lists the hundred most common task tags and the hundred most common method tags among the area's 2025 papers, together with common task-method pairs, each with the share of those papers carrying it and ordered by that share. Figure~\ref{fig:landscape_block} gives the block template and Table~\ref{tab:landscape} what each area fills into it.

\paragraph{Frontier rounds.} The population is split into equal rounds, so a run of 300 proposals is two rounds of 150. 

\section{Language Models and Inference Settings}
\label{app:implementation}

Five roles call a language model: the extractor that turns a document into the idea schema, the tagger that builds the task and method tags, the model that screens papers for relevance to a specific research area, the idea judge that decides schema slot agreement, and the LLM agents that write proposals. Table~\ref{tab:llmroles} lists the model behind each and the settings we set.

\begin{table}[t]
\centering
\small
\setlength{\tabcolsep}{4pt}
\caption{The language model behind each role. Context window is the vendor's stated limit for the checkpoint we called; our prompts stay far below it. Max output of Llama 4 Scout is the 8{,}192 as the model allows. The six LLM agent models are the ones compared in Table~\ref{tab:main}.}
\label{tab:llmroles}
\begin{tabular}{llrrl}
\toprule
Role & Model & Context & Max output & Reply \\
\midrule
Schema extraction & DeepSeek-V4-Pro & 1M & 16{,}384 & JSON \\
Tag extraction & DeepSeek-V4-Flash & 1M & 16{,}384 & JSON \\
Area screening & DeepSeek-V4-Flash & 1M & 4{,}096 & JSON \\
Idea matching & GPT-5.6 Luna & 1M & 1{,}200 & JSON \\
\midrule
LLM agent & Llama 4 Scout & 10M & 8{,}192 & text \\
LLM agent & Claude Haiku 4.5 & 200K & 16{,}384 & text \\
LLM agent & Gemini 3.5 Flash-Lite & 1M & 16{,}384 & text \\
LLM agent & DeepSeek-V4-Flash & 1M & 16{,}384 & text \\
LLM agent & Qwen3.8-27B & 262K & 16{,}384 & text \\
LLM agent & K2 Horizon 375B A23B & 512K & 16{,}384 & text \\
\bottomrule
\end{tabular}
\end{table}

\begin{itemize}[leftmargin=0.3cm, itemsep=-0.3em, topsep=-0.2em]
\item \textbf{Schema extraction.} Both sides of every comparison are described by the same extractor, with the same prompt and the same retry policy. A different extractor on either side would compare two extraction styles rather than two ideas, so the model and the endpoint are pinned in code, independent of whichever model wrote the proposal. We pin DeepSeek-V4-Pro on the vendor's own endpoint, the largest checkpoint in that family rather than the Flash one an LLM agent runs, because extraction is an instrument: a field read wrong here moves every score computed from the schema.
\item \textbf{Tag extraction.} The task and method tags behind \textsc{landscape} are extracted by DeepSeek-V4-Flash in JSON mode. The vocabulary starts empty and grows batch by batch: each batch sees the vocabulary built so far, reuses an existing tag where one fits, and marks a new tag explicitly when none does. A separate merge pass resolves the aliases that survive.
\item \textbf{Area screening.} The area screening model is DeepSeek-V4-Flash in JSON mode. It reads a paper's title and abstract and decides whether the paper is in the area, returning a boolean and a confidence score.
\item \textbf{Idea matching.} Slot agreement is decided by GPT-5.6 Luna in JSON mode, with output capped at 1{,}200 tokens. Each pair is judged once, and which document is shown first is drawn from a hash of the pair and a fixed seed: random across pairs, reproducible on a rerun, and unaffected by worker or retry order. Every call is logged with its input, its output, and the order it drew, since the judge is the only model whose output becomes a number we report; Appendix~\ref{app:eval} reports its agreement with human annotators. No population is judged by its own model family.
\item \textbf{LLM agents.} The six models of Table~\ref{tab:main} run one implementation, one prompt, and one step budget of twenty steps. Replies are plain text, and the action tag is parsed from the text. The output cap is 16{,}384 tokens, except on Bedrock, where Llama 4 Scout is capped at the 8{,}192 the platform allows.
\end{itemize}

The request body carries the inputs and nothing else. Temperature, top-$p$, sampling seed, and tool definitions are never sent, so each endpoint runs at its own default. Two parameters are exceptions. JSON mode is set wherever the reply is parsed as JSON, and a maximum output length is sent where the API requires one or where a stage has its own output budget.

\section{Additional Analyses}

This section reports the checks the main results rest on and the additional analyses of Sec.~\ref{sec:analysis}. The checks ask whether a finding survives a second run, a change of model, or the possibility that a model has already read the 2026 papers. The additional analyses give the counts, schemas, and distributions that the main text only summarizes.

\subsection{Model Selection for Each Analysis}
\label{app:modelsel}

Every analysis in Sec.~\ref{sec:analysis} includes DeepSeek-V4-Flash, so that its results can be read against one another. The model paired with it changes from one analysis to the next, chosen for what that analysis needs and, taken together, to cover as many of the six models as possible.

\begin{itemize}[leftmargin=0.3cm, itemsep=-0.3em, topsep=-0.2em]
  \item \textbf{Direction coverage} (Sec.~\ref{sec:gap}) uses DeepSeek-V4-Flash alone. It labels every proposal of four populations and every paper of the human reference with the task and method vocabulary.
  \item \textbf{Slot decomposition} (Sec.~\ref{sec:slots}) adds Llama 4 Scout, which reaches the highest \MatchIdea{} of the six models. Pairing it with DeepSeek-V4-Flash shows whether a high-matching and a low-matching population fall short in the same slots or in different ones.
  \item \textbf{Case study} (Sec.~\ref{sec:case}) draws Case 1 from Qwen3.8-27B and Case 2 from DeepSeek-V4-Flash. Each case needs one human paper matched by two proposals from a single model, one agreeing on the approach and one not, and these two models supply the clearest such triples.
  \item \textbf{Population scaling} (Sec.~\ref{sec:scale}) adds Gemini 3.5 Flash-Lite, so that the shape of the scaling curve does not rest on one model.
\end{itemize}

Claude Haiku 4.5 and K2 Horizon 375B A23B do not enter these four analyses. Both are covered by Table~\ref{tab:main}, by the human evaluation of Sec.~\ref{sec:human}, by the contribution distribution of Appendix~\ref{sec:contribution}, and by the leakage comparison of Appendix~\ref{app:leakage}, each of which reports all six models.

\subsection{Robustness at a Fixed Population Size}

Table~\ref{tab:main} reports scores for one population of 300 proposals per setting. To assess the stability of the scores and model rankings across runs, we generate a second population of 300 proposals for the \textsc{free} and \textsc{topic} settings with Gemini 3.5 Flash-Lite and DeepSeek-V4-Flash in world models and automated research. Table~\ref{tab:robustness} compares the two runs and reports the change in each metric. Across the tested settings, absolute changes are at most 0.44 percentage points for the match metrics and 2.20 points for the diversity metrics. These results support the stability of these diversity and model-level patterns across the two runs.

\begin{table}[tb]
\centering
\small
\caption{Comparison of two runs with 300 proposals per setting in world models and automated research. Scores are percentages for \MatchCite{}, \MatchIdea{}, \DivCite{}, and \DivIdea{}. $\Delta$ is Run 2 minus Run 1, in percentage points.}
\label{tab:robustness}
\begin{tabular}{ll *{6}{r}}
\toprule
& & \multicolumn{3}{c}{World models} & \multicolumn{3}{c}{Automated research} \\
\cmidrule(lr){3-5}\cmidrule(lr){6-8}
Method & Metric & Run 1 & Run 2 & $\Delta$ & Run 1 & Run 2 & $\Delta$ \\
\midrule
\multicolumn{8}{l}{Gemini 3.5 Flash-Lite~\citep{google2026gemini35flashlite}} \\
\addlinespace[2pt]
\multirow{4}{*}{\textsc{free}} & \MatchCite{} & 4.89 & 4.74 & $-$0.15 & 7.18 & 6.88 & $-$0.30 \\
 & \MatchIdea{} & 3.38 & 3.11 & $-$0.27 & 2.22 & 2.42 & $+$0.20 \\
 & \DivCite{} & 4.76 & 4.99 & $+$0.23 & 5.52 & 5.71 & $+$0.19 \\
 & \DivIdea{} & 18.76 & 18.73 & $-$0.03 & 12.92 & 11.04 & $-$1.88 \\
\addlinespace[3pt]
\multirow{4}{*}{\textsc{topic}} & \MatchCite{} & 5.33 & 5.14 & $-$0.19 & 6.58 & 6.36 & $-$0.22 \\
 & \MatchIdea{} & 3.86 & 3.97 & $+$0.11 & 2.14 & 2.58 & $+$0.44 \\
 & \DivCite{} & 13.35 & 13.53 & $+$0.18 & 20.22 & 20.11 & $-$0.11 \\
 & \DivIdea{} & 52.52 & 54.72 & $+$2.20 & 62.21 & 61.24 & $-$0.97 \\
\midrule
\multicolumn{8}{l}{DeepSeek-V4-Flash~\citep{deepseekai2026v4}} \\
\addlinespace[2pt]
\multirow{4}{*}{\textsc{free}} & \MatchCite{} & 4.77 & 4.70 & $-$0.07 & 6.40 & 5.97 & $-$0.43 \\
 & \MatchIdea{} & 1.45 & 1.78 & $+$0.33 & 1.67 & 1.69 & $+$0.02 \\
 & \DivCite{} & 13.50 & 13.50 & 0.00 & 7.57 & 7.25 & $-$0.32 \\
 & \DivIdea{} & 42.79 & 41.02 & $-$1.77 & 23.16 & 22.52 & $-$0.64 \\
\addlinespace[3pt]
\multirow{4}{*}{\textsc{topic}} & \MatchCite{} & 5.16 & 5.17 & $+$0.01 & 7.72 & 7.54 & $-$0.18 \\
 & \MatchIdea{} & 1.76 & 1.71 & $-$0.05 & 2.09 & 2.01 & $-$0.08 \\
 & \DivCite{} & 32.51 & 32.68 & $+$0.17 & 25.39 & 27.07 & $+$1.68 \\
 & \DivIdea{} & 69.06 & 67.82 & $-$1.24 & 55.80 & 57.00 & $+$1.20 \\
\bottomrule
\end{tabular}
\end{table}

\subsection{The Size of Human Reference}
\label{app:human300}

Every score in Table~\ref{tab:main} is measured against the complete set of 2026 papers in the area, from 1{,}136 to 5{,}188 of them, and the human level applies the same metrics to those papers themselves, each compared only with the others. The Human and LLM agent are sized for different reasons. The reference is complete because it is the record of what a community actually published that year, and the human level is meant to be a reference line for real research. Keeping it fixed also gives every population the same denominator, which is what makes scores comparable across models, methods, and areas. The 300 proposals per population are a deliberate trade-off. Sec.~\ref{sec:scale} grows populations to the size of the reference, 3.8 to 17 times larger, and matching rises slowly and then almost flattens, so we can see that for LLM agents, 300 proposals already place a population close to where a far larger one lands. Asking for more would raise the inference cost of running IdeaTaste without changing what it measures. Therefore, we keep the population size at 300 and let the human reference be whatever it is. In this setting, we can compare the performance of different LLM agents, and also see whether any of them reach the human level.

\subsection{Leakage and Benchmark Robustness}
\label{app:leakage}
 
Four of the six models in Table~\ref{tab:main} were released in 2026 and may have seen part of the reference, which covers January to August 2026, so they could in principle reproduce 2026 papers from memory rather than through research judgment. This choice is deliberate, since a benchmark of research taste should evaluate the models that current AI scientists are built on. We pair these releases with two 2025 models whose training data end before the reference year, which gives a setting free of exposure within the same benchmark. Below we describe one design constraint that limits exposure, test whether exposure helps, and check whether the findings differ between the two settings.
 
\paragraph{Retrieval restricted to the corpus.} An LLM agent cites only papers it retrieves from the corpus of its area, which ends in 2025, so a 2026 paper can never enter its citation set. This constraint does not rule out indirect exposure, since a model that has read a 2026 paper could still reuse its references or restate its idea. The test below addresses this remaining channel.
 
\paragraph{Models released before and after the reference year.} Meta reports an August 2024 pretraining cutoff for Llama 4 Scout, and Anthropic reports that the public internet data of Claude Haiku 4.5 end in February 2025~\citep{meta2025llama4modelcard,anthropic2025haiku45systemcard}. Neither can have seen a paper first released in 2026, so the two serve as temporal controls for Gemini 3.5 Flash-Lite, DeepSeek-V4-Flash, Qwen3.8-27B, and K2 Horizon 375B A23B. If exposure helped, the later releases should lead at the idea level, where memorized content would matter most. They trail instead. Averaged over areas and methods, the two controls reach 3.10 on \MatchIdea{} against 1.68 for the 2026 releases, and Llama 4 Scout, with the earliest cutoff, scores highest of all six. The 2026 releases score slightly higher on \MatchCite{}, 5.06 against 4.14, but Claude Haiku 4.5 comes within 0.6 points of each of them, so this difference more plausibly reflects model capability than exposure. The two levels even move in opposite directions across models, as those with higher \MatchCite{} tend to have lower \MatchIdea{}. Access to the relevant literature is therefore not what limits recovery.
 
\paragraph{Consistent findings in both settings.} The findings of Sec.~\ref{sec:main} hold in both groups. Matching stays far below the human level, at 22\% on citations and 26\% on ideas for the 2025 models and 28\% and 14\% for the 2026 releases. No population in either group reaches the human \DivIdea{}, and \textsc{landscape} ranks last on \MatchCite{} in most settings of both. The only difference is the relatively high \MatchIdea{} of Llama 4 Scout, which runs against exposure rather than with it. Including the 2026 releases thus extends the evaluation to the latest models without changing its conclusions, and IdeaTaste, though built on a time split, does not rest on memorization.
 
\paragraph{Extending to later years.} The construction of IdeaTaste repeats for any year $Y$ by freezing the corpus at the end of $Y-1$, collecting the papers of $Y$ as the reference, and applying the same metrics. Each new model can thus be evaluated against a reference published after its training data end, which removes the concern altogether.

\paragraph{Information access of LLM agents and human researchers.}
LLM agents can retrieve only literature published through 2025, whereas the 2026 papers are observed after the cutoff and their authors may have had access to work released during 2026. However, matching the information of an LLM agent to that of a paper would require knowing what each author had read and restricting retrieval to it, which cannot be reconstructed after publication. IdeaTaste aligns what can be aligned. Every population searches the same corpus, which ends on the same date, and runs under the same budget, prompt, and judge, so this design leaves comparisons among LLM agents intact. The human level is read accordingly, as a reference line for where research actually went rather than as a competing arm under matched conditions.

\subsection{A Post-Cutoff Slice of the Reference for Leakage Analysis}
\label{app:holdout}

The controls above compare models released before and after the reference year. A sharper test splits the reference itself. Gemini 3.5 Flash-Lite is the one 2026 release whose model card states a data cutoff, March 2026, so its training data may cover the reference papers published in January to March 2026 and cannot cover those published from April onward~\citep{google2026gemini35flashlite}. Scoring it separately on the two halves gives a within-model comparison in which only exposure differs. The other three 2026 releases state no data cutoff, and their release dates fall in the last two months of the reference or after it, so no comparable split exists and we leave them out of this analysis.

Table~\ref{tab:holdout} reports matching for Gemini 3.5 Flash-Lite on the full reference and on each half, each half with its own human row. Exposure does not help. Citation matching runs against it, higher after the cutoff in all sixteen cells, and the human papers move the same way, so that difference belongs to the papers rather than to the model. Idea matching, where memorization would matter most, comes out even: read against the human level of its own half, the earlier half leads in nine of the sixteen cells. Its raw scores do lean earlier, in thirteen of sixteen, but so do the human rows in three of the four areas, which is why. The ranking of the four methods is the same in all three blocks, and the shortfall reported in Sec.~\ref{sec:main} holds in each of them.

\begin{table}[tb]
\centering
\small
\setlength{\tabcolsep}{4pt}
\caption{\MatchCite{} and \MatchIdea{} for Gemini 3.5 Flash-Lite against the full 2026 reference and against the two halves that its March 2026 data cutoff separates. Each block header gives the number of reference papers per area, and the Human row scores the human papers of that block against one another. Bold marks the best of the four methods in each column within a block. }
\label{tab:holdout}
\resizebox{\textwidth}{!}{%
\begin{tabular}{l *{8}{r}}
\toprule
& \multicolumn{2}{c}{Machine learning systems} & \multicolumn{2}{c}{Code generation} & \multicolumn{2}{c}{World models} & \multicolumn{2}{c}{Automated research} \\
\cmidrule(lr){2-3}\cmidrule(lr){4-5}\cmidrule(lr){6-7}\cmidrule(lr){8-9}
Population & Cite & Idea & Cite & Idea & Cite & Idea & Cite & Idea \\
\midrule
\rowcolor{lightBlue}\multicolumn{9}{l}{\textbf{Full reference}, January to August 2026 (5{,}188 / 3{,}375 / 1{,}867 / 1{,}136 papers)} \\
\addlinespace[2pt]
Human & 20.84 & 13.56 & 18.75 & 11.25 & 16.38 & 13.87 & 18.51 & 10.53 \\
\addlinespace[2pt]
\textsc{free} & 3.73 & 1.36 & 3.72 & 1.52 & 4.89 & 3.38 & 7.18 & 2.22 \\
\textsc{topic} & 4.59 & 2.29 & 4.23 & \textbf{2.19} & \textbf{5.33} & 3.86 & 6.58 & 2.14 \\
\textsc{landscape} & 3.67 & 1.38 & 3.88 & 1.47 & 4.44 & 2.31 & \textbf{7.18} & 2.14 \\
\textsc{frontier} & \textbf{4.65} & \textbf{2.32} & \textbf{4.23} & 1.85 & 5.28 & \textbf{3.87} & 6.34 & \textbf{2.28} \\
\midrule
\rowcolor{lightBlue}\multicolumn{9}{l}{\textbf{Possibly seen}, January to March 2026 (1{,}876 / 1{,}192 / 599 / 380 papers)} \\
\addlinespace[2pt]
Human & 18.24 & 14.60 & 16.53 & 12.25 & 13.84 & 13.46 & 15.85 & 11.97 \\
\addlinespace[2pt]
\textsc{free} & 2.95 & 1.37 & 3.11 & 1.53 & 3.95 & 3.57 & \textbf{6.29} & 2.46 \\
\textsc{topic} & \textbf{3.75} & 2.18 & 3.79 & \textbf{2.31} & 4.08 & \textbf{4.29} & 5.68 & 2.08 \\
\textsc{landscape} & 2.96 & 1.53 & 3.29 & 1.36 & 3.44 & 2.54 & 6.12 & 2.43 \\
\textsc{frontier} & 3.75 & \textbf{2.59} & \textbf{3.82} & 2.01 & \textbf{4.15} & 4.10 & 5.35 & \textbf{2.51} \\
\midrule
\rowcolor{lightBlue}\multicolumn{9}{l}{\textbf{After the cutoff}, April to August 2026 (3{,}312 / 2{,}183 / 1{,}268 / 756 papers)} \\
\addlinespace[2pt]
Human & 20.55 & 12.97 & 18.35 & 10.70 & 16.18 & 14.06 & 18.21 & 10.44 \\
\addlinespace[2pt]
\textsc{free} & 4.06 & 1.36 & 3.90 & 1.51 & 5.07 & 3.28 & 7.24 & 2.10 \\
\textsc{topic} & 4.79 & \textbf{2.36} & \textbf{4.22} & \textbf{2.11} & \textbf{5.41} & 3.65 & 6.05 & \textbf{2.17} \\
\textsc{landscape} & 3.95 & 1.30 & 4.08 & 1.53 & 4.76 & 2.20 & \textbf{7.40} & 1.99 \\
\textsc{frontier} & \textbf{4.90} & 2.16 & 4.17 & 1.76 & 5.34 & \textbf{3.75} & 6.00 & 2.16 \\
\bottomrule
\end{tabular}%
}
\end{table}

\begin{figure}[t]
\centering
\vspace{-12pt}
\includegraphics[width=\textwidth]{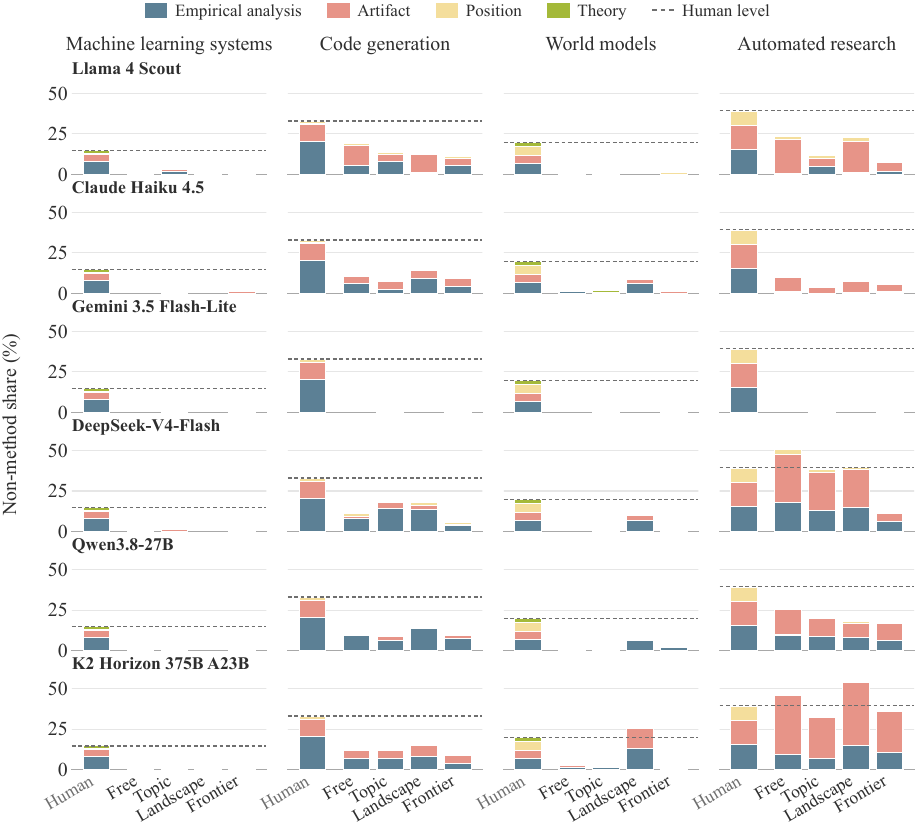}
\vspace{-12pt}
\caption{Contribution types across four research areas for the six models of Table~\ref{tab:main}, one row per model in release-date order. Stacked bars show the shares of non-method contributions in the human reference and each AI population; the remaining share consists of methods. The dashed rule carries the human non-method total of that area across the four arms.}
\label{fig:contribution}
\end{figure}

\subsection{Contribution Distribution of LLM Agents}
\label{sec:contribution}

LLM agents depart from the human record not only in which ideas they reach but in the kind of contribution they make. We label every proposal and 2026 paper with the contribution type its abstract states, one of method, empirical analysis, artifact, position, or theory as shown in Figure~\ref{fig:contribution}. Figure~\ref{fig:contribution} compares contribution types extracted from proposal and paper abstracts, with one row for each of the six models in Table~\ref{tab:main}. AI proposals generally concentrate more heavily on methods than human papers, especially in machine learning systems and world models, while theory and position papers are rare. The concentration is strongest for Gemini 3.5 Flash-Lite, which writes methods in at least $99\%$ of its proposals in every area and arm. Automated research is a partial exception, where DeepSeek-V4-Flash and K2 Horizon 375B A23B both produce more artifact contributions than the human reference, DeepSeek under \textsc{free}, \textsc{topic}, and \textsc{landscape} and K2 Horizon under all four arms. Matching human research taste therefore requires not only choosing problems and methods, but also deciding when to develop a method, conduct an empirical analysis, or build an artifact.

\subsection{Overlap with $\le 2025$ Literature}
\label{sec:prior-overlap}

A proposal may be restating work the corpus already holds. Therefore, we score each document exactly as \MatchIdea{} does, with the same slot judge and the same weights, but against its nearest papers in the corpus $U$ which LLM agent can access. The result is the weighted share of a document's slots that some earlier paper already states. Automated research is used throughout, and every population contributes as many proposals as the area has 2026 papers, so all rows in Table~\ref{tab:prior-overlap} rest on the same number of documents.

Every population repeats the earlier literature less than the 2026 papers do, so the proposals that miss the human record are not restatements of what the corpus already contains. The difference sits in the same two slots as the main result. Human papers restate an earlier approach or research plan many times more often than any population does, and those two slots carry most of the weight, while on problems and motivations the two sides are much closer. A proposal therefore tends to miss both the 2026 papers and the corpus at exactly the point where it commits to a mechanism. This does not make the proposals more original than the published work. The human reference contains cross-year continuations by the same group, which raise its overlap.

\begin{table}[t]
\centering
\small
\caption{Overlap with the corpus $U$ in automated research for DeepSeek-V4-Flash, scored as \MatchIdea{} is but against the papers in $U$. Every row covers 1{,}136 documents as Human 2026 papers. Overlap is the weighted share of a document's slots that an earlier paper already states; the four slot columns give the share of documents whose problem, motivation, proposed approach, or research plan is matched. These values describe measured overlap with earlier literature.}
\label{tab:prior-overlap}
\begin{tabular}{lrrrrr}
\toprule
Population & Overlap & Problem & Motivation & Approach & Plan \\
\midrule
Human & 10.26 & 19.10 & 37.06 & 3.43 & 2.99 \\
\textsc{free} & 6.34 & 12.68 & 30.99 & 0.26 & 0.09 \\
\textsc{topic} & 4.48 & 9.77 & 18.13 & 0.26 & 0.00 \\
\textsc{landscape} & 4.26 & 8.36 & 17.61 & 0.35 & 0.09 \\
\textsc{frontier} & 4.19 & 10.48 & 12.85 & 0.53 & 0.00 \\
\bottomrule
\end{tabular}
\end{table}

\begin{table}[t]
\centering
\small
\caption{Tag use and coverage in automated research for DeepSeek-V4-Flash, by axis and population.}
\label{tab:tagcoverage}
\resizebox{\linewidth}{!}{%
\begin{tabular}{llrrrrrrr}
\toprule
Axis & Population & \makecell{Tags\\used} & \makecell{Per\\doc} & \makecell{Human tags\\covered} & \makecell{Missed human\\mass (\%)} & \makecell{Head of the\\miss (\%)} & \makecell{Top-10\\share (\%)} & TVD \\
\midrule
\multicolumn{9}{l}{\emph{Task} (2025 vocabulary: 318 tags; human 2026 uses 200)} \\
& \popmark{pophuman}Human reference & 200 & 1.71 & --- & --- & --- & 43.5 & --- \\
& \popmark{popfree1136}Free & 36 & 2.17 & 34\,/\,200 & 49.7 & 40.5 & 94.4 & 0.717 \\
& \popmark{poptopic1136}Topic & 125 & 2.19 & 106\,/\,200 & 22.7 & 48.1 & 77.7 & 0.554 \\
& \popmark{poptag1136}Landscape & 39 & 2.18 & 37\,/\,200 & 47.9 & 41.7 & 91.5 & 0.663 \\
& \popmark{poptopicCL}Frontier & 192 & 2.81 & 155\,/\,200 & 5.5 & 53.3 & 55.3 & 0.499 \\
\midrule
\multicolumn{9}{l}{\emph{Method} (2025 vocabulary: 424 tags; human 2026 uses 268)} \\
& \popmark{pophuman}Human reference & 268 & 2.95 & --- & --- & --- & 41.6 & --- \\
& \popmark{popfree1136}Free & 103 & 3.09 & 91\,/\,268 & 25.3 & 32.3 & 68.3 & 0.659 \\
& \popmark{poptopic1136}Topic & 191 & 3.04 & 163\,/\,268 & 10.4 & 43.5 & 56.9 & 0.519 \\
& \popmark{poptag1136}Landscape & 114 & 3.06 & 102\,/\,268 & 22.3 & 28.8 & 64.5 & 0.589 \\
& \popmark{poptopicCL}Frontier & 220 & 3.71 & 178\,/\,268 & 8.9 & 47.3 & 47.7 & 0.494 \\
\midrule
\multicolumn{9}{l}{\emph{Task $\times$ Method} (possible: 318 $\times$ 424 = 134,832; human 2026 uses 2,871)} \\
& \popmark{pophuman}Human reference & 2,871 & 5.23 & --- & --- & --- & 9.7 & --- \\
& \popmark{popfree1136}Free & 710 & 6.76 & 299\,/\,2,871 & 74.0 & 5.9 & 34.9 & 0.853 \\
& \popmark{poptopic1136}Topic & 1,734 & 6.77 & 564\,/\,2,871 & 61.1 & 5.5 & 20.9 & 0.772 \\
& \popmark{poptag1136}Landscape & 847 & 6.76 & 333\,/\,2,871 & 72.1 & 5.6 & 26.9 & 0.831 \\
& \popmark{poptopicCL}Frontier & 3,441 & 10.44 & 889\,/\,2,871 & 49.6 & 5.4 & 12.3 & 0.757 \\
\bottomrule
\end{tabular}%
}
\end{table}

\begin{table}[t]
\centering
\scriptsize
\setlength{\tabcolsep}{4pt}
\caption{The three most used task tags, method tags, and task--method pairs of each population in automated research for DeepSeek-V4-Flash, with the share of that population's tag instances carrying each.}
\label{tab:tagtop3}
\resizebox{\linewidth}{!}{%
\begin{tabular}{@{}l l l l@{}}
\toprule
Population & Top task tags & Top method tags & Top task $\times$ method pairs \\
\midrule
\popmark{pophuman}Human 
  & \makecell[tl]{AI scientist (9\%)\\agent benchmarking (9\%)\\autonomous experimentation (5\%)}
  & \makecell[tl]{llm agent (10\%)\\multi-agent system (7\%)\\benchmark construction (7\%)}
  & \makecell[tl]{agent benchmarking $\times$ benchmark construction (2.3\%)\\AI scientist $\times$ llm agent (1.5\%)\\AI scientist $\times$ multi-agent system (1.1\%)} \\
\addlinespace[2pt]
\popmark{popfree1136}Free
  & \makecell[tl]{agent benchmarking (24\%)\\scientific claim verification (23\%)\\AI scientist safety (16\%)}
  & \makecell[tl]{benchmark construction (27\%)\\execution-based evaluation (12\%)\\provenance tracking (6\%)}
  & \makecell[tl]{agent benchmarking $\times$ benchmark construction (7.7\%)\\scientific claim verification $\times$ benchmark construction (6.3\%)\\scientific claim verification $\times$ execution-based evaluation (4.6\%)} \\
\addlinespace[2pt]
\popmark{poptopic1136}Topic
  & \makecell[tl]{agent benchmarking (23\%)\\experimental design (9\%)\\scientific claim verification (8\%)}
  & \makecell[tl]{benchmark construction (22\%)\\llm agent (8\%)\\execution-based evaluation (5\%)}
  & \makecell[tl]{agent benchmarking $\times$ benchmark construction (7.1\%)\\scientific claim verification $\times$ benchmark construction (2.1\%)\\experimental design $\times$ benchmark construction (1.9\%)} \\
\addlinespace[2pt]
\popmark{poptag1136}Landscape
  & \makecell[tl]{agent benchmarking (21\%)\\scientific hypothesis evaluation (17\%)\\experimental design (14\%)}
  & \makecell[tl]{benchmark construction (24\%)\\execution-based evaluation (8\%)\\llm agent (7\%)}
  & \makecell[tl]{agent benchmarking $\times$ benchmark construction (6.8\%)\\scientific hypothesis evaluation $\times$ benchmark construction (3.6\%)\\experimental design $\times$ benchmark construction (2.9\%)} \\
\addlinespace[2pt]
\popmark{poptopicCL}Frontier
  & \makecell[tl]{experimental design (9\%)\\agent benchmarking (9\%)\\scientific hypothesis evaluation (7\%)}
  & \makecell[tl]{llm agent (13\%)\\benchmark construction (13\%)\\information gain (4\%)}
  & \makecell[tl]{experimental design $\times$ llm agent (1.9\%)\\agent benchmarking $\times$ benchmark construction (1.9\%)\\AI scientist safety $\times$ benchmark construction (1.1\%)} \\
\bottomrule
\end{tabular}%
}
\end{table}

\begin{figure}[htbp]
\centering
\includegraphics[width=0.85\textwidth]{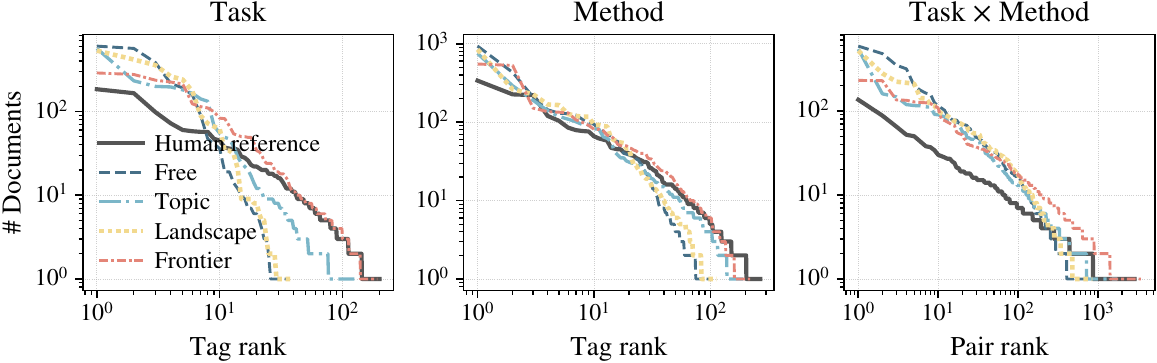}
\caption{Tags ordered by how many documents carry them, for the human reference and for each population in automated research for DeepSeek-V4-Flash. Each LLM agent curve falls away faster than the human curve and ends earlier, which is the concentration in Table~\ref{tab:tagcoverage} seen directly.}
\label{fig:tagrank}
\end{figure}

\subsection{Tag Coverage and Distributions}
\label{app:tags}

Table~\ref{tab:tagcoverage} gives the counts behind Figure~\ref{fig:tagcoverage}, on the task axis, the method axis, and on task--method pairs. Besides coverage it reports how many tags a document carries, how much of a population's tag mass sits in its ten most used tags, and the total variation distance between its tag distribution and the human one. Tags per document matters for reading the rest, since AI proposals carry more tags than the 2026 papers do, and \textsc{frontier} carries twice as many pairs, so part of its coverage comes from tagging more per proposal rather than from spreading further.

Coverage is only part of the picture, since a population can reach a tag and still use it far more often than the community does. Every population puts more of its tag mass in its ten most used tags than the 2026 papers do, and the excess is largest for \textsc{free} and \textsc{landscape}, which place almost all of their task mass there. \textsc{Frontier} comes closest to the human spread on all three axes, and Table~\ref{tab:tagtop3} shows why: its leading task and method tags carry roughly the share the 2026 papers give theirs, while the other populations open with a single tag several times heavier. Figure~\ref{fig:tagrank} gives the full distributions, where each population's curve falls away faster than the human curve and ends earlier.

What the populations concentrate on is the same in every case. Benchmark construction leads or ties for the lead among the method tags of all four, at roughly twice to four times its share among the 2026 papers, and their most used pairs are that tag joined to whichever task they favour. The shortfall is widest on pairs, where even \textsc{frontier} misses about half of the human mass and the other three miss between three fifths and three quarters. Reaching a community's tasks and its methods is therefore easier than combining them the way it does.

\subsection{Case Study}
\label{app:case} 

We provide the schemas behind Sec.~\ref{sec:case} and two additional cases to show where proposals agree with human research and where their mechanisms differ. Each case pairs one human paper with two proposals from the same model. We selected pairs from stored \MatchIdea{} judgments and inspected their descriptions for a shared problem and a clear mechanism comparison. The cases illustrate the metric and do not estimate how often each pattern occurs.

Each box includes the full title and abstract, followed by the four text slots available in the exported schema: problem, motivation, proposed approach, and research plan. Text and bracketed reference numbers are retained from the source records. Proposal A matches the human paper on problem and motivation but not on proposed approach or research plan; proposal B matches all four, according to the stored GPT-5.6 Luna judgments. These are automatic agreement labels, not human quality ratings. Proposal identifiers refer to the source runs; some judgments were also reused in derived population subsets. Comparisons outside the boxes are our interpretations.

\paragraph{Case 1: Expert reuse in memory-constrained inference.}
\label{app:case:remoe}
Both proposals address expert-fetch overhead. ReMoE and LARC fine-tune the router to increase temporal expert reuse. TopicRoute instead leaves the model unchanged and predicts expert demand from prompt topics to guide prefetching and scheduling. The choice is whether to change routing behavior or anticipate it.

{\scriptsize\ttfamily\raggedright
\begin{promptboxlong}
\textbf{Human paper: \href{https://arxiv.org/abs/2605.27081}{ReMoE}~\citep{zhu2026remoe}}\par\smallskip
\textbf{Title.} ReMoE: Boosting Expert Reuse through Router Fine-Tuning in Memory-Constrained MoE LLM Inference\par\smallskip
\textbf{Abstract.} Fine-grained Mixture-of-Experts (MoE) models sparsely activate only a subset of experts per token, reducing activated computation while maintaining high model capacity. However, in memory-constrained inference scenarios, only a small set of experts can be cached. Experts not in the cache must be fetched from slow external storage (e.g., UFS), leading to frequent evictions and substantial I/O overhead. We propose ReMoE, a router fine-tuning framework designed to boost token-wise expert reuse. ReMoE biases the router toward recently selected experts, producing temporally stable routing that better matches cache locality constraints. By increasing short-horizon expert reuse, ReMoE reduces expert fetches from storage without adding inference-time computation. Experiments on DeepSeek and Qwen models show that ReMoE improves expert reuse by 26\% while maintaining downstream task performance. Real-system evaluations further confirm these benefits, improving output throughput by 8.4\% under vLLM GPU-CPU expert offloading and reducing TPOT by 43.6-49.8\% under llama.cpp on Jetson Orin NX, corresponding to a 1.77-1.99$\times$ decode speedup across diverse workloads. Checkpoints and usage instructions are available at \url{https://github.com/BUAA-OSCAR/ReMoE}.\par\smallskip
\textbf{Problem.} In memory-constrained inference for fine-grained Mixture-of-Experts (MoE) models, only a small set of experts can be cached; experts outside the cache must be fetched from slow external storage, causing frequent evictions and substantial I/O overhead.\par\smallskip
\textbf{Motivation.} Reducing expert fetches would cut I/O overhead and improve throughput and latency in memory-constrained deployments.\par\smallskip
\textbf{Proposed approach.} The proposed framework fine-tunes the router of fine-grained MoE models to bias it toward recently selected experts, producing temporally stable routing that better matches cache locality constraints. This increases short-horizon expert reuse and reduces expert fetches from slow external storage without adding inference-time computation. Checkpoints and usage instructions are released.\par\smallskip
\textbf{Research plan.} The paper evaluates the framework through experiments on DeepSeek and Qwen models measuring expert reuse and downstream task performance, and through real-system evaluations under vLLM GPU-CPU expert offloading and llama.cpp on Jetson Orin NX measuring output throughput and time-per-output-token across diverse workloads.\par\smallskip\par
\end{promptboxlong}
}

{\scriptsize\ttfamily\raggedright
\begin{promptboxlong}
\textbf{AI proposal A: TopicRoute}\par\smallskip
\textbf{Model:} Qwen3.8-27B. \textbf{Method:} \textsc{frontier}.\par\smallskip
\textbf{Title.} TopicRoute: Prompt-Only Topic Modeling as a Zero-Forward-Pass Expert Predictor for MoE Offloading and Scheduling\par\smallskip
\textbf{Abstract.} Mixture-of-Experts (MoE) LLMs deployed on memory-constrained devices are bottlenecked by expert cache misses, which trigger costly host-to-device weight transfers on the critical path. Existing predictive prefetchers rely on intermediate forward-pass activations [3,4] or dedicated predictor networks fed by in-flight computation [5,6,10], so none can issue a prefetch before the first layer executes. We propose TopicRoute, a serving system that applies a lightweight online topic model to prompt embeddings at request arrival, predicts the per-request, per-layer expert activation set with zero forward-pass overhead, and uses this single topic signal to drive (i) proactive expert prefetching, (ii) demand-driven cache replacement, and (iii) routing-preserving request co-scheduling. Our core claim is falsifiable: on models with high local routing consistency as measured by the SRP/SCH metrics of recent MoE analysis [1], topic-predicted prefetching reduces time-to-first-token by at least 30\% and host-to-device transfer bytes by at least 20\% versus locality-only (LRU, segment-best) policies at a fixed cache budget, while on low-routing-consistency models the gain collapses to under 5\%, providing a built-in negative control. The approach is complementary to, not competitive with, router-decoupled designs [12] and device-level semantic co-scheduling [18]. We implement TopicRoute atop an open-source MoE offloading runtime and evaluate on Mixtral-8x7B, OLMoE-1B/7B, and DeepSeek-MoE across four workloads against seven offloading baselines including MoE-Infinity [2], ProMoE [3], Fate [4], and DuoServe-MoE [10], with an ablation matrix isolating the topic signal, the confidence gate, and each system component.\par\smallskip
\textbf{Problem.} Mixture-of-Experts LLMs on memory-constrained devices are bottlenecked by expert cache misses, which trigger costly host-to-device weight transfers on the critical path; existing predictive prefetchers depend on intermediate forward-pass activations or dedicated predictor networks fed by in-flight computation, so none can issue a prefetch before the first layer executes.\par\smallskip
\textbf{Motivation.} Removing this first-layer prefetch gap could take host-to-device transfer latency off the critical path and improve time-to-first-token for memory-constrained MoE serving.\par\smallskip
\textbf{Proposed approach.} The proposed serving system applies a lightweight online topic model to prompt embeddings at request arrival, predicts the per-request, per-layer expert activation set with zero forward-pass overhead, and uses this single topic signal to drive proactive expert prefetching, demand-driven cache replacement, and routing-preserving request co-scheduling. A confidence gate is included as a separate component. The system is implemented atop an open-source MoE offloading runtime and evaluated on Mixtral-8x7B, OLMoE-1B/7B, and DeepSeek-MoE across four workloads.\par\smallskip
\textbf{Research plan.} (1) The core claim is evaluated by comparing time-to-first-token and host-to-device transfer bytes against locality-only (LRU, segment-best) policies at a fixed cache budget, separating high- and low-routing-consistency models using the SRP/SCH metrics; the low-consistency condition serves as a built-in negative control in which the predicted gain should fall below 5\%. (2) Evaluation spans Mixtral-8x7B, OLMoE-1B/7B, and DeepSeek-MoE across four workloads against seven offloading baselines including MoE-Infinity [2], ProMoE [3], Fate [4], and DuoServe-MoE [10]. (3) An ablation matrix isolates the topic signal, the confidence gate, and each system component.\par\smallskip
\textbf{Automatic judgments.} Problem: match; motivation: match; proposed approach: no match; research plan: no match.\par
\end{promptboxlong}
}

{\scriptsize\ttfamily\raggedright
\begin{promptboxlong}
\textbf{AI proposal B: LARC}\par\smallskip
\textbf{Model:} Qwen3.8-27B. \textbf{Method:} \textsc{topic}.\par\smallskip
\textbf{Title.} LARC: Post-Training Only the Routers of Frozen MoE LLMs for Local Routing Consistency and Faster Offloaded Inference\par\smallskip
\textbf{Abstract.} Expert offloading for memory-constrained MoE LLM serving hinges on a model property - local routing consistency, the tendency of consecutive tokens to reuse the same experts - that varies widely across models and, in the literature we retrieved, has only ever been measured, never trained [1]. We propose LARC (Local-routing-Aware Router post-Training), a one-time procedure that post-trains only the gating networks of a frozen pretrained MoE LLM with a windowed, differentiable expert-set-overlap objective against a running cache-set surrogate, combined with the language-modeling loss and a global load-balance term. Our falsifiable claim: this model-side change measurably raises local routing consistency (SRP/SCH [1]) while holding perplexity degradation to \ensuremath{\leq} \textasciitilde{}2\%, and - at fixed GPU memory - lowers end-to-end decode latency under expert offloading by \ensuremath{\geq}10\% with zero per-step routing overhead, in contrast to inference-time cache-aware or batch-aware re-routing [2][3], which re-pay a per-token cost and make a per-request quality trade-off on every request. Where offloading systems adapt the system to whatever routing the model exhibits [4][5][6][7][8][9][10][20], LARC adapts the model to the deployment's memory constraint; where prior routing objectives are pretraining-time and target convergence, stability, or communication volume [11][12][13], LARC is a cheap deployment-aware post-processing step on a frozen model. We specify the objective, a router-only training protocol (100-500M tokens; router weights are <0.1\% of model parameters, so this is hours-to-a-day per model), an evaluation plan against system-side, static-compilation, and inference-time re-routing baselines, and the ablations and kill criteria that would falsify the claim.\par\smallskip
\textbf{Problem.} Expert offloading for memory-constrained MoE LLM serving depends on local routing consistency-the tendency of consecutive tokens to reuse the same experts-which varies widely across models and, in prior work, has only been measured, not trained; inference-time cache-aware or batch-aware re-routing incurs per-token cost and per-request quality trade-offs, while offloading systems adapt to whatever routing the model exhibits.\par\smallskip
\textbf{Motivation.} A model-side change that raises local routing consistency while preserving perplexity could lower end-to-end decode latency under expert offloading at fixed GPU memory without per-step routing overhead or per-request quality trade-off, and is cheap because only router weights are trained.\par\smallskip
\textbf{Proposed approach.} The proposed procedure post-trains only the gating networks of a frozen pretrained MoE LLM. It combines a windowed, differentiable expert-set-overlap objective computed against a running cache-set surrogate with the language-modeling loss and a global load-balance term. Training uses a router-only protocol over 100-500M tokens; router weights are less than 0.1\% of model parameters, making the one-time step hours-to-a-day per model.\par\smallskip
\textbf{Research plan.} The paper evaluates the consistency/perplexity claim and the latency claim against system-side, static-compilation, and inference-time re-routing baselines, with ablations and kill criteria to falsify the claims; fixed GPU memory is stated as the matched condition for the latency trade-off claim. The abstract specifies the 100-500M token router-only protocol but does not give exact metrics beyond SRP/SCH or the kill criteria.\par\smallskip
\textbf{Automatic judgments.} Problem: match; motivation: match; proposed approach: match; research plan: match.\par
\end{promptboxlong}
}

\paragraph{Case 2: Online adaptation of world models.}
\label{app:case:spread}
All three works seek adaptation to changing environments while retaining prior capabilities. SPREAD and DynaLoRA update low-rank adapters on a frozen model. ERALD instead stores observed transitions and combines retrieved predictions with the frozen model. DynaLoRA additionally proposes composing adapters for unseen combinations of dynamics changes; this extension is not part of the shared mechanism.

{\scriptsize\ttfamily\raggedright
\begin{promptboxlong}
\textbf{Human paper: \href{https://doi.org/10.1109/LRA.2026.3688061}{SPREAD}~\citep{moon2026spread}}\par\smallskip
\textbf{Title.} SPREAD: Scalable Pre-Trained World Model for Adaptive Dynamics Model\par\smallskip
\textbf{Abstract.} Autonomous robots must be capable of adapting not only to training datasets but also to unfamiliar environments. World models, which learn predictive dynamics of the environment, have been proposed to overcome the task-specific limitations of conventional RL. However, their capability is typically restricted to training distributions. In this paper, we introduce SPREAD, the first framework that enables a pre-trained world model to learn environmental changes online, and continuously. SPREAD freezes the pre-trained model and updates only a set of LoRA adapters. This design jointly achieves three goals: (i) real-time adaptation, (ii) robustness to forgetting without a replay buffer, and (iii) forward transfer. Experiments demonstrate that SPREAD reduces prediction error by 40\% within just 100 steps under diverse visual and dynamic changes, and that such adaptation directly translates into improved task success rates. Moreover, SPREAD validates its effectiveness in continual learning settings. By extending pre-trained world models to the domain of online continual learning, this work highlights world models that autonomously adapt to real-world dynamics.\par\smallskip
\textbf{Problem.} Pre-trained world models learn predictive environment dynamics but their capability is typically restricted to training distributions, so they cannot adapt to environmental changes online and continuously.\par\smallskip
\textbf{Motivation.} Autonomous robots must adapt to unfamiliar environments beyond training datasets; enabling world models to update online would overcome the task-specific limitations of conventional RL and support real-world dynamics.\par\smallskip
\textbf{Proposed approach.} The method freezes a pre-trained world model and updates only a set of low-rank adapters online. This keeps the base model fixed while learning environmental changes, and is designed to provide real-time adaptation, robustness to forgetting without a replay buffer, and forward transfer.\par\smallskip
\textbf{Research plan.} (1) The paper evaluates whether the adaptation reduces prediction error within 100 steps under diverse visual and dynamic changes. (2) It tests whether this adaptation translates into improved task success rates. (3) It validates the method in continual learning settings.\par\smallskip\par
\end{promptboxlong}
}

{\scriptsize\ttfamily\raggedright
\begin{promptboxlong}
\textbf{AI proposal A: ERALD}\par\smallskip
\textbf{Model:} DeepSeek-V4-Flash. \textbf{Method:} \textsc{topic}.\par\smallskip
\textbf{Title.} ERALD: Episodic Retrieval-Augmented Latent Dynamics - Gated Fusion of Parametric and Memory-Based Next-State Prediction for Gradient-Free World-Model Adaptation\par\smallskip
\textbf{Abstract.} Latent world models of the PlaNet/Dreamer family predict environment transitions with fixed network weights and consequently degrade the moment the environment changes (mass, friction, gravity, layout). Existing remedies either fine-tune the transition model on replay mixtures, inheriting a plasticity-forgetting trade-off and stale-data interference, or compress experience into an amortized context vector that fixed weights must still decode. We propose ERALD, a hybrid latent world model that keeps a frozen recurrent state-space model as a parametric prior and adds a bounded, continuously maintained episodic store of latent transitions. At every step of an imagined rollout, a contrastively learned, dynamics-aware key retrieves stored (state, action, next-state) items whose local dynamics match the current regime, and the parametric forecast is fused with the kernel-weighted episodic estimate by an online, residual-driven arbitration rule rather than by learned gating parameters. A regime shift is absorbed by appending observed transitions to the store and re-optimizing the policy inside ERALD imagination, with zero updates to the transition-model weights. We claim that under abrupt, recombinatorial dynamics switches, ERALD recovers a target return in fewer environment steps than replay- and generative-replay-based continual world models, achieves lower multi-step predictive error than both the frozen parametric model and a pure search-based world model given identical memory, and exhibits no measurable forgetting of the source regime. The claim is falsified if the hybrid never beats the better of its two components, if the online arbitration rule adds nothing over a fixed blend, if a replay-based baseline reaches the target return within the same environment budget, or if a dynamics-agnostic key space performs equally.\par\smallskip
\textbf{Problem.} Latent world models of the PlaNet/Dreamer family predict environment transitions with fixed network weights and consequently degrade when the environment changes (mass, friction, gravity, layout); existing remedies either fine-tune the transition model on replay mixtures, inheriting a plasticity-forgetting trade-off and stale-data interference, or compress experience into an amortized context vector that fixed weights must still decode.\par\smallskip
\textbf{Motivation.} Continual model-based control needs to adapt quickly to abrupt environment changes while retaining prior regimes, and avoiding transition-model fine-tuning removes the plasticity-forgetting trade-off and stale-data interference of existing remedies.\par\smallskip
\textbf{Proposed approach.} The proposed hybrid latent world model keeps a frozen recurrent state-space model as a parametric prior and adds a bounded, continuously maintained episodic store of latent transitions. During imagined rollouts, a contrastively learned dynamics-aware key retrieves stored (state, action, next-state) items whose local dynamics match the current regime. The parametric forecast is fused with a kernel-weighted episodic estimate by an online, residual-driven arbitration rule rather than learned gating parameters. Regime shifts are absorbed by appending observed transitions to the store and re-optimizing the policy inside the proposed framework's imagination, with zero updates to the transition-model weights.\par\smallskip
\textbf{Research plan.} The paper evaluates the claims under abrupt, recombinatorial dynamics switches: (1) sample efficiency of target-return recovery versus replay- and generative-replay-based continual world models, (2) multi-step predictive error versus the frozen parametric model and a pure search-based world model with identical memory, and (3) forgetting of the source regime. The stated falsification conditions are: the hybrid never beats the better of its two components, the online arbitration rule adds nothing over a fixed blend, a replay-based baseline reaches the target return within the same environment budget, or a dynamics-agnostic key space performs equally.\par\smallskip
\textbf{Automatic judgments.} Problem: match; motivation: match; proposed approach: no match; research plan: no match.\par
\end{promptboxlong}
}

{\scriptsize\ttfamily\raggedright
\begin{promptboxlong}
\textbf{AI proposal B: DynaLoRA}\par\smallskip
\textbf{Model:} DeepSeek-V4-Flash. \textbf{Method:} \textsc{free}.\par\smallskip
\textbf{Title.} DynaLoRA: Composable Low-Rank Latent-Dynamics Adapters for Forgetting-Free, Few-Shot World-Model Adaptation\par\smallskip
\textbf{Abstract.} Latent-dynamics world models (e.g., Dreamer-style RSSMs) are pretrained in one environment and must be adapted when dynamics change - mass, friction, gravity, or actuation scale - while retaining old skills and ideally composing knowledge of perturbations seen separately into models of combinations never jointly experienced. Existing approaches either fine-tune a monolithic latent residual for a single target environment [1], condition the world model on explicit or inferred context for zero-shot generalization [2, 3], mitigate forgetting with replay or generative rehearsal [4, 5, 6, 7], or discover modular mechanisms by training jointly across environments [8, 9, 10]. None provides a plug-in, parameter-efficient, forgetting-free-by-construction mechanism for adapting a frozen pretrained world model to a stream of dynamics-shifted environments, whose per-environment corrections are closed-form composable so an agent can imagine rollouts under compound perturbations (e.g., mass \ensuremath{\times} friction) it has never jointly experienced. We propose DynaLoRA: low-rank adapters on the linear projections of a frozen RSSM transition model, each trained on a handful of interactions under one perturbation by distilling the frozen posterior's latent sequence, leaving base weights untouched. Because each environment contributes an additive low-rank perturbation \ensuremath{\Delta}W\ensuremath{_{i}} = B\ensuremath{_{i}}A\ensuremath{_{i}} to the same base matrix, DynaLoRA is forgetting-free without replay, costs under \textasciitilde{}0.5\% trainable parameters per environment, and supports closed-form parameter-space composition for held-out compound dynamics. We validate on vision-based DMControl/CARL-style perturbations that few-shot adapters reach \ensuremath{\approx}90\% of converged full-fine-tuning fidelity and control return with under \textasciitilde{}0.5\% of the interaction budget, that a stream of five perturbations causes \textasciitilde{}zero forgetting of prior environments, and that composed adapters outperform matched-budget from-scratch training on unseen combinations.\par\smallskip
\textbf{Problem.} Existing approaches for adapting pretrained latent-dynamics world models to dynamics shift either fine-tune a monolithic residual for a single target environment, condition on context for zero-shot generalization, mitigate forgetting with replay or rehearsal, or discover modular mechanisms by training jointly across environments; none provides a plug-in, parameter-efficient, forgetting-free-by-construction mechanism for adapting a frozen pretrained world model to a stream of dynamics-shifted environments whose per-environment corrections are closed-form composable into models of compound perturbations never jointly experienced.\par\smallskip
\textbf{Motivation.} Such a mechanism would let an agent retain old skills across changing dynamics, adapt with few interactions and parameters, and imagine rollouts under combinations of perturbations seen separately, which is important for real-world deployment where dynamics drift over time and exact joint perturbations cannot be anticipated.\par\smallskip
\textbf{Proposed approach.} The method inserts low-rank adapters on the linear projections of a frozen RSSM transition model. Each adapter is trained on a handful of interactions under a single perturbation by distilling the latent sequence of the frozen posterior, leaving the base weights untouched. Each environment contributes an additive low-rank perturbation to the same base matrix, so the adapters are parameter-efficient and can be composed in closed form in parameter space for compound dynamics. The adapters are intended to be plug-in and forgetting-free by construction.\par\smallskip
\textbf{Research plan.} Validation uses vision-based DMControl/CARL-style dynamics perturbations. The evidence is: (1) few-shot adapters are compared to converged full fine-tuning on fidelity and control return under a reduced interaction budget; (2) a stream of five perturbations is used to measure forgetting of prior environments; (3) composed adapters are compared to matched-budget from-scratch training on unseen compound perturbations. Success conditions implied by the abstract are approaching full-fine-tuning performance, negligible forgetting, and superior performance on composed compounds; no explicit failure thresholds are stated. The comparisons for the composed setting hold interaction budget matched; other matched factors are not detailed.\par\smallskip
\textbf{Automatic judgments.} Problem: match; motivation: match; proposed approach: match; research plan: match.\par
\end{promptboxlong}
}

\paragraph{Case 3: Joint KV-cache eviction and quantization.}
\label{app:case:rdkv}
All three works jointly allocate storage to retention and precision. RDKV and ASYM-RD use a rate-distortion objective and reverse water-filling, with eviction represented by zero bits. TriAdmit learns retention scores for a three-action controller and passes them to an online budget solver. RDKV and ASYM-RD still differ in their distortion definitions and allocation timing. Their agreement concerns the allocation principle, not identical implementations.

{\scriptsize\ttfamily\raggedright
\begin{promptboxlong}
\textbf{Human paper: \href{https://arxiv.org/abs/2605.08317}{RDKV}~\citep{zhang2026rdkv}}\par\smallskip
\textbf{Title.} RDKV: Rate-Distortion Bit Allocation for Joint Eviction and Quantization of the KV Cache\par\smallskip
\textbf{Abstract.} Large language models (LLMs) have shown strong performance across diverse tasks, but their inference with long input contexts is bottlenecked by memory size and bandwidth. The Key-Value (KV) cache size grows linearly with sequence length and needs to be re-read from off-chip high-bandwidth memory (HBM) to on-chip memory at every decoding step, resulting in memory-bound inference. Existing methods reduce the cache by either eviction or quantization, but typically treat the two in isolation. In this paper, we cast KV cache compression as a rate-distortion problem, under which eviction and quantization are two end-points of the same bit allocation scheme. This exposes the need to optimize them jointly, motivating our method, RDKV (Rate-Distortion KV cache compression). RDKV derives the weight of each token or channel from the distortion that compression induces on the attention computation. Based on these weights, it assigns each token or channel a bit-width ranging from full precision down to zero bits guided by reverse water-filling, applied once after the prefilling stage. Experiments on LongBench, RULER, and InfiniteBench show that RDKV outperforms the best evaluated baseline by 9.1\% on average. On LongBench it recovers 97.81\% of full-cache accuracy with only 2.48\% cache retention. Compared with full-cache FlashAttention-2 decoding, it achieves 4.5x decode speedup and 1.9x peak memory reduction with 128K context length, while maintaining comparable performance.\par\smallskip
\textbf{Problem.} Inference with long input contexts is bottlenecked by memory size and bandwidth because the KV cache grows linearly with sequence length and must be re-read from off-chip high-bandwidth memory to on-chip memory at every decoding step, making decoding memory-bound.\par\smallskip
\textbf{Motivation.} Existing KV cache compression methods reduce the cache through eviction or quantization in isolation, leaving unexploited the potential gains from jointly optimizing the two under a shared bit allocation.\par\smallskip
\textbf{Proposed approach.} The method casts KV cache compression as a rate-distortion problem. It derives the weight of each token or channel from the distortion that compression induces on the attention computation, then assigns each token or channel a bit-width from full precision down to zero bits using reverse water-filling, applied once after the prefilling stage.\par\smallskip
\textbf{Research plan.} Evidence consists of experiments on LongBench, RULER, and InfiniteBench comparing against the best evaluated baseline. LongBench evaluation reports full-cache accuracy recovery as a function of cache retention, and 128K-context decoding is compared with full-cache FlashAttention-2 on decode speed and peak memory while checking comparable performance. No explicit failure condition or ablation design is stated.\par\smallskip\par
\end{promptboxlong}
}

{\scriptsize\ttfamily\raggedright
\begin{promptboxlong}
\textbf{AI proposal A: TriAdmit}\par\smallskip
\textbf{Model:} DeepSeek-V4-Flash. \textbf{Method:} \textsc{topic}.\par\smallskip
\textbf{Title.} TriAdmit: Learned Three-Way KV-Cache Admission - Retain, Demote, Evict - under Joint Byte Budgets\par\smallskip
\textbf{Abstract.} Long-context LLM inference is bottlenecked by a KV cache whose bytes and per-decode attention cost grow linearly with context length. Prior work attacks this by learning a binary write-time policy that evicts or keeps each token [1,2,3], by quantizing an entire cache at layer/chunk granularity [7,8,9,10], or by heuristically mixing the two [4,5,6]. These literatures are split along a false axis: a token's storage is not a keep/drop coin flip, nor is precision a property of a layer - each cached element has a rate-retention curve describing how much future-generation utility is lost as its storage is reduced from full precision to low precision to zero bits. We propose TriAdmit, a write-time admission controller that learns, for each (token, layer, head), a joint action over \{retain at full precision, demote to low precision, evict\}, and feeds the resulting per-element retention scores to an online budget solver that respects both byte and attention budgets. The core falsifiable claims are: (C1) there exist byte budgets at which all three actions are jointly necessary - the best two-action oracle cannot match the three-action oracle; and (C2) a lightweight distillation-trained scorer predicts these retention curves well enough to come within a small margin of the oracle. TriAdmit is evaluated on LongBench, RULER, and LongMemEval with Llama-3.1 and Qwen models, in a paged layout where demoted tokens live in separate low-precision pages and evicted tokens cost zero bytes and zero decode bandwidth.\par\smallskip
\textbf{Problem.} Long-context LLM inference is bottlenecked by a KV cache whose bytes and per-decode attention cost grow linearly with context length. Existing methods treat token storage as a binary keep/evict choice or treat precision as a layer/chunk-level property, missing that each cached element has a rate-retention curve describing how much future-generation utility is lost as its storage is reduced.\par\smallskip
\textbf{Motivation.} Modeling storage utility as a per-element rate-retention curve makes it possible to allocate scarce byte and attention budgets where they matter most and avoid paying full cost for tokens with little future-generation utility.\par\smallskip
\textbf{Proposed approach.} The proposed admission controller learns, for each (token, layer, head), a joint action over retaining at full precision, demoting to low precision, and evicting, and feeds the resulting per-element retention scores to an online budget solver that respects both byte and attention budgets. The scorer is trained via distillation to predict retention curves. In a paged layout, demoted tokens live in separate low-precision pages and evicted tokens cost zero bytes and zero decode bandwidth.\par\smallskip
\textbf{Research plan.} Evidence for the first claim is an oracle comparison across byte budgets, with the failure condition being that no budget exists where the three-action oracle beats every two-action oracle. Evidence for the second claim is comparing the distillation-trained scorer against the oracle, with the failure condition being that its predicted retention curves do not come within the claimed small margin. The full system is evaluated on LongBench, RULER, and LongMemEval with Llama-3.1 and Qwen models in a paged layout where demoted tokens live in separate low-precision pages and evicted tokens cost zero bytes and zero decode bandwidth.\par\smallskip
\textbf{Automatic judgments.} Problem: match; motivation: match; proposed approach: no match; research plan: no match.\par
\end{promptboxlong}
}

{\scriptsize\ttfamily\raggedright
\begin{promptboxlong}
\textbf{AI proposal B: ASYM-RD}\par\smallskip
\textbf{Model:} DeepSeek-V4-Flash. \textbf{Method:} \textsc{frontier}.\par\smallskip
\textbf{Title.} ASYM-RD: Rate-Distortion-Optimal Joint Bit-Allocation and Token Retention for the KV Cache under a Single Memory Budget\par\smallskip
\textbf{Abstract.} KV-cache compression is currently attacked one axis at a time: token eviction [1,2,3], uniform or mixed-precision quantization [4,5,6,7,8,9,10,11], or low-rank/channel compression [12]. The token-precision trade-off study [13] shows empirically that more tokens at lower precision usually beats fewer tokens at higher precision, but treats the split as a discrete choice, and existing mixed-precision KV papers allocate precision with layer- or channel-level saliency heuristics [5,6,7,8,9]. We propose to formulate KV compression as one rate-distortion allocation problem, solved once per model on a short calibration pass, whose distortion is the attention-output error, not the K/V reconstruction error. A first-order expansion of the softmax shows that a key error at token j is weighted by p\_j\ensuremath{^{2}}\ensuremath{\Vert}v\_j \ensuremath{-} o\ensuremath{\Vert}\ensuremath{^{2}} while a value error is weighted only by p\_j\ensuremath{^{2}}, so keys and values must be allocated asymmetrically as a matter of theory, not convention. Minimizing D + \ensuremath{\lambda}R over (layer, head, token, channel-group) cells gives a reverse-water-filling rule in which eviction is the interior b = 0 corner rather than a separate heuristic. We will test two falsifiable claims at equal KV bytes: (i) joint allocation beats uniform-bit, attention-saliency, layer-wise mixed-precision, and evict-then-uniform-precision baselines on RULER/LongBench, with the margin growing as the budget falls below 2 bits/token; (ii) the predicted key-sensitivity rank correlates better with measured per-token bit-shift damage than the attention score does, which would explain why low-bit KV methods that protect high-attention tokens still lose retrieval accuracy.\par\smallskip
\textbf{Problem.} KV-cache compression is currently attacked one axis at a time: token eviction, uniform or mixed-precision quantization, or low-rank/channel compression. The token-precision trade-off treats the split as a discrete choice, and existing mixed-precision methods allocate precision with layer- or channel-level saliency heuristics.\par\smallskip
\textbf{Motivation.} A jointly solved rate-distortion allocation with attention-output distortion could enable principled asymmetric key/value allocation and better compression under low KV budgets.\par\smallskip
\textbf{Proposed approach.} Formulate KV compression as one rate-distortion allocation problem over (layer, head, token, channel-group) cells, solved once per model on a short calibration pass, with distortion defined as attention-output error rather than K/V reconstruction error. A first-order expansion of the softmax weights key error at token j by p\_j\ensuremath{^{2}}//v\_j\ensuremath{-}o//\ensuremath{^{2}} and value error by p\_j\ensuremath{^{2}}; minimizing D + \ensuremath{\lambda}R yields a reverse-water-filling rule in which eviction is the interior b = 0 corner.\par\smallskip
\textbf{Research plan.} (1) Joint allocation is compared against uniform-bit, attention-saliency, layer-wise mixed-precision, and evict-then-uniform-precision baselines on RULER/LongBench at equal KV bytes; support requires joint allocation to outperform these baselines with a margin that grows as the budget falls below 2 bits/token, while failure would be baselines matching or reversing the trend. (2) The predicted key-sensitivity rank is correlated with measured per-token bit-shift damage; support requires this correlation to exceed that of the attention score, and failure would be the attention score correlating equally well or better, holding the KV budget matched.\par\smallskip
\textbf{Automatic judgments.} Problem: match; motivation: match; proposed approach: match; research plan: match.\par
\end{promptboxlong}
}

\paragraph{Case 4: Detecting code-benchmark contamination.}
\label{app:case:srdetection}
SrDetection and ReformGap detect leakage by comparing performance on an original problem and semantically equivalent variants. Shrinking Solution Ensembles instead tests whether memorization reduces the diversity of correct solutions. Both routes use controlled contamination experiments, but the experiments test different detection signals. Shared evaluation settings therefore do not by themselves establish agreement on the research plan.

{\scriptsize\ttfamily\raggedright
\begin{promptboxlong}
\textbf{Human paper: \href{https://doi.org/10.18653/v1/2026.findings-acl.252}{SrDetection}~\citep{li2026srdetection}}\par\smallskip
\textbf{Title.} SrDetection: A Self-Referential Framework for Data Leakage Detection in Code Large Language Models\par\smallskip
\textbf{Abstract.} Evaluating code large language models (Code LLMs) requires reliable detection of data leakage, where benchmark performance is artificially inflated by exposure to benchmark data during pre-training. Existing approaches either assume access to proprietary training corpora, rely on brittle heuristics such as timestamp filtering, or use external reference sets with manually tuned, non-generalizable thresholds. To address these limitations, we introduce SrDetection , a unified s elf- r eferential leakage detection framework for both gray-box (access to model logits) and black-box (access to model outputs) settings. SrDetection generates semantically equivalent variants of a benchmark sample and detects leakage by contrasting the model's behavior on the original versus its variants, flagging cases where the original is disproportionately easier for the model. We further design a controlled leakage detection testbed and evaluate SrDetection in this environment. Across different models and training stages, SrDetection improves average F1 by 21.52 points in the gray-box setting and 14.46 points in the black-box setting over strong baselines, demonstrating robust, threshold-independent leakage detection. Finally, a gray-box study of 15 widely used Code LLMs on four popular benchmarks reveals benchmark-specific leakage patterns beyond prior overlap-based analyses[Source code and data are available at \url{https://github.com/SMinL/SrDetectionCode} ].\par\smallskip
\textbf{Problem.} Reliable detection of data leakage is lacking for code large language models: existing approaches either assume access to proprietary training corpora, rely on brittle heuristics such as timestamp filtering, or require external reference sets with manually tuned, non-generalizable thresholds.\par\smallskip
\textbf{Motivation.} Without reliable leakage detection, benchmark performance of code large language models can be artificially inflated by pre-training exposure, so trustworthy evaluation requires a method that works in both gray-box and black-box settings without manual threshold tuning.\par\smallskip
\textbf{Proposed approach.} The proposed framework generates semantically equivalent variants of a benchmark sample and contrasts the model's behavior on the original with its behavior on the variants, flagging cases where the original is disproportionately easier. It supports gray-box operation using model logits and black-box operation using model outputs. A controlled leakage detection testbed is also constructed for evaluation.\par\smallskip
\textbf{Research plan.} (1) The detection claim is evaluated in the controlled leakage detection testbed with known leakage labels, comparing the proposed framework against strong baselines in gray-box and black-box settings using average F1. (2) Robustness and threshold independence are assessed across different models and training stages. (3) A gray-box study of 15 widely used code large language models on four popular benchmarks is conducted to examine benchmark-specific leakage patterns beyond prior overlap-based analyses.\par\smallskip\par
\end{promptboxlong}
}

{\scriptsize\ttfamily\raggedright
\begin{promptboxlong}
\textbf{AI proposal A: Shrinking Solution Ensembles}\par\smallskip
\textbf{Model:} DeepSeek-V4-Flash. \textbf{Method:} \textsc{landscape}.\par\smallskip
\textbf{Title.} Shrinking Solution Ensembles: Detecting Benchmark Contamination in Code LLMs from Sampling Behavior Under Controlled Leakage\par\smallskip
\textbf{Abstract.} Benchmark contamination - the presence of evaluation problems and their reference solutions in a code LLM's training data - is widely suspected of inflating pass@k leaderboards, yet existing contamination detectors are validated only on uncontrolled model-benchmark pairs in which the ground truth of leakage is unknown; several recent studies show these detectors disagree with each other, fail on instruction-tuned and RL-trained models, and can even be evaded. This proposal has two parts. First, we build a controlled-contamination corpus for code generation: open code LLMs are fine-tuned on known fractions (0-50\%) of uncontaminated code benchmarks, giving per-problem ground-truth leakage labels, dose-response measurements of pass@1/pass@k inflation, and quantification of leaderboard rank flips between contaminated and clean evaluations. Second, we introduce LeakScore, a black-box, per-instance contamination test exploiting a signal no existing detector uses: the diversity of a model's functionally correct solution ensemble. Our hypothesis is that leaked problems cause the model's conditional sampling distribution to collapse onto the memorized reference solution, so an executable sample set for a leaked problem has measurably lower cluster entropy and higher self-overlap than the model's samples on matched, certainly-unseen problems. We will test whether LeakScore detects per-problem leakage at fractions where Min-K\% Prob, DiffMIA, CDD, and PEARL-style signals are at chance, and whether the same signal tracks pass@k inflation and ranking distortion across models. If confirmed, LeakScore gives evaluators a leakage test needing no training-data access, no ground-truth contamination labels, and no benchmark rewrite, and provides a quantitative answer to when pass@k numbers should be distrusted.\par\smallskip
\textbf{Problem.} Existing code LLM benchmark contamination detectors are validated only on uncontrolled model-benchmark pairs with unknown leakage, and recent studies show they disagree, fail on instruction-tuned and RL-trained models, and can be evaded.\par\smallskip
\textbf{Motivation.} A leakage test needing no training-data access, ground-truth contamination labels, or benchmark rewrite would make pass@k leaderboards more trustworthy, and controlled fine-tuning of open code LLMs makes it possible to validate such tests against known leakage.\par\smallskip
\textbf{Proposed approach.} The work builds a controlled-contamination corpus by fine-tuning open code LLMs on known fractions (0-50\%) of uncontaminated code benchmarks, producing per-problem ground-truth leakage labels, dose-response measurements of pass@1/pass@k inflation, and quantification of leaderboard rank flips between contaminated and clean evaluations. It also develops a black-box, per-instance contamination test that measures the diversity of a model's functionally correct solution ensemble, using cluster entropy and self-overlap of executable samples relative to matched, certainly-unseen problems to detect memorization-induced sampling collapse.\par\smallskip
\textbf{Research plan.} (1) The controlled corpus will be used to test whether the proposed diversity signal detects per-problem leakage at contamination fractions where Min-K\% Prob, DiffMIA, CDD, and PEARL-style signals are at chance. (2) It will also test whether the same signal tracks pass@k inflation and ranking distortion across models. (3) Ground-truth leakage labels, dose-response measurements, and rank-flip quantification provide the evaluation setting; support for the detection claim requires the proposed signal to detect leakage in regimes where the named existing detectors perform at chance.\par\smallskip
\textbf{Automatic judgments.} Problem: match; motivation: match; proposed approach: no match; research plan: no match.\par
\end{promptboxlong}
}

{\scriptsize\ttfamily\raggedright
\begin{promptboxlong}
\textbf{AI proposal B: ReformGap}\par\smallskip
\textbf{Model:} DeepSeek-V4-Flash. \textbf{Method:} \textsc{free}.\par\smallskip
\textbf{Title.} ReformGap: Detecting Memorized Code-Benchmark Items via Execution-Grounded Prompt Reformulation\par\smallskip
\textbf{Abstract.} Data contamination in code-generation benchmarks such as HumanEval threatens the validity of model evaluation, yet existing detection tools either match against training corpora [5], require token-level loss or output-distribution access that API-only models do not expose [1][2], or sidestep the problem by continuously refreshing benchmarks [8][9]. We propose ReformGap, a purely black-box, per-item contamination detector built on the one asset code benchmarks have and text benchmarks lack: executable hidden test oracles. For each benchmark item, ReformGap generates semantically equivalent natural-language reformulations of the problem statement, accepts only reformulations verified by a reference model to be behavior-identical (execution-grounded equivalence), and measures the target model's reformulation gap \ensuremath{\Gamma} = pass@k(original) \ensuremath{-} pass@k(reformulation). The falsifiable core claim is that items whose (prompt\ensuremath{\to}solution) pair was memorized produce a large \ensuremath{\Gamma}, because recall of the memorized mapping is brittle under surface change, while items solved by genuine competence transfer produce \ensuremath{\Gamma} \ensuremath{\approx} 0. We validate under simulated contamination with known membership ground truth - the gold-standard protocol that real-world contamination studies lack [3][4] - and additionally test on API-only black-box models where no loss signal exists. If memorized items do not exhibit a reformulation gap, or if difficulty-matched competent solves do, the detector fails its own decisive experiment exactly as specified.\par\smallskip
\textbf{Problem.} Data contamination in code-generation benchmarks threatens evaluation validity, but existing detectors require training-corpus matching or internal model access that API-only models do not expose, or avoid detection by continually refreshing benchmarks; no purely black-box, per-item detector exists.\par\smallskip
\textbf{Motivation.} Code benchmarks uniquely provide executable hidden test oracles, so black-box equivalence checking via reformulation can make contamination detection possible without model internals.\par\smallskip
\textbf{Proposed approach.} For each benchmark item, the proposed detector generates semantically equivalent natural-language reformulations of the problem statement, retains only reformulations verified by a reference model to be behavior-identical via execution-grounded equivalence, and computes a gap score as pass@k on the original item minus pass@k on the reformulations. The detector operates purely black-box and uses the benchmark's executable hidden test oracles as the equivalence signal.\par\smallskip
\textbf{Research plan.} The core claim is tested under simulated contamination where membership ground truth is known, the protocol the abstract calls the gold standard, and additionally on API-only black-box models where no loss signal exists. The decisive failure condition is specified: if memorized items do not exhibit a gap score under reformulation, or if difficulty-matched competent solves do, the detector fails.\par\smallskip
\textbf{Automatic judgments.} Problem: match; motivation: match; proposed approach: match; research plan: match.\par
\end{promptboxlong}
}

\end{document}